\documentclass[lettersize,journal]{IEEEtran}
\usepackage{amsmath,amsfonts,amssymb}
\usepackage{algorithmic}
\usepackage{algorithm}
\usepackage{array}
\usepackage[caption=false,font=normalsize,labelfont=sf,textfont=sf]{subfig}
\usepackage{textcomp}
\usepackage{stfloats}
\usepackage{url}
\usepackage{verbatim}
\usepackage{graphicx}

\usepackage{amssymb}
\usepackage[hidelinks]{hyperref}
\usepackage{cite}
\usepackage{xcolor}
\usepackage{bm}
\usepackage[T1]{fontenc}
\usepackage{amsthm} 
\newtheorem{theorem}{Theorem} 

\newtheorem{proposition}{Proposition}
\usepackage{multirow}
\usepackage[table]{xcolor}
\usepackage{bbm}
\usepackage{caption}
\usepackage{hyperref}
 \hypersetup{colorlinks,linkcolor={blue},citecolor={blue},urlcolor={blue}}
 \usepackage{url}
\definecolor{peach}{RGB}{249,196,153}
\definecolor{lightblue}{RGB}{146, 205, 220}
\definecolor{softred}{RGB}{217, 149, 143}
\definecolor{skyblue}{RGB}{0, 176, 240}

\DeclareMathOperator{\Ent}{Ent}

\definecolor{peach}{RGB}{249,196,153}
\definecolor{lightblue}{RGB}{146, 205, 220}
\definecolor{softred}{RGB}{217, 149, 143}
\definecolor{skyblue}{RGB}{0, 176, 240}

\IEEEpubidadjcol

\begin{document}

\title{Post-Hoc Split-Point Self-Consistency Verification for Efficient, Unified Quantification of Aleatoric and Epistemic Uncertainty in Deep Learning}

\author{Zhizhong Zhao, and Ke  Chen,~\IEEEmembership{Senior Member,~IEEE}
\thanks{The authors are with the Department of Computer Science, University of Manchester, Manchester M13 9PL, U.K. (e-mail: zhizhong.zhao@postgrad.manchester.ac.uk; ke.chen@manchester.ac.uk).
}
}


\maketitle

\begin{abstract}
Uncertainty quantification (UQ) is vital for trustworthy deep learning, yet existing methods are either computationally intensive, such as Bayesian or ensemble methods, or provide only partial, task-specific estimates, such as single-forward-pass techniques.
In this paper, we propose a post-hoc single-forward-pass framework that jointly captures aleatoric and epistemic uncertainty without modifying or retraining pretrained models.
Our method applies \emph{Split-Point Analysis} (SPA) to decompose predictive residuals into upper and lower subsets, computing \emph{Mean Absolute Residuals} (MARs) on each side. We prove that, under ideal conditions, the total MAR equals the harmonic mean of subset MARs; deviations define a novel \emph{Self-consistency Discrepancy Score} (SDS) for fine-grained epistemic estimation across regression and classification. 
For regression, side-specific quantile regression yields prediction intervals with improved empirical coverage, which are further calibrated via SDS. 
For classification, when calibration data are available, we apply SPA-based calibration identities to adjust the softmax outputs and then compute predictive entropy on these calibrated probabilities.
Extensive experiments on diverse regression and classification benchmarks demonstrate that our framework matches or exceeds several state-of-the-art UQ methods while incurring minimal overhead.
\end{abstract}

\begin{IEEEkeywords}
Uncertainty quantification, split-point self-consistency, aleatoric-epistemic disentanglement,  calibration, trustworthy deep learning 
\end{IEEEkeywords}

\section{Introduction}
\label{sect:intro}

\IEEEPARstart{U}{ncertainty} quantification (UQ) in machine learning (ML) aims to quantify uncertainties associated with model predictions, typically distinguishing between \textit{aleatoric} (data) uncertainty, which stems from intrinsic data variability, and \textit{epistemic} (model) uncertainty, which arises from limitations in the model itself \cite{malinin2018predictive, KIUREGHIAN2009105, Hullermeier2021}. UQ is not only critical for improving the reliability and interpretability of ML models but also indispensable in safety-critical applications such as autonomous driving and AI-based medical diagnostics \cite{abdar2021review, begoli2019need, 10.1145/3461702.3462571, vashney2022trustworthy}.

Numerous UQ methods have been proposed \cite{depeweg2018decomposition, valdenegro2022deeper, lockwood2022review, gawlikowski2023survey, he2023survey, angelopoulos2022gentleintroductionconformalprediction}. Bayesian inference and ensemble approaches yield high quality uncertainty estimates, but their use in deep learning (DL) is hindered by substantial computational cost. Conformal prediction offers robust guarantees, yet it requires an exchangeable calibration set and does not distinguish aleatoric from epistemic uncertainty. Accordingly, recent efforts have shifted towards efficient single-forward-pass UQ methods for DL \cite{gawlikowski2023survey}.  
Despite their efficiency, these techniques suffer from four main limitations: 
(i) reliance on explicit distributional assumptions, causing misaligned calibration \cite{Salem2020PredictionIS,malinin2018predictive,sensoy2018evidential,amini2020deep,mukhoti2023deep};   
(ii) imprecise epistemic estimates that function more as \textit{out-of-distribution} (OOD) detectors than fine-grained uncertainty measures \cite{he2023survey, gawlikowski2023survey, bengs2022pitfalls, bengs2023second, shen2024uncertaintyquantificationcapabilitiesevidential}; 
(iii) separate estimation of aleatoric and epistemic uncertainty in regression, leading to misaligned \textit{predictive intervals} (PIs) and calibration errors \cite{Salem2020PredictionIS};  and
(iv) with the sole exception of \cite{tagasovska2019single}, no unified method quantifies both uncertainty types, supports diverse ML tasks, and integrates seamlessly with an already deployed DL model (hereinafter termed the \emph{base model}) without modifying its architecture or retraining.

\begin{figure*}[tbp]
  \centering
  \includegraphics[width=0.9\linewidth]{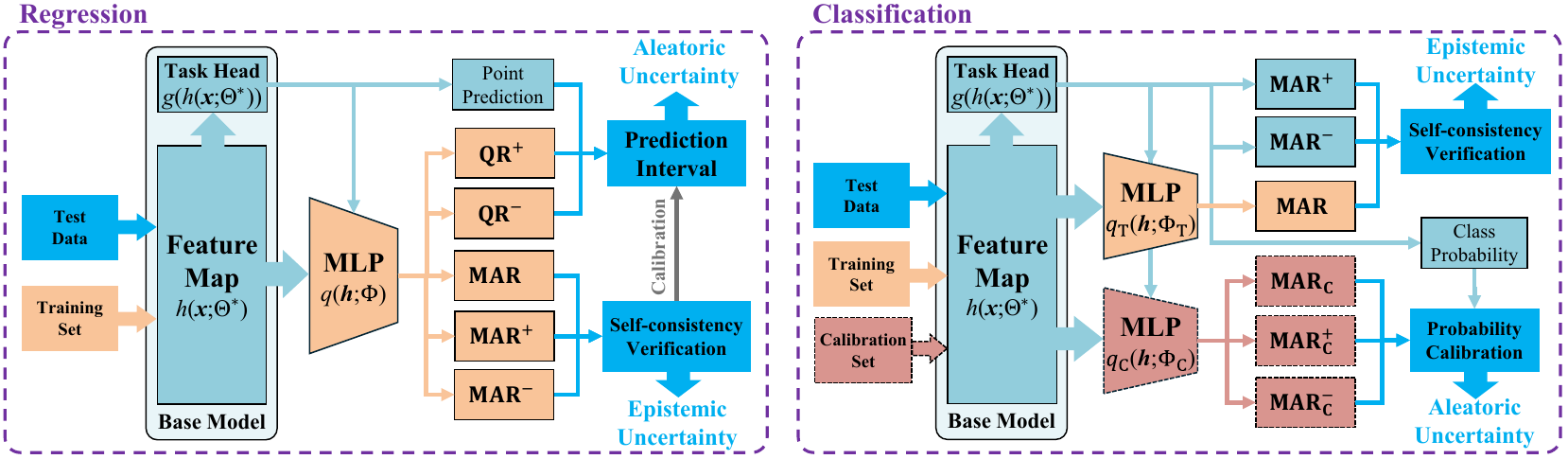}
  \caption{Our unified UQ framework, based on the split-point analysis and the self-consistency principle (Section \ref{sect:problem-foundation}), operates as follows. \textbf{In the left panel (Regression)}, we leverage the \textbf{\textcolor{lightblue}{base model}} and employ an \textbf{\textcolor{peach}{MLP}} regressor to jointly learn \textbf{\textcolor{peach}{split-point QR}} (\textcolor{peach}{$\mathrm{QR}^+,\mathrm{QR}^-$}) and the three \textbf{\textcolor{peach}{MARs}} (\textcolor{peach}{$\mathrm{MAR},\mathrm{MAR}^+,\mathrm{MAR}^-$}) on the \textbf{\textcolor{peach}{training set}} (Section \ref{subsect:net-learning-regress}). For \textbf{\textcolor{skyblue}{test data}}, the trained MLP produces MAR estimates via its three MAR heads for self-consistency verification to quantify \textbf{\textcolor{skyblue}{epistemic uncertainty}} (Section \ref{subsect:epistemic_uq}); the split-point QR heads, together with the base model’s \textbf{\textcolor{lightblue}{point prediction}}, yield \textbf{\textcolor{skyblue}{PIs}} for \textbf{\textcolor{skyblue}{aleatoric uncertainty}}, which are further calibrated by self-consistency verification (Section \ref{subsect:aleatoric-regress}).
\textbf{In the right panel (Classification)}, we leverage the \textbf{\textcolor{lightblue}{base model}} and employ an \textbf{\textcolor{peach}{MLP}} to learn the total \textbf{\textcolor{peach}{MAR}} 
on the \textbf{\textcolor{peach}{training set}} (Section \ref{subsect:net-learning-class}). 
For \textbf{\textcolor{skyblue}{test data}}, the trained MLP’s total \textbf{\textcolor{peach}{MAR}} estimate and two base-model derived \textcolor{lightblue}{$\mathrm{MAR}^+$, $\mathrm{MAR}^-$} are used together in self‐consistency verification to quantify \textbf{\textcolor{skyblue}{epistemic uncertainty}} (Section~\ref{subsect:epistemic_uq}).
When a \textbf{\textcolor{softred}{calibration set}} is available, we train another \textbf{\textcolor{softred}{MLP}} to learn the three \textbf{\textcolor{softred}{MARs}} (\textcolor{softred}{$\mathrm{MAR}_{\mathrm{C}},\mathrm{MAR}_{\mathrm{C}}^+,\mathrm{MAR}_{\mathrm{C}}^-$}) on this set (Section \ref{subsect:net-learning-class}). For \textbf{\textcolor{skyblue}{test data}}, its MAP heads produce three estimates used in self-consistency verification to calibrate the base model's softmax confidence to capture \textbf{\textcolor{skyblue}{aleatoric uncertainty}} (Section \ref{subsect:aleatoric-class}).}  
  \vspace*{-2mm}
  \label{fig: Model_structure}
\end{figure*}

In this paper, we propose a unified UQ framework that directly addresses these limitations, as illustrated in Fig. ~\ref{fig: Model_structure}. To address limitation (i), our method avoids any distributional assumptions and, leveraging an existing DL base model, quantifies both aleatoric and epistemic uncertainty in a single forward pass via \emph{split-point analysis} (SPA). In SPA, predictive residuals are partitioned around the point-prediction into upper and lower subsets, and the corresponding split-point \emph{mean absolute residuals} (MARs) are estimated independently. Under heteroscedastic conditions, we prove that, for a perfect model, the total MAR equals the harmonic mean of its subset MARs; this self-consistency constraint forms our theoretical foundation. For an imperfect model, deviations from the harmonic identity then yield a fine-grained measure of epistemic uncertainty for both regression and classification, hence improving upon coarse OOD-style detectors and addressing limitation (ii). 
In regression, our method jointly applies split-point quantile regression (QR) to the upper and lower subsets, producing PIs and estimating MARs, thereby addressing limitation (iii). Unlike simultaneous QR on the full dataset in prior work \cite{tagasovska2019single}, our split-point QR achieves improved empirical PI coverage; these intervals can then be calibrated via MAR-based self-consistency verification on the original training data to incorporate epistemic uncertainty without extra calibration sets. When calibration data are available for classification, we combine predictive-distribution entropy for aleatoric uncertainty \cite{kendall2017uncertainties} with SPA-based calibration identities derived from zero-included MARs to correct base model over- or under-confidence. Finally, our framework addresses limitation (iv) by operating post-hoc and model-agnostically on already deployed DL base models, as depicted in Fig. ~\ref{fig: Model_structure}.

Our main contributions are summarized as follows:
\begin{enumerate}
\item[(i)] We propose a single-pass unified UQ framework that quantifies both aleatoric and epistemic uncertainty, supports regression and classification, and seamlessly integrates with deployed DL models.
\item[(ii)] We provide rigorous theoretical underpinnings for the SPA-based self-consistency principle, establishing its validity for efficient and reliable UQ.
\item[(iii)] The self-consistency principle enables joint uncertainty modeling without any distributional assumptions, post-hoc calibration of PIs in regression without extra calibration sets, and confidence correction in classification when calibration data are available.
\item[(iv)] We conduct extensive evaluations on diverse benchmark and real-world datasets, demonstrating that our method is competitive with or outperforms several state-of-the-art UQ methods.
\end{enumerate}

\section{Background and Related Work}
\label{sect:related-work}
In this section, we review key UQ methods and situate our proposed UQ framework within this context.

\subsection{Single-Forward-Pass UQ Methods}
\label{subsect:single-pass} 
In general, these methods quantify uncertainty using a single forward pass of a deterministic task model \cite{gawlikowski2023survey}, either by modifying the model internally or by appending a post-hoc external uncertainty estimator without modifying the model. 

\subsubsection{Internal Methods}
\label{subsubsect:internal}
Internal methods require modifying a base model’s architecture or training loss to produce uncertainty estimates. Traditional methods, such as quantile regression \cite{QR_Koenker1978} and heteroscedastic regression \cite{kendall2017uncertainties}, train a new model from scratch with a specialized loss for regression. More recent internal methods include evidential approaches \cite{malinin2018predictive,sensoy2018evidential,amini2020deep}, deterministic UQ (DUQ) \cite{van2020uncertainty}, and distance-aware models like SNGP \cite{liu2020simple}. These techniques are usually tailored to a single task and often capture only aleatoric or epistemic uncertainty; when both are modeled, they rely on strong prior or distributional assumptions and may require additional posterior inference at test time, complicating integration with existing base models. In contrast, our UQ framework requires no prior or distributional assumptions, no extra posterior inference at test time, and efficiently captures both aleatoric and epistemic uncertainty for different tasks, all while integrating seamlessly with the base model without modifying its architecture or retraining.

\subsubsection{External Methods}
\label{subsubsect:external}

External methods operate post hoc by attaching a separate module to a base model to estimate uncertainty from its predictions and extracted features. Common techniques model the feature distribution \cite{10204383, lahlou2023deup}, often under a specific distributional assumption. While easy to apply, these methods are generally limited to classification and capture only abrupt epistemic uncertainty for OOD detection rather than providing fine grain epistemic estimates.
A unified external method, SQR-OC \cite{tagasovska2019single}, estimates aleatoric uncertainty in regression via simultaneous QR and epistemic uncertainty via one class classification independently. However, SQR treats the full target set globally, which can produce misaligned PIs for complex distributions, and OC, used as a one-class classifier relying on a linear feature-space assumption, functions solely as an OOD detector.
Our framework also acts externally but differs in several key respects: (i) it requires no explicit distributional assumptions and delivers fine-grained, reliable epistemic estimates for both in-Distribution (iD) and OOD data across regression and classification; (ii) in regression, it jointly quantifies aleatoric and epistemic uncertainty via the SPA, yielding PIs with improved coverage and an epistemic-score based interval calibration under complex distributions; and (iii) in classification, when calibration data are available, it applies split-point self-consistency verification to adjust softmax outputs, providing a novel calibration method superior to commonly used temperature scaling~\cite{guo2017calibration}.

\subsection{Multi-Forward-Pass UQ Methods}
\label{subsect:multi-pass}

In general, these methods require multiple forward passes at test time to estimate uncertainty.

\subsubsection{Bayesian and Ensemble}
\label{subsubsect:bayes-ensemble}
Bayesian and ensemble methods \cite{neal2012bayesian,welling2011bayesian,gal2016dropout,graves2011practical,hernandez2015probabilistic,louizos2017multiplicative,maddox2019simple,Lakshminarayanan2016SimpleAS,osband2016deep,wenzel2020good} require multiple forward passes at test time, yielding high quality uncertainty estimates, but incurring substantial computational and memory overhead which limits their practical use. In contrast, our framework requires only a single forward pass yet achieves UQ quality comparable to that of ensemble methods, since the three MAR estimates act as an implicit ensemble under the self-consistency constraint.

\subsubsection{Post-hoc Augmentation}
\label{subsubsect:augment}
Post hoc augmentation methods \cite{ayhan2018test,wang2019aleatoric} estimate aleatoric uncertainty via test-time data augmentation, requiring multiple forward passes over perturbed inputs, despite no model architectural change. In contrast, our framework delivers both aleatoric and epistemic uncertainty estimates in real time with a single forward pass through a dedicated UQ network.

\subsection{Conformal Prediction}
\label{subsect:conformal}
Conformal prediction (CP) \cite{angelopoulos2022gentleintroductionconformalprediction} is a model agnostic, distribution free, post hoc method that provides theoretical coverage guarantees for any ML model. However, CP requires a high quality calibration set and does not distinguish aleatoric from epistemic uncertainty. By contrast, our framework designed for DL remains model agnostic within that domain, disentangles aleatoric and epistemic uncertainty, and calibrates PIs in regression without additional calibration data, although it does not provide formal coverage guarantees.

\section{Problem Formulation and  Foundations}
\label{sect:problem-foundation}

In this section, we formulate the problem statement and establish the foundational elements of our UQ framework, including the split point analysis and the self consistency constraint.

\subsection{Problem Formulation}
\label{subsect:problem}

Consider a supervised dataset $\mathcal D = \{(\bm{x}_i, \bm{y}_i)\}_{i=1}^{|\mathcal D|}$, where each input $\bm{x}_i\in\mathcal X$ is paired with a target $\bm{y}_i\in\mathcal Y$.
A deployed DL model $f(\bm{x}; \Theta^*)$, parameterized by fixed parameters $\Theta^*$ and hereafter termed the \textit{base} model, has been trained on $\mathcal{D}$ to approximate the true mapping $F\colon \mathcal{X}\to\mathcal{Y}$, such that $f(\bm{x}; \Theta^*)\approx F(\bm{x})$.
 Given an unseen test set $\hat{\mathcal D}= \{\hat{\bm{x}}_i\}_{i=1}^{|\hat{\mathcal D}|}$ ($\mathcal D \cap \hat{\mathcal D} = \emptyset$), our goal is to accurately estimate both the aleatoric uncertainty, which stems from the inherent data noise $\varepsilon(\hat{\bm{x}}_i)$, and the epistemic uncertainty, which arises from the model's approximation $f(\hat{\bm{x}_i}; \Theta^*)$ of the true function $F(\hat{\bm{x}}_i)$.

For regression under heteroscedastic conditions, the underlying data-generating process is inherently stochastic and can be formalized as:
\begin{equation}
\label{eq:general-problem}
\bm{y} = F(\bm{x}) + \varepsilon(\bm{x}),
\end{equation}
where $F:\mathcal X\to\mathcal Y$ is the true deterministic function and $\varepsilon(\bm{x})$ is input-dependent noise.

To establish a unified UQ framework applicable to both regression and classification tasks, we extend the heteroscedastic regression setting in \eqref{eq:general-problem} to classification. For multi-class classification, from a probabilistic perspective, we interpret the softmax output\footnote{In binary classification, the labels naturally follow a Bernoulli distribution.}, which follows a multinomial distribution, as the expectation of a generalized Bernoulli distribution \cite{chen1999improved}:
\[
P(\bm{y}; \bm{p}) = \prod_{k=1}^{K} p_k^{y_k} (1 - p_k)^{1 - y_k},
\]
where $\bm{y} = (y_1, y_2, \cdots, y_K) \in \{0, 1\}^K$ is the one-hot encoded label vector, and $\bm{p} = (p_1, p_2, \cdots, p_K)$ denotes the predicted Bernoulli probabilities for the $K$ classes. Given a softmax output vector, $\tilde{\bm{y}} = f(\bm{x}; \Theta^*) = (\tilde{y}_1, \tilde{y}_2, \cdots, \tilde{y}_K )$,
we view each element $\tilde{y}_k \in \tilde{\bm{y}}$
as the expected value of a class-specific Bernoulli distribution, where $\tilde{y}_k$ signifies the probability that $\bm{x}$ belongs to class $k$, and $1-\tilde{y}_k$ is the probability that $\bm{x}$ belongs to any class other than $k$.

Leveraging this interpretation, we formulate heteroscedastic classification as a per-class problem for a given input $\bm{x}$:
\begin{equation}
\tilde{y}_k = F_k(\bm{x}) + \varepsilon_k(\bm{x}).
\label{eq:cls-obj}
\end{equation}
Here, $F_k(\bm{x}) \in \{0,1\}$ is the true binary indicator for class $k$, and 
$\varepsilon_k(\bm{x})$ captures the input-dependent noise specifically associated with class $k$. Under this formulation, the softmax score for each class, $\tilde{y}_k$, naturally serves as the expected value of the noisy indicator $y_k$.

\subsection{Split-point Analysis}
\label{subsect:spa}

As described in Section~\ref{sect:intro}, \emph{split-point analysis} (SPA) underpins our UQ framework. We apply SPA separately to regression (continuous targets) and classification (discrete labels), and adopt element-wise notation throughout in the rest of this section for clarity and consistency.

\subsubsection{SPA for Regression}
\label{subsubsect:spa-regress}

For each pair $(\bm x, y)\in\mathcal D$, let the base model prediction be $\tilde y = f(\bm{x}; \Theta^*)$ and define the residual $r = y - \tilde y$. We collect the set of input-residual pairs where the residual is non-zero:
\(
    \mathcal{R} = \{(\bm{x}, r) \mid r \neq 0, (\bm{x}, y) \in \mathcal{D}\}
\).
This set is then partitioned based on the sign of the residual into a set of positive residuals: 
\(
    \mathcal{R}^+ = \{(\bm{x}, r) \mid r > 0, (\bm{x}, y) \in \mathcal{D}\}
\),
and a set of negative residuals: 
\(
    \mathcal{R}^- = \{(\bm{x}, r) \mid r < 0, (\bm{x}, y) \in \mathcal{D}\}
\).
Here, $\mathcal R^+$ captures underestimation errors while $\mathcal R^-$ captures overestimation errors. Residuals with $r = 0$ are omitted, since they occur with negligible probability and have minimal effect on uncertainty estimation.

Based on the above SPA, we derive the total, upper side and lower side \emph{mean absolute residuals} (MARs) for any prediction \(\tilde y\) on input \(\bm{x} \in \mathcal{D}\):
\begin{align}
\mathrm{MAR}(\tilde y|\bm{x})   &= \mathbb{E}\bigl[|r| \mid (\bm {x'}, r)\in\mathcal R, \bm{x'}=\bm{x} \bigr], \nonumber\\
\mathrm{MAR}^{+}(\tilde y|\bm{x}) &= \mathbb{E}\bigl[|r| \mid (\bm {x'}, r)\in\mathcal R^+, \bm{x'}=\bm{x} \bigr], \label{eq:reg-MARs}\\
\mathrm{MAR}^{-}(\tilde y|\bm{x}) &= \mathbb{E}\bigl[|r| \mid (\bm {x'}, r)\in\mathcal R^-, \bm{x'}=\bm{x} \bigr]. \nonumber
\end{align}
These respectively measure the average magnitude of all residuals, underestimations, and overestimations.

\subsubsection{SPA for Classification}
\label{subsubsect:spa-class}

\label{subsubsect:spa-class}
In classification, labels are discrete. For each class \(k\), we use the base model's softmax probability \(\tilde{y}_k \in (0,1)\) and the one-hot label \(y_k \in \{0,1\}\), yielding a residual \(r_k = y_k - \tilde{y}_k\). Accordingly, we form \(\mathcal{R}_k = \{(\bm{x}, r_k) \mid r_k \neq 0, (\bm{x}, y_k) \in \mathcal{D}\}\), \(\mathcal{R}^+_k = \{(\bm{x}, r_k) \mid r_k > 0, (\bm{x}, y_k) \in \mathcal{D}\}\), and \(\mathcal{R}^-_k = \{(\bm{x}, r_k) \mid r_k < 0, (\bm{x}, y_k) \in \mathcal{D}\}\). 
Based on the heteroscedastic classification formulation in \eqref{eq:cls-obj}, we derive the total, upper-side and lower-side MARs for any prediction \(\tilde y_k\) on input \(\bm x \in \mathcal{D}\):
\begin{subequations} 
\label{eq:class_k-MARs}
\begin{align}
\mathrm{MAR}(\tilde{y}_k|\bm{x})   &= \mathbb{E} \bigl[|r_k| \mid (\bm {x'}, r_k)\in\mathcal R_k, \bm{x'}=\bm{x} \bigr] \nonumber\\
&=P_k(\bm x) (1 - \tilde{y}_{k}) + (1 - P_k(\bm x)) \tilde{y}_{k}, \label{eq:class_k-MARs-total}\\
\mathrm{MAR}^+(\tilde{y}_k|\bm{x}) &= \mathbb{E} \bigl[|r_k| \mid (\bm {x'}, r_k)\in\mathcal R^+_k, \bm{x'}=\bm{x} \bigr] \nonumber\\
&= 1 - \tilde{y}_{k},\label{eq:class_k-MARs-upper}\\
\mathrm{MAR}^-(\tilde{y}_k|\bm{x}) &= \mathbb{E} \bigl[|r_k| \mid (\bm {x'}, r_k)\in\mathcal R^-_k, \bm{x'}=\bm{x} \bigr] \nonumber\\
&= \tilde{y}_{k}. \label{eq:class_k-MARs-lower}
\end{align}
\end{subequations}
Here, \(P_k(\bm x)\) denotes the conditional class-frequency of class \(k\) for input $\bm x$ estimated from the training data in $\mathcal D$. 
Notably, $\mathrm{MAR}(\tilde{y}_k|\bm x)$ depends on the training data distribution, whereas $\mathrm{MAR}^{+}(\tilde{y}_k|\bm x)$ and $\mathrm{MAR}^{-}(\tilde{y}_k|\bm x)$ are derived directly from the base model's softmax output.

When a calibration dataset $\mathcal D_{\mathrm C} = \{(\bm{x}_{{\mathrm C},i}, \bm{y}_{{\mathrm C},i})\}_{i=1}^{|\mathcal D_{\mathrm C}|}~(\mathcal D_{\mathrm C} \ne \mathcal D)$ is available, we can use it to calibrate the initial UQ. To capture both the \emph{magnitude} and \emph{frequency} of under- or over-prediction made by the base model, we define  residuals for class \(k\) for any data point in 
$\mathcal D_{\mathrm C}$: 
\(r_{k} = y_{{\mathrm C},k} - \tilde{y}_{{\mathrm C},k}\), 
$r^+_{k} = \max\{r_{k}, 0\}$, and $r^-_{k} = \min\{r_{k}, 0\}$.
Based on the zero-included residuals, we form 
\(\operatorname{C}_{k} = \bigl\{(\bm{x}_{\mathrm C}, r_k) | (\bm x_{\mathrm C}, y_{{\mathrm C},k})\in\mathcal D_{\mathrm C} \bigr\} \), 
\(\operatorname{C}^+_k = \bigl\{(\bm{x}_{\mathrm C}, r_k^+) | (\bm x_{\mathrm C}, y_{{\mathrm C},k})\in\mathcal D_{\mathrm C} \bigr\} \), and 
\(\operatorname{C}^-_k = \bigl\{(\bm{x}_{\mathrm C}, r_k^-) | (\bm x_{\mathrm C}, y_{{\mathrm C},k})\in\mathcal D_{\mathrm C} \bigr\} \). 
Thus, we derive the total, upper-side, and lower-side \textit{zero-included MARs} for any prediction \(\tilde y_{{\mathrm C},k}\) on input \(\bm x \in \mathcal D_{\mathrm C}\):
\begin{align}
\mathrm{MAR}_{\mathrm{C}}(\tilde y_{{\mathrm C},k}|\bm{x})   &= \mathbb{E} \left[ |r_k| \mid (\bm {x'}, r_k)\in\operatorname C_k, \bm{x'}=\bm{x} \right] \nonumber\\
&=P_{{\mathrm C},k}(\bm x) (1 - \tilde y_{{\mathrm C},k}) + (1 - P_{{\mathrm C},k}(\bm x)) \tilde y_{{\mathrm C},k}, \nonumber\\
\mathrm{MAR}_{\mathrm{C}}^+(\tilde y_{{\mathrm C},k}|\bm{x}) &= \mathbb{E} \left[ |r_k| \mid (\bm {x'}, r_k)\in\operatorname C^+_k, \bm{x'}=\bm{x} \right] \nonumber\\
&= P_{{\mathrm C},k}(\bm x)\bigl( 1 - \tilde y_{{\mathrm C},k} \bigr), \label{eq:class_k-MARs-c}\\
\mathrm{MAR}_{\mathrm{C}}^-(\tilde y_{{\mathrm C},k}|\bm{x}) &= \mathbb{E} \left[ |r_k| \mid (\bm {x'}, r_k)\in\operatorname C^-_k, \bm{x'}=\bm{x} \right] \nonumber\\
&= \bigl( 1 - P_{{\mathrm C},k}(\bm x) \bigr) \tilde y_{{\mathrm C},k}. \nonumber
\end{align}
Here, \(P_{C,k}(\bm x)\) is the frequency of class \(k\) conditioned on input $\bm x$ and estimated from the calibration set \(\mathcal D_{C}\). In contrast to \eqref{eq:class_k-MARs}, \eqref{eq:class_k-MARs-c} employs these statistical estimates to weight each residual by its empirical likelihood, resulting in adjustments that vanish in well-calibrated regions and grow only where both error magnitude and occurrence frequency are high. The derivations of MARs in 
\eqref{eq:reg-MARs}, \eqref{eq:class_k-MARs} and \eqref{eq:class_k-MARs-c} are provided in Appendix~A. 

\subsection{Split-Point Self-Consistency Principle}
\label{subsect:self-consistency-principle}

We observe a general relationship among split-point statistical quantities that holds under any distribution over a finite set, and refer to this property as \textit{self-consistency constraint}:
\begin{theorem}[Self-Consistency Constraint]
\label{th:sc}
Let \(Y\) be a real-valued random variable with \(|\mathbb{E}[Y]|<\infty\). For a split-point $t \in \mathbb{R}$, define the total Mean Absolute Deviation (MAD), upper-side $\mathrm{MAD}^+$, and lower-side 
$\mathrm{MAD}^-$ by 
\begin{align*}
\mathrm{MAD} &= \mathbb{E}\bigl[\lvert Y - t \rvert \mid Y \ne t \bigr], \\
\mathrm{MAD}^+ &= \mathbb{E}\bigl[ Y - t \mid Y > t \bigr], \\
\mathrm{MAD}^- &= \mathbb{E}\bigl[ t - Y \mid Y < t \bigr].
\end{align*}
When $t = \mathbb{E}[Y]$, assuming 
$P(Y > t) > 0$ and $P(Y < t) > 0$, the following identity holds: 
\begin{equation}
\label{eq:sc}
\mathrm{MAD} = H\bigl(\mathrm{MAD}^+,\, \mathrm{MAD}^-\bigr) = \frac{2\, \mathrm{MAD}^+\, \mathrm{MAD}^-} {\mathrm{MAD}^+ + \mathrm{MAD}^-}, 
\end{equation} 
where $H(a,b) = 2ab/(a + b)$ denotes the harmonic mean. 
\end{theorem}
Theorem~\ref{th:sc} implies the following proposition:
\begin{proposition}[Minimum Discrepancy]
\label{prop-sc}
For any $t \in \mathbb{R}$ with $P(Y > t)>0$ and $P(Y < t)>0$, define the self-consistency discrepancy:
\[
\Delta(t) := \left| \mathrm{MAD} - H\bigl(\mathrm{MAD}^+,\, \mathrm{MAD}^-\bigr) \right|.
\]
Then $\Delta(t)$ attains its global minimum of zero when $t = \mathbb{E}[Y]$ and at any balance points where $\mathrm{MAD}^+ = \mathrm{MAD}^-$.
\end{proposition}
Proofs of Theorem~\ref{th:sc} and Proposition~\ref{prop-sc} are provided in Appendix~A. 
Under the self-consistency constraint of Theorem~\ref{th:sc}, Proposition~\ref{prop-sc} suggests that the discrepancy 
\(\Delta(\tilde{t})\) quantifies the deviation of any estimate \(\tilde{t}\in Y\) from the true mean \(\mathbb{E}[Y]\).
If predictive bias is interpreted as epistemic uncertainty and the MAD components can be estimated, then $\Delta(\tilde{t})$ serves as a natural metric for quantifying this uncertainty. 
Moreover, since \eqref{eq:sc} is homogeneous in $\mathrm{MAD}$, $\mathrm{MAD}^+$ and $\mathrm{MAD}^-$, $\Delta(\tilde{t})$ is invariant under their uniform scaling. Thus, aleatoric noise, which merely rescales all deviations, leaves $\Delta(\tilde{t})$ unchanged, whereas epistemic bias, by skewing $\mathrm{MAD}^+$ against $\mathrm{MAD}^-$, alters it, enabling separation of epistemic and aleatoric uncertainty in UQ.

For the zero-included MARs on calibration data, we have the\textit{ zero-included self-consistency constraint}:
\begin{proposition}[Calibration Identity]
\label{prop-sc-C}
Let \(\mathcal D_{\mathrm C}=\{(\bm{x}_{{\mathrm C},i},\bm{y}_{{\mathrm C},i})\}_{i=1}^{|\mathcal D_{\mathrm C}|}\) be a calibration set, where \(\mathcal D_{\mathrm C}\neq\mathcal D\). 
Then the zero‑included MARs from \eqref{eq:class_k-MARs-c} satisfy
\begin{subequations}\label{eq:MAR-C_relation}
\begin{align}
\mathrm{MAR}_{\mathrm C}(\tilde y_{{\mathrm C},k}|\bm{x})
&= \mathrm{MAR}^+_{\mathrm C}(\tilde y_{{\mathrm C},k}|\bm{x}) + \mathrm{MAR}^-_{\mathrm C}(\tilde y_{{\mathrm C},k}|\bm{x}), \label{eq:MAR-C_relation-1}\\
P_{\mathrm C,k}
= \tilde y_{\mathrm C,k} & + \mathrm{MAR}^+_{\mathrm C}(\tilde y_{{\mathrm C},k}|\bm{x}) - \mathrm{MAR}^-_{\mathrm C}(\tilde y_{{\mathrm C},k}|\bm{x}). \label{eq:MAR-C_relation-2}
\end{align}
\end{subequations}
\end{proposition}
The proof of Proposition~\ref{prop-sc-C} is provided in Appendix~A. 

According to statistical decision theory~\cite[Section~2.4]{hastie2009elements}, the optimal prediction under the mean squared loss is the conditional mean \( \mathbb{E}[Y|X] \). 
For unbiased model prediction \( \tilde{t} = \mathbb{E}[Y|X] \), the MAR coincides with the MAD.  In this case, for any  prediction \( \tilde{t} \) on $\bm x \in X$, the observed MARs in a sample serve as empirical estimates of the conditional-level MADs: 
$\operatorname{MAR}(\tilde{t}|\bm x) = \operatorname{MAD}(\tilde{t}|\bm x),~
\operatorname{MAR}^+(\tilde{t}|\bm x) = \operatorname{MAD}^+(\tilde{t}|\bm x)$ and
$\operatorname{MAR}^-(\tilde{t}|\bm x) = \operatorname{MAD}^-(\tilde{t}|\bm x)$. 

Therefore, the theoretical results in Theorem~\ref{th:sc} and Proposition~\ref{prop-sc} apply directly to conditional MARs and serve as the theoretical grounding for our UQ framework based on split-point self-consistency verification. 
Moreover, when a calibration set is available, we apply the self-consistency identities in Proposition~\ref{prop-sc-C} to recalibrate softmax outputs, enhancing the reliability of aleatoric uncertainty estimates in classification.

\section{UQ Network Architecture and Learning}
\label{sect:net-learning}

In this section, we present our UQ network architecture for learning the quantities required for UQ through split-point self-consistency verification, as detailed in Section~\ref{sect:method}, along with its training procedure.

As illustrated in Fig. ~\ref{fig: Model_structure}, working in a post-hoc 
manner\footnote{\label{fn:model}If a base model does not exist, our method allows training the base model and the UQ network stagewise or jointly.}, our UQ network builds upon an established base model,
$
f(\bm{x};\Theta^*) = g\bigl(h(\bm{x};\Theta^*)\bigr),
$
where $h(\bm{x};\Theta^*)$ denotes the last hidden layer features or the penultimate layer output, and $g(\cdot)$ is the output layer’s transfer activation. Hereafter, we denote $\bm h = h(\bm{x};\Theta^*)$ as the feature map of $\bm x$ extracted by $h(\bm{x};\Theta^*)$.
Notably, for any input $\bm x$, our UQ network requires only the feature map $\bm h$ and the base model output $\tilde{\bm y}=f(\bm{x};\Theta^*)$ to perform the SPA described in Section~\ref{subsect:spa}.

\subsection{Architecture and Learning for Regression}
\label{subsect:net-learning-regress}

Based on a training dataset $\mathcal{D} = \{(\bm{x}_i, y_i)\}_{i=1}^{|\mathcal{D}|}$\footnote{\label{fn:data}The dataset may differ from the base model’s training set, provided it follows the same distribution. Here, we present the method for univariate regression; its extension to multivariate regression is straightforward.}, our UQ network operates on a regression base model to perform SPA-based quantile regression for aleatoric UQ and to estimate three MARs described in Section~\ref{subsubsect:spa-regress} for epistemic UQ.

As shown in the left plot of Fig. ~\ref{fig: Model_structure}, we employ a fully connected MLP regressor $q(\bm h;\Phi)$
with parameters $\Phi$, as our UQ network. We train $q(\bm h;\Phi)$ to learn
\[
Q\colon h(\mathcal{X})\to \mathcal{QR}^+ \times \mathcal{QR}^- \times \mathcal{Z}_{\mathrm{MAR}}
\times \mathcal{Z}_{\mathrm{MAR}^+} \times \mathcal{Z}_{\mathrm{MAR}^-},
\]
where 
$
\mathcal{QR}^+,~
\mathcal{QR}^-,~
\mathcal{Z}_{\mathrm{MAR}} ,~ 
\mathcal{Z}_{\mathrm{MAR}^+}, ~  
\mathcal{Z}_{\mathrm{MAR}^-} \subset \mathbb{R}
$
are, respectively, the spaces of upper-side residuals, lower-side residuals, and the total, upper-side, and lower-side MARs.
Hence, its output is
$
q\bigl(\bm h;\Phi\bigr)
= \bigl({q}^{+}, {q}^{-}, {z}, {z}^{+}, {z}^{-}\bigr).
$
Here, ${q}^{+}$ and ${q}^{-}$ 
are two independent quantile regression (QR) heads corresponding to $\mathcal{R}^+$ and $\mathcal{R}^-$, while ${z}$, ${z}^{+}$, and ${z}^{-}$ 
estimate the total, upper-side, and lower-side MARs defined in \eqref{eq:reg-MARs}.

We construct two training sets for the split-point quantile regression:
\begin{align*}
\mathcal{D}_{\mathrm{QR}}^+ &= \bigl\{ \bigl( \bm h_i, r_{i} \bigr )| r_{i} \in \mathcal{R}^+\bigr\}_{i=1}^{|\mathcal{R}^+|},\\
\mathcal{D}_{\mathrm{QR}}^- &= \bigl\{ \bigl( \bm h_i, -r_i \bigr )| r_{i} \in \mathcal{R}^-\bigr\}_{i=1}^{|\mathcal{R}^-|}.
\end{align*}
Based on these datasets, we train two QR heads to learn
\(
Q_{\tau^+},\,Q_{\tau^-}\colon h(\mathcal{X})\to\mathbb{R},
\)
with $Q_{\tau^+}$ fitted on $\mathcal{D}_{\mathrm{QR}}^+$ and $Q_{\tau^-}$ fitted on $\mathcal{D}_{\mathrm{QR}}^-$, where $\tau^+,\tau^-\in(0,1)$ denote the marginal confidence levels for the upper and lower quantiles, respectively. 

For notational simplicity, we drop the explicit conditioning on \(\tilde{y}_i|\bm x_i\), and write
\(\mathrm{MAR}(\tilde{y}_i|\bm x_i)\), \(\mathrm{MAR}^+(\tilde{y}_i|\bm x_i)\) and \(\mathrm{MAR}^-(\tilde{y}_i|\bm x_i)\)
as the shorthand \(\mathrm{MAR}_i\), \(\mathrm{MAR}_i^+\), \(\mathrm{MAR}_i^-\), respectively.
Furthermore, we construct the training set for estimating the three MARs defined in \eqref{eq:reg-MARs}:
\begin{align*}
\mathcal{D}_{\mathrm{MAR}} 
&= \bigl\{ \bigl(\bm h_i, \mathrm{MAR}_i\bigr)\bigr\}_{i=1}^{|\mathcal{D}|},\\
\mathcal{D}_{\mathrm{MAR}}^+ 
&= \bigl\{ \bigl(\bm h_i, \mathrm{MAR}_i^{+}\bigr)\bigr\}_{i=1}^{|\mathcal{D}|},\\
\mathcal{D}_{\mathrm{MAR}}^-
&= \bigl\{ \bigl(\bm h_i, \mathrm{MAR}_i^{-}\bigr)\bigr\}_{i=1}^{|\mathcal{D}|}.
\end{align*}
Using these datasets, we train three MAR heads to learn
$Q_{\mathrm{MAR}}\colon h(\mathcal{X})\to \mathcal{Z}_{\mathrm{MAR}}$.
During training, we use the calibration-aware loss \cite{chung2021beyond}, 
$L_{\mathrm{QR}}(\mathcal{D}_{\mathrm{QR}}^+, \mathcal{D}_{\mathrm{QR}}^-, \tau^+, \tau^-; \Phi)$, for quantile regression, and the mean square error (MSE) loss $L_{\mathrm{MSE}}(\mathcal{D}_{\mathrm{MAR}}, \mathcal{D}_{\mathrm{MAR}}^+, \mathcal{D}_{\mathrm{MAR}}^-; \Phi)$ for MAR prediction. Hence, the optimal shared parameters are obtained by
\begin{align*}
\Phi^* &= \arg\min_{\Phi}\Bigl[\,L_{\mathrm{QR}}\bigl(\mathcal{D}_{\mathrm{QR}}^+,\mathcal{D}_{\mathrm{QR}}^-,\tau^+, \tau^-;\Phi\bigr)\\
&\quad \quad \quad \quad \quad +L_{\mathrm{MSE}}\bigl(\mathcal{D}_{\mathrm{MAR}},\mathcal{D}_{\mathrm{MAR}}^+,\mathcal{D}_{\mathrm{MAR}}^-;\Phi\bigr)\Bigr].
\end{align*}

\subsection{Architecture and Learning for Classification}
\label{subsect:net-learning-class}

Based on a training dataset $\mathcal{D} = \{(\bm{x}_i, \bm y_i)\}_{i=1}^{|\mathcal{D}|}$, our UQ network operates on a classification base model to estimate only the total MAR described in \eqref{eq:class_k-MARs-total}, as $\mathrm{MAR}^{+}$ and $\mathrm{MAR}^{-}$ in \eqref{eq:class_k-MARs-upper} and \eqref{eq:class_k-MARs-lower} are derived directly from the base model's softmax output. As shown in the right plot of Fig. ~\ref{fig: Model_structure}, we employ a fully connected MLP regressor $q_{\mathrm{T}}(\bm h, \Phi_{\mathrm{T}})$ with parameters $\Phi_{\mathrm{T}}$ to learn
$$
Q_{\mathrm{T}}\colon h(\mathcal{X}) \to \mathcal{Z}_{\mathrm{MAR}},
$$
where $\mathcal{Z}_{\mathrm{MAR}} \subset \mathbb{R}^K$ is the space of total MAR. 
Hence, its output is the estimated total MAR:
$
q_{\mathrm{T}}\bigl(\bm h;\Phi_{\mathrm{T}}\bigr) = \bm z,
$
the estimated $\mathrm{MAR}$ across the $K$ classes for any input $\bm x$ to the base model via its feature map $\bm h$.
We construct the training set for estimating the total MAR defined in \eqref{eq:class_k-MARs-total}:
$
\mathcal{D}_{\mathrm{MAR}} 
= \bigl\{ \bigl(\bm h_i, (\mathrm{MAR}_{ik})_{k=1}^K\bigr)\bigr\}_{i=1}^{|\mathcal{D}|},
$
where $\mathrm{MAR}_{ik}$ is the shorthand \(\mathrm{MAR}(\tilde{y}_{ik}|\bm x_i)\).
We employ the MSE loss $L_{\mathrm{MSE}}(\mathcal{D}_{\mathrm{MAR}};\Phi_{\mathrm{T}})$ for learning total MAR prediction. Therefore, the optimal parameters are obtained by
$$
\Phi^*_{\mathrm{T}} = \arg\min_{\Phi_{\mathrm{T}}}\Bigl[L_{\mathrm{MSE}}\bigl(\mathcal{D}_{\mathrm{MAR}};\Phi_{\mathrm{T}}\bigr)\Bigr].
$$

When a calibration dataset $\mathcal D_{\mathrm C} = \{(\bm{x}_{{\mathrm C},i}, \bm{y}_{{\mathrm C},i})\}_{i=1}^{|\mathcal D_{\mathrm C}|}$ is available, we learn to estimate the three MARs defined in \eqref{eq:class_k-MARs-c} for calibration. As shown in the right plot of Fig. ~\ref{fig: Model_structure}, we employ another fully connected MLP regressor $q_{\mathrm{C}}(\bm h, \Phi_{\mathrm{C}})$ with parameters $\Phi_{\mathrm{C}}$ to learn
$$
Q_{\mathrm{C}}\colon h(\mathcal{X}) \to \mathcal{Z}_{{\mathrm C}}
\times \mathcal{Z}_{{\mathrm C}^+} \times \mathcal{Z}_{{\mathrm C}^-},
$$
where 
$
\mathcal{Z}_{{\mathrm C}},~ 
\mathcal{Z}_{{\mathrm C}^+},~ 
\mathcal{Z}_{{\mathrm C}^-} \subset \mathbb{R}^K
$
are the spaces of total, upper-side, and lower-side zero-included MARs. 
Hence, its output is
$
q_{\mathrm{C}}\bigl(\bm h;\Phi_{\mathrm{C}}\bigr)
= \bigl({\bm z}_{\mathrm{C}},{\bm z}_{\mathrm{C}}^{+}, {\bm z}_{\mathrm{C}}^{-}\bigr).
$
Here, ${\bm z}_{\mathrm{C}}$, ${\bm z}_{\mathrm{C}}^{+}$, and ${\bm z}_{\mathrm{C}}^{-}$ 
estimate the total, upper-side, and lower-side zero-included MARs defined in \eqref{eq:class_k-MARs-c} for all $K$ classes.
We construct the training set for estimating these zero-included MARs:
\begin{align*}
\mathcal{D}_{\mathrm{MAR}_{\mathrm C}}
&= \bigl\{ \bigl(\bm h_{\mathrm{C},i}, (\mathrm{MAR}_{{\mathrm C},ik})_{k=1}^K\bigr)\bigr\}_{i=1}^{|\mathcal D_{\mathrm C}|},\\
\mathcal{D}_{\mathrm{MAR}_{\mathrm C}^+} 
&= \bigl\{ \bigl(\bm h_{\mathrm{C},i}, (\mathrm{MAR}_{{\mathrm C},ik}^+)_{k=1}^K\bigr)\bigr\}_{i=1}^{|\mathcal D_{\mathrm C}|},\\
\mathcal{D}_{\mathrm{MAR}_{\mathrm C}^-} &= \bigl\{ \bigl(\bm h_{\mathrm{C},i}, (\mathrm{MAR}_{{\mathrm C},ik}^-)_{k=1}^K\bigr)\bigr\}_{i=1}^{|\mathcal D_{\mathrm C}|}.
\end{align*}
Here, $\mathrm{MAR}_{{\mathrm C},ik}, \mathrm{MAR}_{{\mathrm C},ik}^+,
\mathrm{MAR}_{{\mathrm C},ik}^-$ are the shorthand 
$\mathrm{MAR}_{\mathrm C}(\tilde y_{{\mathrm C},ik}|\bm x_i),
\mathrm{MAR}_{\mathrm C}^+(\tilde y_{{\mathrm C},ik}|\bm x_i),
\mathrm{MAR}_{\mathrm C}^-(\tilde y_{{\mathrm C},ik}|\bm x_i)$.
We use the MSE loss 
$L_{\mathrm{MSE}}(\mathcal{D}_{\mathrm{MAR}_{\mathrm C}}, \mathcal{D}_{\mathrm{MAR}_{\mathrm C}^+},\mathcal{D}_{\mathrm{MAR}_{\mathrm C}^-};\Phi_{\mathrm{C}})$ 
for learning the prediction of three MARs used in calibration. Hence, the optimal shared parameters are obtained by
$$
\Phi^*_{\mathrm{C}} = \arg\min_{\Phi_{\mathrm{C}}}\Bigl[L_{\mathrm{MSE}}(\mathcal{D}_{\mathrm{MAR}_{\mathrm C}}, \mathcal{D}_{\mathrm{MAR}_{\mathrm C}^+},\mathcal{D}_{\mathrm{MAR}_{\mathrm C}^-};\Phi_{\mathrm{C}})\Bigr].
$$

All the loss function definitions, the pseudo-code of the learning algorithms and a computational complexity analysis are provided in Appendix B.

\section{UQ Network Architecture and Learning}
\label{sect:net-learning}

In this section, we present our UQ network architecture for learning the quantities required for UQ through split-point self-consistency verification, as detailed in Section~\ref{sect:method}, along with its training procedure.

As illustrated in Fig. ~\ref{fig: Model_structure}, working in a post-hoc 
manner\footnote{\label{fn:model}If a base model does not exist, our method allows training the base model and the UQ network stagewise or jointly.}, our UQ network builds upon an established base model,
$
f(\bm{x};\Theta^*) = g\bigl(h(\bm{x};\Theta^*)\bigr),
$
where $h(\bm{x};\Theta^*)$ denotes the last hidden layer features or the penultimate layer output, and $g(\cdot)$ is the output layer’s transfer activation. Hereafter, we denote $\bm h = h(\bm{x};\Theta^*)$ as the feature map of $\bm x$ extracted by $h(\bm{x};\Theta^*)$.
Notably, for any input $\bm x$, our UQ network requires only the feature map $\bm h$ and the base model output $\tilde{\bm y}=f(\bm{x};\Theta^*)$ to perform the SPA described in Section~\ref{subsect:spa}.

\subsection{Architecture and Learning for Regression}
\label{subsect:net-learning-regress}

Based on a training dataset $\mathcal{D} = \{(\bm{x}_i, y_i)\}_{i=1}^{|\mathcal{D}|}$\footnote{\label{fn:data}The dataset may differ from the base model’s training set, provided it follows the same distribution. Here, we present the method for univariate regression; its extension to multivariate regression is straightforward.}, our UQ network operates on a regression base model to perform SPA-based quantile regression for aleatoric UQ and to estimate three MARs described in Section~\ref{subsubsect:spa-regress} for epistemic UQ.

As shown in the left plot of Fig. ~\ref{fig: Model_structure}, we employ a fully connected MLP regressor $q(\bm h;\Phi)$
with parameters $\Phi$, as our UQ network. We train $q(\bm h;\Phi)$ to learn
\[
Q\colon h(\mathcal{X})\to \mathcal{QR}^+ \times \mathcal{QR}^- \times \mathcal{Z}_{\mathrm{MAR}}
\times \mathcal{Z}_{\mathrm{MAR}^+} \times \mathcal{Z}_{\mathrm{MAR}^-},
\]
where 
$
\mathcal{QR}^+,~
\mathcal{QR}^-,~
\mathcal{Z}_{\mathrm{MAR}} ,~ 
\mathcal{Z}_{\mathrm{MAR}^+}, ~  
\mathcal{Z}_{\mathrm{MAR}^-} \subset \mathbb{R}
$
are, respectively, the spaces of upper-side residuals, lower-side residuals, and the total, upper-side, and lower-side MARs.
Hence, its output is
$
q\bigl(\bm h;\Phi\bigr)
= \bigl({q}^{+}, {q}^{-}, {z}, {z}^{+}, {z}^{-}\bigr).
$
Here, ${q}^{+}$ and ${q}^{-}$ 
are two independent quantile regression (QR) heads corresponding to $\mathcal{R}^+$ and $\mathcal{R}^-$, while ${z}$, ${z}^{+}$, and ${z}^{-}$ 
estimate the total, upper-side, and lower-side MARs defined in \eqref{eq:reg-MARs}.

We construct two training sets for the split-point quantile regression:
\begin{align*}
\mathcal{D}_{\mathrm{QR}}^+ &= \bigl\{ \bigl( \bm h_i, r_{i} \bigr )| r_{i} \in \mathcal{R}^+\bigr\}_{i=1}^{|\mathcal{R}^+|},\\
\mathcal{D}_{\mathrm{QR}}^- &= \bigl\{ \bigl( \bm h_i, -r_i \bigr )| r_{i} \in \mathcal{R}^-\bigr\}_{i=1}^{|\mathcal{R}^-|}.
\end{align*}
Based on these datasets, we train two QR heads to learn
\(
Q_{\tau^+},\,Q_{\tau^-}\colon h(\mathcal{X})\to\mathbb{R},
\)
with $Q_{\tau^+}$ fitted on $\mathcal{D}_{\mathrm{QR}}^+$ and $Q_{\tau^-}$ fitted on $\mathcal{D}_{\mathrm{QR}}^-$, where $\tau^+,\tau^-\in(0,1)$ denote the marginal confidence levels for the upper and lower quantiles, respectively. 

For notational simplicity, we drop the explicit conditioning on \(\tilde{y}_i|\bm x_i\), and write
\(\mathrm{MAR}(\tilde{y}_i|\bm x_i)\), \(\mathrm{MAR}^+(\tilde{y}_i|\bm x_i)\) and \(\mathrm{MAR}^-(\tilde{y}_i|\bm x_i)\)
as the shorthand \(\mathrm{MAR}_i\), \(\mathrm{MAR}_i^+\), \(\mathrm{MAR}_i^-\), respectively.
Furthermore, we construct the training set for estimating the three MARs defined in \eqref{eq:reg-MARs}:
\begin{align*}
\mathcal{D}_{\mathrm{MAR}} 
&= \bigl\{ \bigl(\bm h_i, \mathrm{MAR}_i\bigr)\bigr\}_{i=1}^{|\mathcal{D}|},\\
\mathcal{D}_{\mathrm{MAR}}^+ 
&= \bigl\{ \bigl(\bm h_i, \mathrm{MAR}_i^{+}\bigr)\bigr\}_{i=1}^{|\mathcal{D}|},\\
\mathcal{D}_{\mathrm{MAR}}^-
&= \bigl\{ \bigl(\bm h_i, \mathrm{MAR}_i^{-}\bigr)\bigr\}_{i=1}^{|\mathcal{D}|}.
\end{align*}
Using these datasets, we train three MAR heads to learn
$Q_{\mathrm{MAR}}\colon h(\mathcal{X})\to \mathcal{Z}_{\mathrm{MAR}}$.
During training, we use the calibration-aware loss \cite{chung2021beyond}, 
$L_{\mathrm{QR}}(\mathcal{D}_{\mathrm{QR}}^+, \mathcal{D}_{\mathrm{QR}}^-, \tau^+, \tau^-; \Phi)$, for quantile regression, and the mean square error (MSE) loss $L_{\mathrm{MSE}}(\mathcal{D}_{\mathrm{MAR}}, \mathcal{D}_{\mathrm{MAR}}^+, \mathcal{D}_{\mathrm{MAR}}^-; \Phi)$ for MAR prediction. Hence, the optimal shared parameters are obtained by
\begin{align*}
\Phi^* &= \arg\min_{\Phi}\Bigl[\,L_{\mathrm{QR}}\bigl(\mathcal{D}_{\mathrm{QR}}^+,\mathcal{D}_{\mathrm{QR}}^-,\tau^+, \tau^-;\Phi\bigr)\\
&\quad \quad \quad \quad \quad +L_{\mathrm{MSE}}\bigl(\mathcal{D}_{\mathrm{MAR}},\mathcal{D}_{\mathrm{MAR}}^+,\mathcal{D}_{\mathrm{MAR}}^-;\Phi\bigr)\Bigr].
\end{align*}

\subsection{Architecture and Learning for Classification}
\label{subsect:net-learning-class}

Based on a training dataset $\mathcal{D} = \{(\bm{x}_i, \bm y_i)\}_{i=1}^{|\mathcal{D}|}$, our UQ network operates on a classification base model to estimate only the total MAR described in \eqref{eq:class_k-MARs-total}, as $\mathrm{MAR}^{+}$ and $\mathrm{MAR}^{-}$ in \eqref{eq:class_k-MARs-upper} and \eqref{eq:class_k-MARs-lower} are derived directly from the base model's softmax output. As shown in the right plot of Fig. ~\ref{fig: Model_structure}, we employ a fully connected MLP regressor $q_{\mathrm{T}}(\bm h, \Phi_{\mathrm{T}})$ with parameters $\Phi_{\mathrm{T}}$ to learn
$$
Q_{\mathrm{T}}\colon h(\mathcal{X}) \to \mathcal{Z}_{\mathrm{MAR}},
$$
where $\mathcal{Z}_{\mathrm{MAR}} \subset \mathbb{R}^K$ is the space of total MAR. 
Hence, its output is the estimated total MAR:
$
q_{\mathrm{T}}\bigl(\bm h;\Phi_{\mathrm{T}}\bigr) = \bm z,
$
the estimated $\mathrm{MAR}$ across the $K$ classes for any input $\bm x$ to the base model via its feature map $\bm h$.
We construct the training set for estimating the total MAR defined in \eqref{eq:class_k-MARs-total}:
$
\mathcal{D}_{\mathrm{MAR}} 
= \bigl\{ \bigl(\bm h_i, (\mathrm{MAR}_{ik})_{k=1}^K\bigr)\bigr\}_{i=1}^{|\mathcal{D}|},
$
where $\mathrm{MAR}_{ik}$ is the shorthand \(\mathrm{MAR}(\tilde{y}_{ik}|\bm x_i)\).
We employ the MSE loss $L_{\mathrm{MSE}}(\mathcal{D}_{\mathrm{MAR}};\Phi_{\mathrm{T}})$ for learning total MAR prediction. Therefore, the optimal parameters are obtained by
$$
\Phi^*_{\mathrm{T}} = \arg\min_{\Phi_{\mathrm{T}}}\Bigl[L_{\mathrm{MSE}}\bigl(\mathcal{D}_{\mathrm{MAR}};\Phi_{\mathrm{T}}\bigr)\Bigr].
$$

When a calibration dataset $\mathcal D_{\mathrm C} = \{(\bm{x}_{{\mathrm C},i}, \bm{y}_{{\mathrm C},i})\}_{i=1}^{|\mathcal D_{\mathrm C}|}$ is available, we learn to estimate the three MARs defined in \eqref{eq:class_k-MARs-c} for calibration. As shown in the right plot of Fig. ~\ref{fig: Model_structure}, we employ another fully connected MLP regressor $q_{\mathrm{C}}(\bm h, \Phi_{\mathrm{C}})$ with parameters $\Phi_{\mathrm{C}}$ to learn
$$
Q_{\mathrm{C}}\colon h(\mathcal{X}) \to \mathcal{Z}_{{\mathrm C}}
\times \mathcal{Z}_{{\mathrm C}^+} \times \mathcal{Z}_{{\mathrm C}^-},
$$
where 
$
\mathcal{Z}_{{\mathrm C}},~ 
\mathcal{Z}_{{\mathrm C}^+},~ 
\mathcal{Z}_{{\mathrm C}^-} \subset \mathbb{R}^K
$
are the spaces of total, upper-side, and lower-side zero-included MARs. 
Hence, its output is
$
q_{\mathrm{C}}\bigl(\bm h;\Phi_{\mathrm{C}}\bigr)
= \bigl({\bm z}_{\mathrm{C}},{\bm z}_{\mathrm{C}}^{+}, {\bm z}_{\mathrm{C}}^{-}\bigr).
$
Here, ${\bm z}_{\mathrm{C}}$, ${\bm z}_{\mathrm{C}}^{+}$, and ${\bm z}_{\mathrm{C}}^{-}$ 
estimate the total, upper-side, and lower-side zero-included MARs defined in \eqref{eq:class_k-MARs-c} for all $K$ classes.
We construct the training set for estimating these zero-included MARs:
\begin{align*}
\mathcal{D}_{\mathrm{MAR}_{\mathrm C}}
&= \bigl\{ \bigl(\bm h_{\mathrm{C},i}, (\mathrm{MAR}_{{\mathrm C},ik})_{k=1}^K\bigr)\bigr\}_{i=1}^{|\mathcal D_{\mathrm C}|},\\
\mathcal{D}_{\mathrm{MAR}_{\mathrm C}^+} 
&= \bigl\{ \bigl(\bm h_{\mathrm{C},i}, (\mathrm{MAR}_{{\mathrm C},ik}^+)_{k=1}^K\bigr)\bigr\}_{i=1}^{|\mathcal D_{\mathrm C}|},\\
\mathcal{D}_{\mathrm{MAR}_{\mathrm C}^-} &= \bigl\{ \bigl(\bm h_{\mathrm{C},i}, (\mathrm{MAR}_{{\mathrm C},ik}^-)_{k=1}^K\bigr)\bigr\}_{i=1}^{|\mathcal D_{\mathrm C}|}.
\end{align*}
Here, $\mathrm{MAR}_{{\mathrm C},ik}, \mathrm{MAR}_{{\mathrm C},ik}^+,
\mathrm{MAR}_{{\mathrm C},ik}^-$ are the shorthand 
$\mathrm{MAR}_{\mathrm C}(\tilde{y}_{ik}|\bm x_i),
\mathrm{MAR}_{\mathrm C}^+(\tilde{y}_{ik}|\bm x_i),
\mathrm{MAR}_{\mathrm C}^-(\tilde{y}_{ik}|\bm x_i)$.
We use the MSE loss 
$L_{\mathrm{MSE}}(\mathcal{D}_{\mathrm{MAR}_{\mathrm C}}, \mathcal{D}_{\mathrm{MAR}_{\mathrm C}^+},\mathcal{D}_{\mathrm{MAR}_{\mathrm C}^-};\Phi_{\mathrm{C}})$ 
for learning the prediction of three MARs used in calibration. Hence, the optimal shared parameters are obtained by
$$
\Phi^*_{\mathrm{C}} = \arg\min_{\Phi_{\mathrm{C}}}\Bigl[L_{\mathrm{MSE}}(\mathcal{D}_{\mathrm{MAR}_{\mathrm C}}, \mathcal{D}_{\mathrm{MAR}_{\mathrm C}^+},\mathcal{D}_{\mathrm{MAR}_{\mathrm C}^-};\Phi_{\mathrm{C}})\Bigr].
$$

All the loss function definitions, the pseudo-code of the learning algorithms and a computational complexity analysis are provided in Appendix B.

\section{Uncertainty Quantification via Split-Point Self-Consistency Verification}
\label{sect:method}

In this section, we introduce our UQ method, guided by the self-consistency principle (Section~\ref{subsect:self-consistency-principle}) and supported by trained UQ networks (Section~\ref{sect:net-learning}).

As shown in Fig. ~\ref{fig: Model_structure}, for each test point \(\hat{\bm x}\in\hat{\mathcal D}=\{\hat{\bm x}_i\}_{i=1}^{|\hat{\mathcal D}|}\), the pretrained base model and the trained UQ networks supply all quantities required by our UQ method. Specifically, the base model produces the feature map \(\hat{\bm h}=h(\hat{\bm x};\Theta^*)\) and prediction \(\hat{\bm y}=f(\hat{\bm x};\Theta^*)\). In regression, the UQ network \(q(\hat{\bm h};\Phi^*)\) outputs \((\hat{q}^+,\hat{q}^-,\hat{z},\hat{z}^+,\hat{z}^-)\). In classification, \(q_{\mathrm{T}}(\hat{\bm h};\Phi_{\mathrm{T}}^*)\) yields \(\hat{\bm z}\), while \(q_{\mathrm{C}}(\hat{\bm h};\Phi_{\mathrm{C}}^*)\) yields \((\hat{\bm z}_{\mathrm{C}},\hat{\bm z}_{\mathrm{C}}^+,\hat{\bm z}_{\mathrm{C}}^-)\).

\subsection{Quantifying Epistemic Uncertainty}
\label{subsect:epistemic_uq}

\subsubsection{Self-Consistency Discrepancy Score}
\label{subsect:SDS}

To quantify epistemic uncertainty, we adapt Proposition~\ref{prop-sc} to define the \emph{self-consistency discrepancy score} (SDS):
\begin{equation}
\label{eq:sds}
\Delta^\prime
= \bigl|2\,\mathrm{MAR}^+\,\mathrm{MAR}^- - \mathrm{MAR}\,(\mathrm{MAR}^+ + \mathrm{MAR}^-)\bigr|.
\end{equation}
This formulation avoids the division in the harmonic mean in \eqref{eq:sc}, thereby reducing numerical instability and extreme values. It preserves the core self-consistency discrepancy and yields more robust estimates of epistemic uncertainty.

The SDS at a test point \(\hat{\bm{x}}\) reflects two factors: (i) the bias of the base model’s prediction \(\hat{\bm{y}}\) relative to the true conditional expectation \(\mathbb{E} \big [\bm{y}| \hat{\bm{x}} \big ]\), and (ii) the model’s lack of knowledge in the neighbourhood of \(\hat{\bm{x}}\), which induces stochastic deviations that break the consistency among MAR estimates. Thus, \(\mathrm{SDS}\) captures both predictive bias and distributional mismatch, enabling fine‑grained quantification of epistemic uncertainty.

In regression, the UQ network \(q(\hat{\bm h};\Phi^*)\) provides the three MAR estimates \(\hat z\), \(\hat z^+\) and \(\hat z^-\). According to \eqref{eq:sds}, for a test prediction \(\hat{y}\) on \(\hat{\bm x}\),  the SDS is simply
\begin{equation}
\label{eq:sds-regress}
\Delta^\prime(\hat{y}|\hat{\bm x}) 
= \bigl|2\hat z^+\hat z^- - \hat z(\hat z^+ + \hat z^-)\bigr|.
\end{equation}

In classification, we directly obtain the upper- and lower-side MARs defined in \eqref{eq:class_k-MARs-upper} and \eqref{eq:class_k-MARs-lower} from the base model, as
\[
\hat{\bm z}^+ = \bm{1} - \hat{\bm y},
\quad
\hat{\bm z}^- = \hat{\bm y}.
\]
Here, \(\bm{1}\) is the all-ones vector. 
Together with the total MAR \(\hat{\bm z}\) estimated by the UQ network \(q_{\mathrm{T}}(\hat{\bm h};\Phi_{\mathrm{T}}^*)\), according to \eqref{eq:sds}, for a test prediction \(\hat{\bm y}\) on \(\hat{\bm x}\), the SDS becomes
\begin{equation}
\label{eq:sds-class}
\Delta^\prime(\hat{\bm y}|\hat{\bm x})
= \lVert
2\hat{\bm z}^+\odot\hat{\bm z}^- 
- \hat{\bm z}\odot(\hat{\bm z}^+ + \hat{\bm z}^-)
\bigr\rVert_{1},
\end{equation}
where $\lVert\cdot\rVert_{1}$ is the $\mathfrak{•}{l}_1$ norm and $\odot$ is the operator for element-wise multiplication.

\subsubsection{Out-of-Distribution Detection}
\label{subsect:OOD}

While the $\mathrm{SDS}$ serves as a \textit{fine-grained} metric for quantifying epistemic uncertainty on iD data, it can also detect OOD points and other high-uncertainty cases. In practice, we first compute $\mathrm{SDS}$ values on a held‑out iD validation set to build a reference distribution (e.g., via histogram) and choose a threshold $\Delta^\prime_0$ as its upper $\alpha$‑quantile (e.g., 95th percentile). At inference, for any test input $\hat{\bm x}$, we compute
$
\Delta^\prime(\hat{\bm y}|\hat{\bm x})
$
and flag the point as OOD if
$
\Delta^\prime(\hat{\bm y}|\hat{\bm x}) > \Delta^\prime_0.
$

\subsection{Quantifying Aleatoric Uncertainty}
\label{subsect:aleatoric_uq}

\subsubsection{Aleatoric Uncertainty in Regression}
\label{subsect:aleatoric-regress}

The UQ network for regression, \(q(\hat{\bm h};\Phi^*)\) with \(\hat{\bm h}=h(\hat{\bm x};\Theta^*)\), has two QR heads that produce the estimates \(\hat q^+\) and \(\hat q^-\) corresponding to marginal confidence levels \(\tau^+\) and \(\tau^-\). For a test point \(\hat{\bm x}\in\hat{\mathcal D}\) with prediction \(\hat{y}=f(\hat{\bm x};\Theta^*)\), we define the \emph{split-point prediction interval} (SPI) as
\begin{equation}
\label{eq:spi}
y \;\in\; \bigl[\hat{y}-\hat q^-,\;\hat{y}+\hat q^+\bigr].
\end{equation}
By separating over- and under-estimation residuals and constructing bounds for each side, the SPI yields a more informative quantification of aleatoric uncertainty.

Nevertheless, the estimates $\hat q^+$ and $\hat q^-$ may be noisy, especially when $\Delta^\prime(\hat{y}|\hat{\bm x})$ in \eqref{eq:sds-regress} is large, rendering the SPI unreliable. To improve robustness, we enforce the self-consistency constraint from Theorem~\ref{th:sc} using the shared UQ network of the two QR heads and three MAR heads. From \eqref{eq:sc}, the MAR outputs must satisfy the harmonic relation, which implies
$$
\hat z^+_{\mathrm{C}}
= \frac{\hat z\,\hat z^-}{2\,\hat z^- - \hat z},
\quad
\hat z^-_{\mathrm{C}}
= \frac{\hat z\,\hat z^+}{2\,\hat z^+ - \hat z}.
$$
To enhance reliability in regions prone to under-coverage, we define the \textit{calibration factors} as follows:
$$
s^+_{\mathrm{C}} = \frac{\max\bigl(\hat z^+,\,\hat z^+_{\mathrm{C}}\bigr)}{\hat z^+},
\quad
s^-_{\mathrm{C}} = \frac{\max\bigl(\hat z^-,\,\hat z^-_{\mathrm{C}}\bigr)}{\hat z^-}.
$$
Applying these scaling factors yields the \emph{calibrated} SPI:
\begin{equation}
\label{eq:calibrated-SPI}
y \;\in\; \bigl[\hat{y} - s^-_{\mathrm{C}}\,\hat q^-,\;\hat{y} + s^+_{\mathrm{C}}\,\hat q^+\bigr].
\end{equation}

\subsubsection{Aleatoric Uncertainty in Classification}
\label{subsect:aleatoric-class}

For a test input \(\hat{\bm x}\) with softmax output \(\hat{\bm y}\), its aleatoric uncertainty is quantified via the predictive entropy \cite{kendall2017uncertainties}:
\begin{equation}
\label{eq:entroy-predictive}
\Ent(\hat{\bm y}| \hat{\bm x})
= -\sum_{k=1}^K \hat y_k \,\log\bigl(\hat y_k\bigr).
\end{equation}

When calibration data are available, we utilize the zero-included MAR self-consistency encoded in 
\eqref{eq:MAR-C_relation-1} and \eqref{eq:MAR-C_relation-2}
for calibrating a base model's prediction.
Based on the vectorial form of \eqref{eq:MAR-C_relation-1} and the UQ network output 
\(q_{\mathrm{C}}(\hat{\bm h};\Phi_{\mathrm{C}}^*)=\bigl(\hat{\bm z}_{\mathrm{C}},\hat{\bm z}_{\mathrm{C}}^+,\hat{\bm z}_{\mathrm{C}}^-\bigr)
\), we define a \textit{calibration quality-assurance factor} for a test point \(\hat{\bm x}\) with prediction \(\hat{\bm y}\):
$$
\delta_{\mathrm{C}}(\hat{\bm y}|\hat{\bm x})
= \lVert\,\hat{\bm z}_{\mathrm{C}} - \hat{\bm z}_{\mathrm{C}}^+ - \hat{\bm z}_{\mathrm{C}}^-\,\rVert_{1}.
$$
A small \(\delta_{\mathrm{C}}(\hat{\bm y}|\hat{\bm x})\) indicates reliable calibration, so we adjust \(\hat{\bm y}\) only if \(\delta_{\mathrm{C}}(\hat{\bm y}|\hat{\bm x})<\delta_{0}\), where \(\delta_{0}\) is chosen via cross-validation. Applying the vectorial form of \eqref{eq:MAR-C_relation-2}, the calibrated prediction $\hat{\bm y}_{\mathrm{C}}$ becomes
\begin{equation}
\label{eq:calibrated-predic}
\hat{\bm y}_{\mathrm{C}}
= \hat{\bm y} + 
\mathbbm{1}\bigl(\delta_{\mathrm{C}}(\hat{\bm y}|\hat{\bm x})<\delta_{0}\bigr)(\hat{\bm z}_{\mathrm{C}}^+ - \hat{\bm z}_{\mathrm{C}}^-),
\end{equation}
where $\mathbbm{1}(\cdot)$ is the indicator function. 
In practice, we find that setting \(\delta_{0}=0.01\) yields stable and reliable results across all experimental datasets.

\section{Experiments}
\label{sect:experiment}

In this section, we describe our experimental setup, report results for the regression and classification tasks, respectively, and summarize extended experimental findings in the supplementary materials.

\subsection{Experimental Setup}
\label{subsect:setup}

Below, we outline our experimental setup, and all the details are provided in Appendix~C to ensure reproducibility.

\subsubsection{Datasets}
\label{subsubsect:dataset}
For regression, we first consider a synthetic cubic regression task \cite{hernandez2015probabilistic, Lakshminarayanan2016SimpleAS, amini2020deep},
\[
y = x^3 + \epsilon(x) - \mathbb{E}[\epsilon(x)], \quad \epsilon(x) \sim \mathrm{LogNormal}(1.5, 1),
\]
which enables an illustrative study of asymmetric noise. Next, we use nine UCI regression benchmarks \cite{UCI2025}, widely used for UQ evaluation. 
Finally, we evaluate on a high-dimensional monocular depth estimation dataset \cite{silberman2012indoor, huang2018apolloscape} to assess performance in complex real-world scenarios. To test fine-grained uncertainty estimation, we generate adversarial variants using the Fast Gradient Sign Method (FGSM)~\cite{goodfellow2014explaining}, where the perturbation magnitude is controlled by a parameter \(\epsilon\).

For classification, we use CIFAR-10, CIFAR-100 \cite{Krizhevsky2009}, and ImageNet-1K \cite{Deng2009} as iD datasets, augmenting them with FGSM adversarial variants. For OOD detection, we evaluate on SVHN \cite{Netzer2011}, Tiny ImageNet \cite{le2015tiny}, and ImageNet-O/A \cite{Hendrycks2019}. Finally, we assess performance in real-world multimodal scenarios using the LUMA benchmark \cite{luma_dataset2025}.

\begin{figure*}[thbp]
  \centering
  \includegraphics[width=1\linewidth]{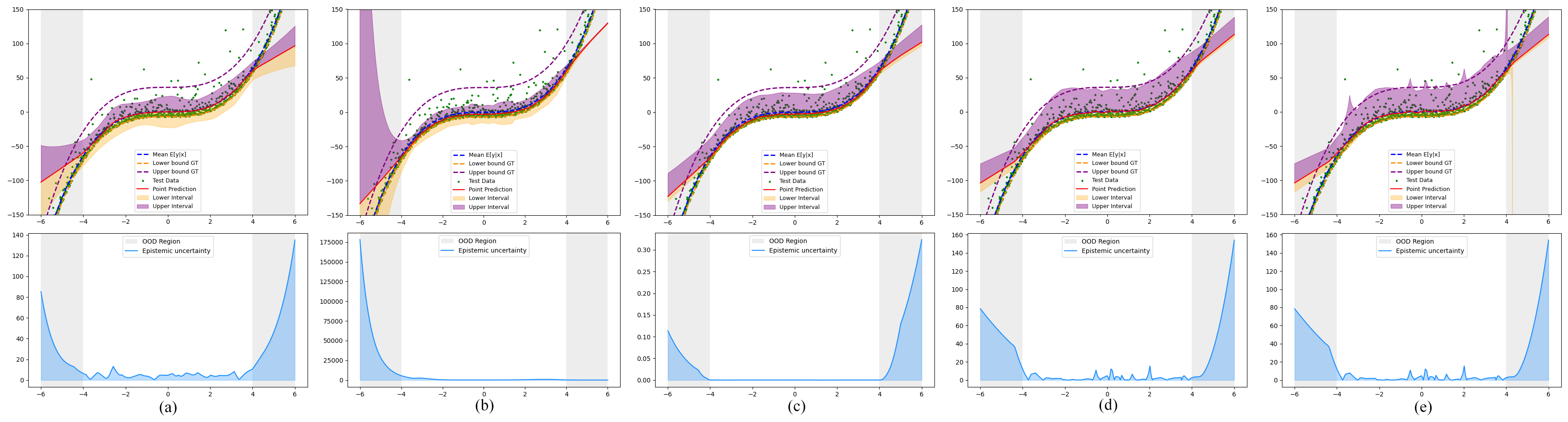}
  \caption{Uncertainties quantified for the cubic regression using (a) Deep Ensemble (DE), (b) Evidential Regression (EDL-R), (c) SQR-OC, (d) our method without calibration, and (e) our method with calibration. \textbf{Top row} shows aleatoric uncertainty estimates, \textbf{bottom row} shows epistemic uncertainty estimates. Ground truth and true PI boundaries are shown as dashed lines.
}
  \label{fig:unc-decomposition}
\vspace*{-3mm}
\end{figure*}

\subsubsection{Baselines and Base Models} 
\label{subsubsect:base}

In our comparative study, we adopt several state-of-the-art baselines within single-forward-pass, including internal and external methods, and multi-forward-pass categories (see Section \ref{sect:related-work}):  
(i) Bayesian-based methods, \textit{MC-Dropout} (MD)~\cite{gal2016dropout} and \textit{Laplace Approximation} (LA)~\cite{daxberger2021laplace};  
(ii) ensemble-based method, \textit{Deep Ensemble} (DE)~\cite{Lakshminarayanan2016SimpleAS};  
(iii) internal (evidential) methods, \textit{Evidential Regression} (EDL-R)~\cite{amini2020deep}, \textit{Evidential Quantile Regression} (EDL-QR)~\cite{huttel2023deep}, and \textit{Evidential Classification} (EDL-C)~\cite{sensoy2018evidential};  
(iv) external methods, \textit{SQR-OC}~\cite{tagasovska2019single} and \textit{DDU}~\cite{mukhoti2023deep}.

As reviewed in Section \ref{sect:related-work}, external methods operate on a base model. For regression, we train fully connected MLPs as base models on the synthetic cubic regression task and the nine UCI benchmarks. For monocular depth estimation, we train a U-Net~\cite{ronneberger2015u} as the base model. In image classification, we employ two CNN architectures, VGG-16~\cite{Simonyan2015} and Wide ResNet~\cite{zagoruyko2016wide}, as base models. For CIFAR-10 and CIFAR-100, both networks are trained from scratch, whereas for ImageNet-1K we use pretrained models~\cite{torchvision2024models}. Finally, as the base model for the multimodal task, following \cite{luma_dataset2025}, we train a CNN to encode the visual modality and Transformer encoders to process the audio and text modalities.

\subsubsection{Experimental Protocol} 
\label{subsubsect:protocal}

We evaluate all models under identical settings, including the same training, validation and test splits, and a consistent hyperparameter search. For regression, MD and DE use Gaussian-likelihood regression, while other methods rely on their own evidential or quantile-based distributions to construct 95\% prediction intervals (PIs). Point predictions are defined as the predictive mean in Gaussian-based models, the 50th percentile in quantile regression, and the MSE-optimal output in our framework. For classification, methods without a built-in aleatoric calibration mechanism employ Temperature Scaling (TS)~\cite{guo2017calibration}. For hyperparameter tuning, we randomly reserve 10\% of the training data as a validation set and perform $k$-fold cross validation, with $k=20$ for synthetic, UCI datasets and CIFAR-10/CIFAR-100, and $k = 5$ for the remaining datasets due to computational constraints. To simulate calibration data, we further sample 10\% of the training set without data leakage. All models are evaluated on the predefined test sets.

For our method, we also conduct extended experiments to assess joint training of the base model and UQ network from scratch\textsuperscript{\ref{fn:model}}, evaluate robustness to training data volume\textsuperscript{\ref{fn:data}}, and test our confidence calibration under the same settings.

\subsubsection{Evaluation Criteria} 
\label{subsubsect:evaluation}

We assess UQ performance using the following criteria: 
(i) \textbf{Accuracy}, measured by \textit{root mean squared error} (RMSE) for regression point predictions and by prediction accuracy for classification; 
(ii) \textbf{Aleatoric uncertainty}, quantified by \textit{expected calibration error} (ECE)~\cite{naeini2015obtaining} for classification, and by \textit{prediction interval expected calibration error} (PIECE) \cite{levi2022evaluating} and Winkler score~\cite{winkler1972decision} for regression. 
Motivated by our SPA, we also adopt fine grained split-point metrics \(\text{PIECE}^{+}\) and \(\text{PIECE}^{-}\) on the upper and lower split point intervals, respectively. This decomposition measures overestimation and underestimation separately and applies to any model yielding point predictions;
(iii) \textbf{Epistemic uncertainty}, evaluated via the Spearman \textit{correlation} coefficient~\cite{spearman1961proof} for regression and \textit{area under the receiver operating characteristic curve} (AUROC) for classification, reflecting the separability of adversarial, OOD and error samples; 
(iv) \textbf{Efficiency}, assessed by training and inference time.

\subsubsection{Implementation} 
\label{subsubsect:implement}

We implement our framework in Python using PyTorch and its built-in Adam and SGD optimizers\footnote{Our source code is available at \url{https://github.com/zzz0527/SPC-UQ}.}. Experiments run on an NVIDIA A100 GPU with 16\,GB of memory, while ImageNet-1K experiments use an NVIDIA V100 GPU with 80\,GB of memory. For each baseline, we adapt the original authors' open-source code on the same platform and retain their default hyperparameters.

\begin{table*}
\caption{Results on UCI Regression Benchmarks}
\caption*{\small The best and second-best results per column are indicated by \underline{\textbf{bold underlining}} and \underline{\textit{italic underlining}}. This notation applies to all tables.}
\label{tab:UCI-results1}
\centering
\scalebox{0.8}{
\begin{tabular}{cllllllllll}
\hline
\multicolumn{1}{l}{\multirow{2}{*}{Metric}}                              & \multirow{2}{*}{Method}                                                      & \multicolumn{9}{c}{Dataset}                                                                                                                                                                                                                                             \\ \cline{3-11} 
\multicolumn{1}{l}{}                                                     &                                                                              & Boston                      & Concrete                    & Energy                     & Kin8nm                     & Naval                      & Power                       & Protein                     & Wine                       & Yacht                       \\ \hline
\multirow{6}{*}{RMSE}                                                    & MD                                                                           & 3.99 ± 0.17                 & 7.75 ± 0.13                 & 2.92 ± 0.08                & 0.14 ± 0.00                & \underline{\textbf{0.00 ± 0.00}} & 4.18 ± 0.03                 & 4.78 ± 0.01                 & \underline{\textit{0.64 ± 0.01}} & 5.53 ± 0.30                 \\
                                                                         & DE                                                                           & \underline{\textit{3.70 ± 0.15}}  & \underline{\textbf{6.99 ± 0.13}}  & \underline{\textit{2.70 ± 0.08}} & 0.11 ± 0.00                & \underline{\textbf{0.00 ± 0.00}} & \underline{\textbf{3.97 ± 0.03}}  & \underline{\textit{4.59 ± 0.01}}  & \underline{\textbf{0.63 ± 0.01}} & 6.24 ± 0.48                 \\
                                                                         & EDL-R                                                                        & 3.81 ± 0.17                 & 7.01 ± 0.14                 & 2.78 ± 0.08                & 0.13 ± 0.00                & \underline{\textbf{0.00 ± 0.00}} & 3.98 ± 0.03                 & 4.80 ± 0.01                 & \underline{\textit{0.64 ± 0.01}} & 6.41 ± 0.51                 \\
                                                                         & EDL-QR                                                                       & 4.07 ± 0.17                 & 7.61 ± 0.15                 & 2.86 ± 0.09                & 0.12 ± 0.00                & \underline{\textbf{0.00 ± 0.00}} & 4.02 ± 0.03                 & 4.75 ± 0.01                 & \underline{\textit{0.64 ± 0.01}} & 7.89 ± 0.62                 \\
                                                                         & SQR-OC                                                                       & 3.97 ± 0.17                 & 7.64 ± 0.13                 & 2.80 ± 0.09                & \underline{\textit{0.10 ± 0.00}} & \underline{\textbf{0.00 ± 0.00}} & 4.03 ± 0.03                 & 4.67 ± 0.02                 & 0.65 ± 0.01                & \underline{\textit{4.90 ± 0.34}}  \\
                                                                         & Ours                                                                         & \underline{\textbf{3.69 ± 0.15}}  & \underline{\textit{7.09 ± 0.14}}  & \underline{\textbf{2.49 ± 0.06}} & \underline{\textbf{0.09 ± 0.00}} & \underline{\textbf{0.00 ± 0.00}} & \underline{\textbf{3.97 ± 0.03}}  & \underline{\textbf{4.46 ± 0.01}}  & \underline{\textit{0.64 ± 0.01}} & \underline{\textbf{4.56 ± 0.22}}  \\ \hline
\multirow{7}{*}{\begin{tabular}[c]{@{}c@{}}Winkler\\ Score\end{tabular}} & MD                                                                           & 20.78 ± 1.04                & 35.82 ± 0.71                & 11.57 ± 0.25               & 0.58 ± 0.00                & 0.01 ± 0.00                & 20.66 ± 0.20                & 21.94 ± 0.08                & \underline{\textbf{3.12 ± 0.05}} & 18.08 ± 0.97                \\
                                                                         & DE                                                                           & \underline{\textbf{19.43 ± 1.08}} & \underline{\textbf{32.37 ± 0.72}} & \underline{\textit{8.33 ± 0.17}} & \underline{\textit{0.43 ± 0.00}} & \underline{\textbf{0.00 ± 0.00}} & 19.40 ± 0.23                & 21.23 ± 0.09                & 3.17 ± 0.05                & \underline{\textbf{15.39 ± 1.29}} \\
                                                                         & EDL-R                                                                        & 22.13 ± 1.42                & 33.85 ± 0.89                & 9.58 ± 0.20                & 0.46 ± 0.01                & 0.01 ± 0.00                & 19.70 ± 0.26                & 23.64 ± 0.15                & 3.40 ± 0.05                & 20.54 ± 2.00                \\
                                                                         & EDL-QR                                                                       & 21.35 ± 0.89                & 37.65 ± 0.74                & 9.49 ± 0.24                & 0.50 ± 0.01                & 0.01 ± 0.00                & 19.25 ± 0.24                & 17.94 ± 0.05                & \underline{\textit{3.13 ± 0.04}} & 21.17 ± 1.70                \\
                                                                         & SQR-OC                                                                       & 21.41 ± 0.99                & 39.43 ± 0.85                & 9.56 ± 0.18                & 0.47 ± 0.00                & \underline{\textbf{0.00 ± 0.00}} & \underline{\textit{18.95 ± 0.23}} & \underline{\textit{17.31 ± 0.03}} & 3.22 ± 0.06                & 21.00 ± 1.15                \\
                                                                         & \multirow{2}{*}{\begin{tabular}[c]{@{}l@{}}Ours\\ Ours-Calib\end{tabular}} & 20.33 ± 1.17                & 33.37 ± 0.89                & \underline{\textbf{7.82 ± 0.15}} & \underline{\textbf{0.41 ± 0.01}} & \underline{\textbf{0.00 ± 0.00}} & \underline{\textbf{18.73 ± 0.24}} & \underline{\textbf{16.55 ± 0.04}} & \underline{\textit{3.13 ± 0.06}} & 16.36 ± 1.19                \\
                                                                         &                                                                              & \underline{\textit{19.65 ± 0.98}} & \underline{\textit{32.86 ± 0.81}} & \underline{\textit{8.12 ± 0.14}} & \underline{\textit{0.43 ± 0.01}} & \underline{\textbf{0.00 ± 0.00}} & \underline{\textit{18.75 ± 0.24}} & 17.80 ± 0.10                & 3.32 ± 0.10                & \underline{\textit{15.94 ± 1.61}} \\ \hline
\multirow{7}{*}{PIECE}                                                   & MD                                                                           & \underline{\textit{0.04 ± 0.00}}  & \underline{\textbf{0.05 ± 0.00}}  & 0.06 ± 0.00                & 0.03 ± 0.00                & \underline{\textit{0.04 ± 0.00}} & 0.03 ± 0.00                 & 0.02 ± 0.00                 & \underline{\textit{0.03 ± 0.00}} & \underline{\textbf{0.05 ± 0.00}}  \\
                                                                         & DE                                                                           & 0.05 ± 0.00                 & 0.06 ± 0.00                 & \underline{\textbf{0.05 ± 0.00}} & \underline{\textbf{0.01 ± 0.00}} & 0.05 ± 0.00                & \underline{\textbf{0.02 ± 0.00}}  & 0.02 ± 0.00                 & 0.04 ± 0.00                & 0.08 ± 0.01                 \\
                                                                         & EDL-R                                                                        & 0.06 ± 0.00                 & 0.06 ± 0.00                 & 0.07 ± 0.01                & 0.02 ± 0.00                & 0.05 ± 0.01                & \underline{\textbf{0.02 ± 0.00}}  & \underline{\textbf{0.01 ± 0.00}}  & 0.05 ± 0.00                & 0.11 ± 0.01                 \\
                                                                         & EDL-QR                                                                       & 0.07 ± 0.00                 & 0.06 ± 0.00                 & 0.07 ± 0.00                & 0.03 ± 0.00                & 0.07 ± 0.01                & \underline{\textbf{0.02 ± 0.00}}  & 0.02 ± 0.00                 & \underline{\textit{0.03 ± 0.00}} & 0.13 ± 0.01                 \\
                                                                         & SQR-OC                                                                       & 0.05 ± 0.00                 & 0.06 ± 0.00                 & 0.08 ± 0.00                & 0.02 ± 0.00                & 0.05 ± 0.01                & \underline{\textbf{0.02 ± 0.00}}  & \underline{\textbf{0.01 ± 0.00}}  & 0.04 ± 0.00                & 0.15 ± 0.01                 \\
                                                                         & \multirow{2}{*}{\begin{tabular}[c]{@{}l@{}}Ours\\ Ours-Calib\end{tabular}} & 0.05 ± 0.00                 & 0.06 ± 0.00                 & 0.08 ± 0.01                & 0.02 ± 0.00                & \underline{\textbf{0.02 ± 0.00}} & \underline{\textbf{0.02 ± 0.00}}  & \underline{\textbf{0.01 ± 0.00}}  & \underline{\textit{0.03 ± 0.00}} & 0.14 ± 0.01                 \\
                                                                         &                                                                              & \underline{\textbf{0.03 ± 0.00}}  & \underline{\textbf{0.05 ± 0.00}}  & \underline{\textbf{0.05 ± 0.00}} & \underline{\textbf{0.01 ± 0.00}} & \underline{\textit{0.03 ± 0.00}} & \underline{\textbf{0.02 ± 0.00}}  & 0.02 ± 0.00                 & \underline{\textbf{0.02 ± 0.00}} & \underline{\textit{0.07 ± 0.01}}  \\ \hline
\multirow{7}{*}{$\text{PIECE}^+$}                                        & MD                                                                           & \underline{\textit{0.03 ± 0.01}}  & \underline{\textbf{0.02 ± 0.00}}  & \underline{\textbf{0.03 ± 0.00}} & 0.03 ± 0.00                & 0.04 ± 0.00                & 0.03 ± 0.00                 & 0.02 ± 0.00                 & 0.03 ± 0.00                & \underline{\textit{0.06 ± 0.01}}  \\
                                                                         & DE                                                                           & 0.05 ± 0.01                 & 0.04 ± 0.00                 & \underline{\textbf{0.03 ± 0.01}} & \underline{\textbf{0.01 ± 0.00}} & 0.04 ± 0.00                & \underline{\textbf{0.01 ± 0.00}}  & 0.01 ± 0.00                 & 0.04 ± 0.01                & 0.13 ± 0.02                 \\
                                                                         & EDL-R                                                                        & 0.07 ± 0.01                 & 0.04 ± 0.01                 & 0.05 ± 0.01                & \underline{\textbf{0.01 ± 0.00}} & 0.03 ± 0.00                & \underline{\textbf{0.01 ± 0.00}}  & 0.03 ± 0.00                 & 0.05 ± 0.01                & 0.14 ± 0.02                 \\
                                                                         & EDL-QR                                                                       & 0.04 ± 0.01                 & 0.04 ± 0.00                 & \underline{\textbf{0.03 ± 0.01}} & \underline{\textbf{0.01 ± 0.00}} & 0.05 ± 0.01                & \underline{\textbf{0.01 ± 0.00}}  & \underline{\textbf{0.00 ± 0.00}}  & \underline{\textbf{0.02 ± 0.00}} & 0.07 ± 0.01                 \\
                                                                         & SQR-OC                                                                       & \underline{\textit{0.03 ± 0.00}}  & 0.04 ± 0.01                 & 0.05 ± 0.01                & \underline{\textbf{0.01 ± 0.00}} & \underline{\textit{0.02 ± 0.00}} & \underline{\textbf{0.01 ± 0.00}}  & \underline{\textbf{0.00 ± 0.00}}  & 0.03 ± 0.00                & 0.07 ± 0.01                 \\
                                                                         & \multirow{2}{*}{\begin{tabular}[c]{@{}l@{}}Ours\\ Ours-Calib\end{tabular}} & \underline{\textit{0.03 ± 0.00}}  & 0.04 ± 0.01                 & 0.09 ± 0.01                & 0.02 ± 0.00                & \underline{\textbf{0.01 ± 0.00}} & \underline{\textbf{0.01 ± 0.00}}  & 0.01 ± 0.00                 & 0.03 ± 0.01                & 0.10 ± 0.02                 \\
                                                                         &                                                                              & \underline{\textbf{0.02 ± 0.00}}  & \underline{\textit{0.03 ± 0.01}}  & \underline{\textbf{0.03 ± 0.01}} & \underline{\textbf{0.01 ± 0.00}} & 0.03 ± 0.00                & \underline{\textbf{0.01 ± 0.00}}  & 0.02 ± 0.00                 & \underline{\textbf{0.02 ± 0.00}} & \underline{\textbf{0.05 ± 0.01}}  \\ \hline
\multirow{7}{*}{$\text{PIECE}^-$}                                        & MD                                                                           & 0.03 ± 0.00                 & 0.03 ± 0.00                 & 0.05 ± 0.00                & 0.03 ± 0.00                & \underline{\textit{0.04 ± 0.00}} & 0.02 ± 0.00                 & 0.05 ± 0.00                 & \underline{\textbf{0.01 ± 0.00}} & \underline{\textit{0.05 ± 0.00}}  \\
                                                                         & DE                                                                           & \underline{\textbf{0.02 ± 0.00}}  & \underline{\textbf{0.02 ± 0.00}}  & \underline{\textit{0.03 ± 0.00}} & \underline{\textbf{0.01 ± 0.00}} & 0.05 ± 0.00                & \underline{\textbf{0.01 ± 0.00}}  & 0.04 ± 0.00                 & 0.02 ± 0.00                & \underline{\textbf{0.04 ± 0.00}}  \\
                                                                         & EDL-R                                                                        & \underline{\textbf{0.02 ± 0.00}}  & \underline{\textbf{0.02 ± 0.00}}  & \underline{\textit{0.03 ± 0.00}} & 0.02 ± 0.00                & 0.05 ± 0.02                & \underline{\textbf{0.01 ± 0.00}}  & 0.04 ± 0.00                 & 0.02 ± 0.00                & 0.07 ± 0.01                 \\
                                                                         & EDL-QR                                                                       & 0.03 ± 0.01                 & \underline{\textbf{0.02 ± 0.00}}  & \underline{\textbf{0.02 ± 0.00}} & \underline{\textbf{0.01 ± 0.00}} & 0.05 ± 0.01                & \underline{\textbf{0.01 ± 0.00}}  & 0.03 ± 0.00                 & 0.02 ± 0.00                & 0.09 ± 0.02                 \\
                                                                         & SQR-OC                                                                       & 0.04 ± 0.01                 & 0.03 ± 0.00                 & 0.05 ± 0.01                & \underline{\textbf{0.01 ± 0.00}} & \underline{\textit{0.04 ± 0.01}} & \underline{\textbf{0.01 ± 0.00}}  & \underline{\textbf{0.00 ± 0.00}}  & 0.03 ± 0.01                & 0.09 ± 0.02                 \\
                                                                         & \multirow{2}{*}{\begin{tabular}[c]{@{}l@{}}Ours\\ Ours-Calib\end{tabular}} & 0.03 ± 0.00                 & 0.03 ± 0.01                 & 0.04 ± 0.01                & \underline{\textbf{0.01 ± 0.00}} & \underline{\textbf{0.01 ± 0.00}} & \underline{\textbf{0.01 ± 0.00}}  & \underline{\textit{0.01 ± 0.00}}  & 0.02 ± 0.00                & 0.12 ± 0.02                 \\
                                                                         &                                                                              & \underline{\textbf{0.02 ± 0.00}}  & 0.03 ± 0.00                 & \underline{\textit{0.03 ± 0.01}} & \underline{\textbf{0.01 ± 0.00}} & \underline{\textit{0.03 ± 0.00}} & \underline{\textbf{0.01 ± 0.00}}  & 0.02 ± 0.00                 & \underline{\textbf{0.01 ± 0.00}} & \underline{\textit{0.05 ± 0.01}}  \\ \hline
\multirow{6}{*}{Correlation}                                             & MD                                                                           & \underline{\textbf{0.34 ± 0.02}}  & 0.36 ± 0.02                 & 0.53 ± 0.02                & 0.37 ± 0.01                & 0.45 ± 0.01                & 0.11 ± 0.01                 & 0.44 ± 0.00                 & 0.19 ± 0.01                & 0.70 ± 0.02                 \\
                                                                         & DE                                                                           & \underline{\textit{0.33 ± 0.02}}  & \underline{\textbf{0.42 ± 0.02}}  & \underline{\textbf{0.71 ± 0.01}} & \underline{\textbf{0.44 ± 0.01}} & \underline{\textbf{0.79 ± 0.01}} & 0.18 ± 0.01                 & 0.52 ± 0.00                 & 0.21 ± 0.01                & \underline{\textbf{0.83 ± 0.02}}  \\
                                                                         & EDL-R                                                                        & 0.26 ± 0.02                 & 0.38 ± 0.02                 & \underline{\textbf{0.71 ± 0.02}} & \underline{\textit{0.43 ± 0.01}} & \underline{\textbf{0.79 ± 0.01}} & \underline{\textit{0.21 ± 0.01}}  & -0.19 ± 0.01                & \underline{\textit{0.25 ± 0.01}} & \underline{\textit{0.81 ± 0.02}}  \\
                                                                         & EDL-QR                                                                       & 0.27 ± 0.01                 & 0.35 ± 0.02                 & \underline{\textit{0.61 ± 0.02}} & 0.42 ± 0.01                & \underline{\textit{0.78 ± 0.01}} & 0.19 ± 0.01                 & \underline{\textit{0.53 ± 0.00}}  & \underline{\textbf{0.26 ± 0.01}} & 0.79 ± 0.02                 \\
                                                                         & SQR-OC                                                                       & 0.31 ± 0.02                 & 0.27 ± 0.02                 & 0.44 ± 0.02                & 0.36 ± 0.01                & 0.56 ± 0.02                & 0.14 ± 0.01                 & 0.32 ± 0.00                 & 0.19 ± 0.01                & 0.73 ± 0.02                 \\
                                                                         & Ours                                                                         & 0.25 ±   0.02               & \underline{\textit{0.40 ± 0.03}}  & 0.60 ± 0.02                & 0.30 ± 0.01                & 0.58 ± 0.02                & \underline{\textbf{0.27 ± 0.01}}  & \underline{\textbf{0.60 ± 0.00}}  & 0.23 ± 0.01                & 0.61 ± 0.03                 \\ \hline
\end{tabular}
}
\end{table*}

\subsection{Experimental Results for Regression}
\label{subsect:res-regression}

\subsubsection{Illustration and Results on Cubic Regression}
\label{subsubsect:cubic}

In Fig. ~\ref{fig:unc-decomposition}, we illustrate the results produced by the baselines DE, EDL-R and SQR, and by our method without/with calibration.

For aleatoric estimates in the top row of Fig. ~\ref{fig:unc-decomposition}, DE and EDL-R yield accurate point predictions that align with the Gaussian expectation, while SQR targets the median, resulting in a visible bias between prediction and ground truth. In PI calibration, DE and EDL-R suffer from poor calibration due to prior mismatch, whereas SQR captures noise asymmetry but fails to provide adequate coverage, as indicated by its narrower upper bound. By comparison, our method without calibration in \eqref{eq:spi} improves alignment in both point predictions and PI boundaries, and our method with the SDS based calibration in \eqref{eq:calibrated-SPI} adaptively expands under covered intervals, which reduces smoothness and introduces mild over coverage but effectively mitigates local under coverage.

For epistemic estimates in the bottom row of Fig. ~\ref{fig:unc-decomposition}, within the iD region, DE shows mild fluctuations reflecting model inherent uncertainty in iD data, whereas EDL-R yields unstable and hard to interpret estimates due to distributional mismatch. OC maintains an almost constant uncertainty level because it is designed for OOD detection and thus fails to capture model uncertainty. In contrast, our method closely follows DE behavior owing to the implicit ensemble induced by the self consistent constraint. In OOD regions, both DE and OC display a sharp rise in epistemic uncertainty, successfully identifying OOD samples, while EDL-R continues to produce unreliable estimates. Our method similarly succeeds with a sharp rise, demonstrating its ability to support both fine grained epistemic estimation and OOD detection.

Additional results for the cubic regression including alternative noise distributions appear in Appendix~D-A1. 

\subsubsection{Results on UCI Benchmarks}
\label{subsubsect:uci-res}

Table~\ref{tab:UCI-results1} compares five baselines and our method. Our method preserve regression accuracy, achieving the lowest RMSE on seven of nine datasets, and improve PI quality across Winkler Score, PIECE, PIECE$^+$, and PIECE$^-$. While baselines often share similar PIECE, their divergent PIECE$^+$ and PIECE$^-$ expose calibration imbalance; our method maintains balanced, strong performance on both metrics, demonstrating fine-grained calibration.

Regarding epistemic uncertainty, DE and the two EDL-based regression methods exhibit stronger Spearman correlations between uncertainty estimates and RMSE, reflecting their reliance on repeated sampling or explicit distributional assumptions. In contrast, our method only ranks first on two and second on one of nine benchmarks due to its distribution-agnostic formulation. Nevertheless, unlike OC, it remains sensitive to model-related uncertainty in the iD regime.

When calibrated via \eqref{eq:calibrated-SPI}, our PIs become more reliable: both       $\text{PIECE}^+$ and $\text{PIECE}^-$ decrease on most datasets, effectively mitigating sub-interval under-coverage. While calibration marginally reduces PI sharpness according to the Winkler Score, the calibrated intervals remain among the top two on seven out of nine datasets, a favorable trade-off in safety-critical settings where slight over-coverage is preferable to under-coverage.

Overall, our method consistently ranks among the top performers across the nine UCI benchmarks in both accuracy and PI quality. Additional results under different synthetic noise distributions are provided in Appendix~D-A2.

\subsubsection{Results on Monocular Depth Estimation}
\label{subsubsect:depth-estimate}

Table~\ref{tab:Monocular-Depth-result} compares baseline methods and ours. Our method balances UQ performance and efficiency, retaining point prediction accuracy (RMSE) and achieving the lowest PIECE, $\text{PIECE}^+$ and $\text{PIECE}^-$ after calibration via \eqref{eq:calibrated-SPI}, without degrading PI sharpness as indicated by an unchanged Winkler Score. For epistemic uncertainty, OC performs poorly due to base model incompatibility, while our method remains highly competitive in OOD detection with an AUROC of 0.98. Although MD and DE slightly outperform us in AUROC, both require substantially longer inference times, and DE also has the longest training time of all methods.

\begin{table*}
\caption{Results on Monocular Depth Estimation}
\label{tab:Monocular-Depth-result}
\centering
\scalebox{0.8}{
\begin{tabular}{lllllllll}
\hline
                                                                             & RMSE                       & Winkler Score              & PIECE                      & $\text{PIECE}^+$           & $\text{PIECE}^-$           & AUROC                      & Training time (s)           & Inference time (ms)        \\ \hline
MD                                                                           & \underline{\textit{0.02 ± 0.00}} & \underline{\textit{0.12 ± 0.00}} & \underline{\textbf{0.01 ± 0.00}} & \underline{\textit{0.02 ± 0.00}} & \underline{\textbf{0.01 ± 0.01}} & \underline{\textit{0.99 ± 0.00}} & \underline{\textbf{25.34 ± 0.24}} & 64.42 ± 0.59               \\
DE                                                                           & \underline{\textbf{0.01 ± 0.00}} & \underline{\textbf{0.11 ± 0.01}} & 0.03 ± 0.00                & 0.03 ± 0.00                & 0.03 ± 0.00                & \underline{\textbf{1.00 ± 0.00}} & 99.92 ± 0.54                & 34.50 ± 0.03               \\
EDL-R                                                                        & \underline{\textit{0.02 ± 0.00}} & 0.14 ± 0.00                & 0.03 ± 0.01                & \underline{\textit{0.02 ± 0.01}} & 0.02 ± 0.01                & 0.98 ± 0.01                & 25.79 ± 0.64                & 3.11 ± 0.26                \\
EDL-QR                                                                       & \underline{\textit{0.02 ± 0.00}} & 0.15 ± 0.01                & 0.03 ± 0.01                & \underline{\textit{0.02 ± 0.01}} & 0.03 ± 0.01                & 0.97 ± 0.01                & 26.15 ± 0.62                & \underline{\textbf{3.06 ± 0.01}} \\
SQR-OC                                                                       & \underline{\textit{0.02 ± 0.00}} & 0.13 ± 0.00                & 0.05 ± 0.01                & 0.05 ± 0.01                & 0.04 ± 0.01                & 0.61 ± 0.01                & \underline{\textit{25.40 ± 0.18}} & \underline{\textit{3.08 ± 0.48}} \\
\multirow{2}{*}{\begin{tabular}[c]{@{}l@{}}Ours\\ Ours-Calib\end{tabular}} & \underline{\textit{0.02 ± 0.00}} & 0.13 ± 0.00                & 0.02 ± 0.00                & \underline{\textbf{0.00 ± 0.01}} & 0.02 ± 0.01                & 0.98 ± 0.01                & 26.87 ± 0.54                & 3.49 ± 0.26                \\
                                                                             & \underline{\textit{0.02 ± 0.00}} & 0.13 ± 0.00                & \underline{\textbf{0.01 ± 0.00}} & \underline{\textbf{0.00 ± 0.00}} & \underline{\textbf{0.01 ± 0.01}} & 0.98 ± 0.01                & 26.87 ± 0.54                & 4.20 ± 0.82                \\ \hline
\end{tabular}
}
\end{table*}

\begin{figure*}
  \centering
  \includegraphics[width=1\linewidth]{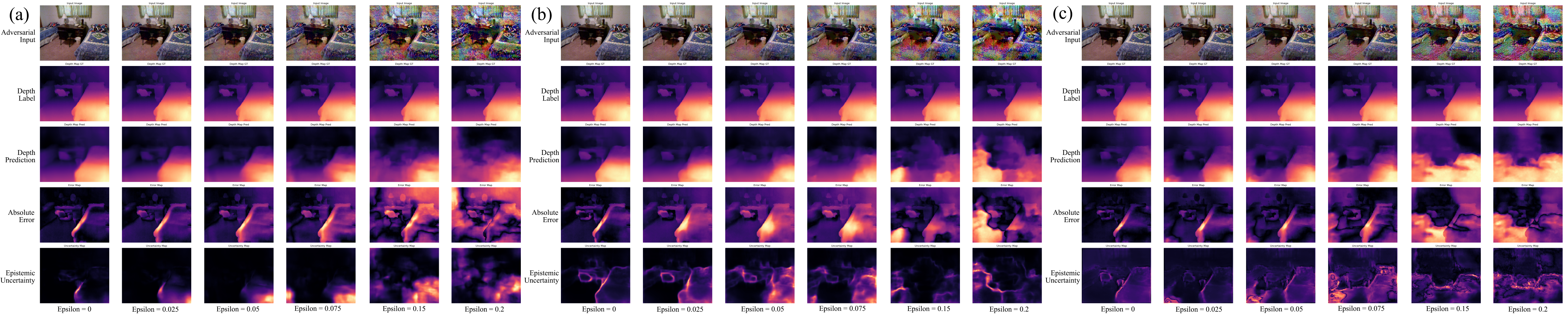}
  \caption{Visualization of model response to varying adversarial perturbations in monocular depth estimation. Rows (from top to bottom) show perturbed input images, ground truth depth maps, predicted outputs, error maps and estimated epistemic uncertainty. Columns correspond to (a) Deep Ensemble (DE), (b) Evidential Regression (EDL-R) and (c) our method.}
  \label{fig:Monocular-Depth-Visualization}
  \vspace*{-2mm}
\end{figure*}

To examine the correlation between model-inherent uncertainty and prediction error, we apply DE and EDL-R (the strongest competitors in Table~\ref{tab:Monocular-Depth-result}) alongside our method to adversarial images with varying noise strengths \(\epsilon\) and track the resulting uncertainty estimates. Fig. ~\ref{fig:Monocular-Depth-Visualization} illustrates their behavior.
As \(\epsilon\) increases, it is observed that DE in panel (a) exhibits minimal error map degradation due to ensemble averaging, with only small growth in highlighted regions, and it weakens the correlation between error and uncertainty since high error areas appear dark in the uncertainty map; EDL-R in panel (b) aligns uncertainty with error under mild perturbations but becomes unstable at higher \(\epsilon\); our method in panel (c) consistently highlights regions of both high error and high uncertainty, and its SDS increases in proportion to \(\epsilon\), reflecting a stable and accurate correlation as a fine-grain estimator.

These findings underscore our method’s practical potential. Additional results and analysis of perturbation strength versus uncertainty magnitude are provided in Appendix~D-B1.

\subsection{Experimental Results for Classification}
\label{subsect:res-class}

\subsubsection{Results on Image Classification Benchmarks}
\label{subsubsect:res-class-image}

Table~\ref{tab:CLS-results1} presents results on six configurations (three datasets $\times$ two architectures). For accuracy, OC, DDU, and our method share the same base model, so their TS-calibrated accuracies coincide; we therefore report only OC+TS and DDU+TS, and we also calibrate DE’s outputs with TS. As expected, DE+TS achieves the highest accuracy, while the other methods perform slightly worse but remain comparable across four configurations\footnote{No results for DE and LA on ImageNet due to computational constraints.}. 
In terms of ECE, TS generally improves the performance on CIFAR10 and CIFAR100, but can worsen it (e.g., VGG-16 on ImageNet). LA and EDL-C, which replace the softmax layer, exhibit higher ECE and cannot use standard calibration. In contrast, our SPA-based calibration via \eqref{eq:calibrated-predic} lowers ECE in five of six configurations than TS-calibration, demonstrating greater stability and reliability. 
For AUROC (error) on test data, our method ranks first in two configurations and second in three, demonstrating robust performance in diverse settings, especially for large, complex datasets such as ImageNet.

\begin{table*}

\caption{Epistemic Uncertainty Estimation and Confidence Calibration Results (\%) on Image Classification Benchmarks}

\label{tab:CLS-results1}

\centering

\scalebox{0.77}{

\begin{tabular}{cl|lllll|lllll}

\hline

\multirow{2}{*}{Dataset}  & \multicolumn{1}{c|}{\multirow{2}{*}{Method}} & \multicolumn{5}{c|}{VGG16}                                                                                                                         & \multicolumn{5}{c}{Wide-ResNet}                                                                                                                             \\ \cline{3-12}

                          & \multicolumn{1}{c|}{}                        & Accuracy                    & ECE                        & AUROC(error)                & AUROC(adv)                  & AUROC(ood)                  & Accuracy                             & ECE                        & AUROC(error)                & AUROC(adv)                  & AUROC(ood)                  \\ \hline

\multirow{6}{*}{CIFAR10}  & OC+TS                                        & \underline{\textit{93.62 ± 0.03}} & 1.75 ± 0.06                & 73.79 ± 0.50                & 40.36 ± 0.27                & 81.88 ± 0.95                & \underline{\textit{96.01 ± 0.03}}          & 0.92 ± 0.08                & 71.15 ± 0.21                & 38.74 ± 0.23                & 69.65 ± 1.81                \\

                          & DDU+TS                                       & \underline{\textit{93.62 ± 0.03}} & 1.75 ± 0.06                & 91.82 ± 0.10                & 59.66 ± 0.14                & \underline{\textit{89.77 ± 0.36}} & \underline{\textit{96.01 ± 0.03}}          & 0.92 ± 0.08                & 92.13 ± 0.11                & 71.12 ± 0.20                & \underline{\textit{95.27 ± 0.11}} \\

                          & EDL-C                                        & 93.41 ± 0.04                & 6.88 ± 0.05                & 90.37 ± 0.16                & \underline{\textbf{66.62 ± 0.27}} & 88.26 ± 0.52                & 95.77 ± 0.02                         & 9.15 ± 0.04                & 91.22 ± 0.26                & 78.16 ± 0.16                & 90.64 ± 0.40                \\

                          & LA                                           & \underline{\textit{93.62 ± 0.03}} & 5.62 ± 0.04                & 91.71 ± 0.10                & 65.59 ± 0.15                & 88.15 ± 0.45                & \underline{\textbf{96.02 ± 0.03}} & 2.59 ± 0.11                & \underline{\textit{94.34 ± 0.09}} & \underline{\textbf{78.69 ± 0.07}} & 92.92 ± 0.27                \\

                          & DE+TS                                        & \underline{\textbf{94.87 ± 0.03}} & \underline{\textbf{0.93 ± 0.04}} & \underline{\textbf{93.49 ± 0.08}} & 65.66 ± 0.22                & \underline{\textbf{91.02 ± 0.15}} & 96.00 ± 0.03                         & \underline{\textbf{0.60 ± 0.03}} & \underline{\textbf{95.06 ± 0.04}} & 75.75 ± 0.18                & \underline{\textbf{95.68 ± 0.09}} \\

                          & Ours-Calib                                   & 93.61 ± 0.03                & \underline{\textit{1.08 ± 0.06}} & \underline{\textit{91.86 ± 0.10}} & \underline{\textit{66.09 ± 0.17}} & 89.46 ± 0.30                & 96.00 ± 0.03                         & \underline{\textit{0.91 ± 0.15}} & 94.03 ± 0.11                & \underline{\textit{78.35 ± 0.71}} & 93.27 ± 0.18                \\ \hline

\multirow{6}{*}{CIFAR100} & OC+TS                                        & 73.51 ± 0.06                & \underline{\textit{3.04 ± 0.07}} & 76.29 ± 0.12                & 45.95 ± 0.05                & 77.61 ± 0.43                & \underline{\textit{80.88 ± 0.05}}          & \underline{\textit{3.82 ± 0.06}} & 54.85 ± 0.28                & 43.42 ± 0.20                & 38.49 ± 1.10                \\

                          & DDU+TS                                       & 73.51 ± 0.06                & \underline{\textit{3.04 ± 0.07}} & 84.46 ± 0.12                & 53.64 ± 0.06                & \underline{\textbf{79.99 ± 0.39}} & \underline{\textit{80.88 ± 0.05}}          & \underline{\textit{3.82 ± 0.06}} & 77.22 ± 0.11                & 63.26 ± 0.16                & 84.44 ± 0.39                \\

                          & EDL-C                                        & \underline{\textit{73.55 ± 0.07}} & 26.20 ± 0.06               & 84.98 ± 0.33                & \underline{\textit{59.78 ± 0.06}} & 77.73 ± 0.46                & 75.80 ± 0.15                         & 34.33 ± 0.11               & 61.62 ± 0.59                & 61.52 ± 0.12                & 82.75 ± 0.38                \\

                          & LA                                           & 73.47 ± 0.06                & 58.15 ± 0.10               & 83.10 ± 0.10                & 54.50 ± 0.08                & 76.84 ± 0.45                & 80.65 ± 0.04                         & 71.87 ± 0.07               & 85.17 ± 0.11                & 66.33 ± 0.08                & \underline{\textit{85.37 ± 0.19}} \\

                          & DE+TS                                        & \underline{\textbf{77.63 ± 0.10}} & \underline{\textbf{2.37 ± 0.17}} & \underline{\textbf{87.39 ± 0.04}} & \underline{\textbf{59.83 ± 0.11}} & \underline{\textit{79.59 ± 0.16}} & \underline{\textbf{83.35 ± 0.04}}          & 4.09 ± 0.08                & \underline{\textit{87.83 ± 0.03}} & \underline{\textbf{68.30 ± 0.11}} & \underline{\textbf{86.74 ± 0.15}} \\

                          & Ours-Calib                                   & 73.38 ± 0.05                & 11.14 ± 0.06               & \underline{\textit{86.20 ± 0.10}} & 57.13 ± 0.07                & 75.00 ± 0.36                & 80.71 ± 0.06                         & \underline{\textbf{3.22 ± 0.09}} & \underline{\textbf{88.00 ± 0.08}} & \underline{\textit{67.68 ± 0.05}} & 83.35 ± 0.26                \\ \hline

\multirow{4}{*}{ImageNet} & OC+TS                                        & \underline{\textbf{71.59 ± 0.00}} & 7.80 ± 0.00                & 31.44 ± 0.08                & 28.77 ± 0.33                & 45.87 ± 0.57                & \underline{\textbf{81.30 ± 0.00}}          & 8.18 ± 1.14                & 69.36 ± 0.09                & 81.30 ± 0.11                & 56.22 ± 0.20                \\

                          & DDU+TS                                       & \underline{\textbf{71.59 ± 0.00}} & 7.80 ± 0.00                & 63.95 ± 0.00                & 66.90 ± 0.00                & \underline{\textbf{69.11 ± 0.00}} & \underline{\textbf{81.30 ± 0.00}}          & 8.18 ± 1.14                & 67.75 ± 0.00                & \underline{\textit{86.86 ± 0.00}} & \underline{\textit{71.02 ± 0.00}} \\

                          & EDL-C                                        & 61.96 ± 0.41                & \underline{\textbf{2.28 ± 0.33}} & 83.14 ± 2.13                & 72.72 ± 4.33                & 59.07 ± 0.52                & 77.11 ± 0.20                         & \underline{\textbf{5.83 ± 0.03}} & \underline{\textbf{88.80 ± 0.12}} & 83.25 ± 0.14                & 56.56 ± 0.12                \\

                          & Ours-Calib                                   & \underline{\textbf{71.59 ± 0.01}} & \underline{\textit{2.82 ± 0.01}} & \underline{\textbf{82.02 ± 0.02}} & \underline{\textbf{83.70 ± 0.18}} & \underline{\textit{60.13 ± 0.02}} & \underline{\textit{80.91 ± 0.04}}          & \underline{\textit{5.98 ± 0.00}} & \underline{\textit{73.38 ± 0.02}} & \underline{\textbf{89.19 ± 0.00}} & \underline{\textbf{73.92 ± 0.02}} \\ \hline

\end{tabular}

}

\end{table*}

\begin{table*}[ht]
\caption{Results on Multimodal LUMA Benchmark}
\label{tab:LUMA-results}
\centering
\scalebox{0.8}{
\begin{tabular}{ll|cc|cc|cc|cc}
\hline
\multirow{2}{*}{Modal} & \multicolumn{1}{c|}{\multirow{2}{*}{Method}} & \multicolumn{2}{c|}{Clean}                   & \multicolumn{2}{c|}{$\searrow$ Diversity}         & \multicolumn{2}{c|}{$\nearrow$ Label Noise}       & \multicolumn{2}{c}{$\nearrow$ Sample Noise}     \\
                       & \multicolumn{1}{c|}{}                        & Accuracy               & AUROC               & Aleatoric               & Epistemic               & Aleatoric               & Epistemic               & Aleatoric              & Epistemic              \\ \hline
\multirow{3}{*}{Image} & MD                                           & 32.69\%                & \underline{\textit{0.56}} & \underline{\textbf{-15.73\%}} & -11.66\%                & \underline{\textbf{59.20\%}}  & \underline{\textbf{54.51\%}}  & \underline{\textbf{4.44\%}}  & \underline{\textbf{2.18\%}}  \\
                       & DE                                           & \underline{\textbf{40.31\%}} & 0.51                & -37.49\%                & \underline{\textbf{-8.54\%}}  & -7.43\%                 & \underline{\textit{0.24\%}}   & -18.46\%               & -3.22\%                \\
                       & Ours-Calib                                         & \underline{\textit{38.21\%}} & \underline{\textbf{0.62}} & \underline{\textit{-15.86\%}} & \underline{\textit{-10.02\%}} & \underline{\textit{21.97\%}}  & -2.38\%                 & \underline{\textit{-6.51\%}} & \underline{\textit{-1.26\%}} \\ \hline
\multirow{3}{*}{Audio} & MD                                           & 83.38\%                & 0.50                & \underline{\textbf{-5.54\%}}  & \underline{\textbf{2.16\%}}   & 96.63\%                 & \underline{\textit{54.49\%}}  & 23.12\%                & 14.40\%                \\
                       & DE                                           & \underline{\textbf{91.60\%}} & \underline{\textit{0.54}} & -27.39\%                & \underline{\textit{-3.34\%}}  & \underline{\textit{156.40\%}} & 50.43\%                 & \underline{\textbf{70.26\%}} & \underline{\textit{34.41\%}} \\
                       & Ours-Calib                                         & \underline{\textit{87.36\%}} & \underline{\textbf{0.74}} & \underline{\textit{-15.67\%}} & -7.23\%                 & \underline{\textbf{294.29\%}} & \underline{\textbf{152.64\%}} & \underline{\textit{63.21\%}} & \underline{\textbf{51.69\%}} \\ \hline
\multirow{3}{*}{Text}  & MD                                           & 96.62\%                & 0.50                & -3.91\%                 & -2.62\%                 & \underline{\textit{93.59\%}}  & \underline{\textit{2.41\%}}   & \underline{\textit{64.96\%}} & \underline{\textit{-2.03\%}} \\
                       & DE                                           & \underline{\textbf{97.00\%}} & \underline{\textit{0.56}} & \underline{\textbf{5.02\%}}   & -6.15\%                 & 81.26\%                 & -0.51\%                 & 62.24\%                & -7.11\%                \\
                       & Ours-Calib                                         & \underline{\textit{96.02\%}} & \underline{\textbf{0.83}} & \underline{\textit{4.38\%}}   & \underline{\textbf{1.86\%}}   & \underline{\textbf{359.46\%}} & \underline{\textbf{186.45\%}} & \underline{\textbf{70.03\%}} & \underline{\textbf{88.89\%}} \\ \hline
\multirow{4}{*}{Multi} & MD                                           & 98.93\%                & 0.50                & -8.52\%                 & -1.21\%                 & \underline{\textit{122.44\%}} & 11.60\%                 & \underline{\textbf{59.14\%}} & 9.89\%                 \\
                       & DE                                           & \underline{\textbf{99.48\%}} & 0.53                & -22.80\%                & -3.40\%                 & 115.15\%                & 20.62\%                 & 45.97\%                & 5.54\%                 \\
                       & RCML (EDL)                                          & 94.86\%                & \underline{\textbf{0.91}} & \underline{\textit{8.34\%}}   & \underline{\textbf{16.16\%}}  & 64.72\%                 & \underline{\textit{106.16\%}} & 36.19\%                & \underline{\textit{58.21\%}} \\
                       & Ours-Calib                                        & \underline{\textit{99.10\%}} & \underline{\textit{0.90}} & \underline{\textbf{9.76\%}}   & \underline{\textit{12.05\%}}  & \underline{\textbf{340.16\%}} & \underline{\textbf{348.06\%}} & \underline{\textit{55.80\%}} & \underline{\textbf{59.52\%}} \\ \hline
\end{tabular}
}
\end{table*}

According to the AUROC (adv) results, DE, EDL-C and LA demonstrate satisfactory adversarial detection on four configurations involving CIFAR10 and CIFAR100, while all other baselines underperform across six configurations. In contrast, our method ranks first on two ImageNet configurations and second on the remaining four, indicating its fine-grained adversarial detection capability. Nevertheless, DE and LA are multi-pass methods that incur intensive training and substantially longer inference times, whereas EDL-C is an internal method based on distributional assumptions and is difficult to integrate with an already deployed DL model.

According to the AUROC (ood) results, DE and DDU are the strongest performers overall, jointly placing in the top two for five of six configurations. This reflects DDU’s specialization in OOD detection and DE’s ensemble robustness. 
Our method ranks first on one ImageNet configuration and remains competitive on the other baselines, notably outperforming OC, designed specifically for OOD detection, demonstrating that our SDS metric serves both as a fine-grained estimator and an effective OOD detector.

Additional analysis of perturbation strength versus uncertainty magnitude appears in Appendix~D-B2.

\subsubsection{Results on Multimodal Classification}
\label{subsubsect:multimodal-res}

The LUMA benchmark~\cite{luma_dataset2025} supplies a pretrained base model and results for MD and DE in both uni- and multimodal settings, as well as an EDL-based multimodal UQ baseline, RCML~\cite{xu2024reliable}. We adopt the same base model and directly compare our method to these baselines. Following the LUMA protocol, Table~\ref{tab:LUMA-results} presents performance on clean data and uncertainty estimates across varying dataset conditions.

On clean data, DE achieves the highest accuracy in both uni- and multimodal settings, as expected. Our calibrated method then outperforms all other baselines across these settings, falling marginally behind DE only in the text and multimodal cases. For AUROC, our approach ranks first in every setting except the multimodal one, where it slightly trails RCML. These results highlight the effectiveness of our method across diverse data modalities.

When models are trained on less diverse subsets, most methods show reduced sensitivity to both aleatoric and epistemic uncertainty in image and audio modalities; MD is the exception, exhibiting increased epistemic uncertainty. For text and multimodal data, our method increases both uncertainties, with only DE producing a larger aleatoric rise on text and RCML yielding a larger epistemic rise on multimodal data. Despite using data of reduced diversity, our method effectively quantifies both uncertainty types in multimodal scenarios.
  
In the noisy-label scenario, our method outperforms all others in quantifying both uncertainty types for audio, text, and multimodal data; only MD surpasses ours on both metrics for images, indicating our method’s greater sensitivity to label-noise uncertainty in the other modalities.

In the noisy-sample scenario, our method outperforms others in both aleatoric and epistemic measures for audio, text, and multimodal data; DE surpasses ours on the aleatoric metric for audio, and MD does so for multimodal. Although our method exceeds DE on both metrics for images, MD achieves the highest scores in that modality. Overall, our method effectively handles sample noise across all modalities except images.

\subsection{Summary of Extended Experimental Findings}
\label{subsect:other-res}

We also experimented with joint training of a task base model and our UQ network from scratch, varying the volumes of training and calibration data and the UQ network’s MLP architecture (see Appendix E for details). The main findings are: (i) joint training can slightly degrade base-model performance due to increased optimization complexity but often improves epistemic uncertainty accuracy; (ii) the UQ network can be trained with limited data, though uncertainty estimates may suffer; (iii) softmax calibration in classification generally enhances reliability, but larger calibration sets do not guarantee better performance; and (iv) a shallow MLP regressor is typically sufficient for the UQ network.

\section{Conclusion}
\label{sect:conclude}

We have proposed a unified post-hoc UQ framework for deep learning grounded in the split-point self-consistency principle. Our method overcomes key limitations of existing methods and integrates seamlessly with already deployed DL models. Extensive comparative evaluations demonstrate that split-point quantile regression yields more accurate prediction interval coverage in regression, and that the Self-Consistency Discrepancy Score (SDS) is a theoretically sound, fine grained epistemic metric applicable to both regression and classification, and can be further utilized to enhance interval coverage in regression and improve confidence calibration in classification.

Nevertheless, our method has several potential limitations. 
First, self-consistency verification locates a single zero-minimum of the SDS landscape; if multiple zero-minima exist, it may in theory select an improper one, causing systematic biases to evade detection or skew epistemic estimates.
Second, while SDS provides fine-grained epistemic measurements, high-uncertainty in-distribution samples can be mistaken for OOD, reducing its specificity compared to dedicated detectors. 
Third, by eschewing distributional assumptions, our framework cannot exploit known data priors and may underperform methods tailored to specific distributions. 
Fourth, our current implementation relies on flat, vectorial feature maps and may not generalize to structured representations (e.g., graphs or sequences) without adapting the UQ regressor. 
Finally, our method applies only to supervised learning and does not yet extend to unsupervised, semi-supervised, or reinforcement learning settings.

Our future outlook tackles these challenges on multiple fronts: investigating self-consistency criteria and robust optimization to align predictions with one proper SDS zero-minimum; combining SDS with complementary in-distribution measures for stronger OOD discrimination; incorporating soft priors (e.g., noise models or physics constraints) into SPA; developing mesh- or graph-based UQ regressors for structured feature spaces; extending the framework to reinforcement learning and generative modeling; and validating in real-world, uncertainty-aware domains such as autonomous driving, medical diagnosis, and climate modeling.

\appendices

\section{Derivation and Proofs}
\label{appendix-A}

\subsection{Derivation of MARs}
\label{app:mar-derivation}

\subsubsection{Derivation of MARs in Regression}
\label{subsubsect:mar-derivation-regress}

Recall the definition from the main text: for each pair $(\bm{x}, y) \in \mathcal{D}$, let the base model prediction be $\tilde{y} = f(\bm{x}; \Theta^*)$ and define the residual $r = y - \tilde{y}$. We collect the set of input-residual pairs where the residual is non-zero:
\(
    \mathcal{R} = \{(\bm{x}, r) \mid r \neq 0, (\bm{x}, y) \in \mathcal{D}\}
\).
This set is then partitioned based on the sign of the residual into a set of positive residuals: 
\(
    \mathcal{R}^+ = \{(\bm{x}, r) \mid r > 0, (\bm{x}, y) \in \mathcal{D}\}
\),
and a set of negative residuals: 
\(
    \mathcal{R}^- = \{(\bm{x}, r) \mid r < 0, (\bm{x}, y) \in \mathcal{D}\}
\).

Based on this partitioning, we need to derive the total, upper-side, and lower-side \emph{mean absolute residuals} (MARs) for any prediction \(\tilde y\) as functions of the input \(\bm{x}\):
\begin{align}
\mathrm{MAR}(\tilde y|\bm{x})   &= \mathbb{E}\bigl[|r| \mid (\bm {x'}, r)\in\mathcal R, \bm{x'}=\bm{x} \bigr], \nonumber\\
\mathrm{MAR}^{+}(\tilde y|\bm{x}) &= \mathbb{E}\bigl[|r| \mid (\bm {x'}, r)\in\mathcal R^+, \bm{x'}=\bm{x} \bigr], \nonumber\\
\mathrm{MAR}^{-}(\tilde y|\bm{x}) &= \mathbb{E}\bigl[|r| \mid (\bm {x'}, r)\in\mathcal R^-, \bm{x'}=\bm{x} \bigr]. \nonumber
\end{align}

The MARs align with the heteroscedastic regression formulation in (1) of the main text. 
When the base model $f(\bm{x}; \Theta^*)$ coincides with the true function $F(\bm{x})$, 
the MARs quantify the absolute expectations of the data noise $\varepsilon(\bm{x})$:
\begin{align}
\mathrm{MAR}(\tilde y|\bm{x}) = \mathbb{E}\!\left[\,|\varepsilon(\bm{x})| \,\middle|\, \varepsilon(\bm{x}) \neq 0 \right], \nonumber\\
\mathrm{MAR}^{+}(\tilde y|\bm{x}) = \mathbb{E}\!\left[\,|\varepsilon(\bm{x})| \,\middle|\, \varepsilon(\bm{x}) > 0 \right], \nonumber\\
\mathrm{MAR}^{-}(\tilde y|\bm{x}) = \mathbb{E}\!\left[\,|\varepsilon(\bm{x})| \,\middle|\, \varepsilon(\bm{x}) < 0 \right]. \nonumber
\end{align}

While these theoretical definitions are formulated at a single point $\bm{x}$, their practical estimation requires a smoothness assumption \cite{chung2021beyond}: that the conditional distribution of the residual does not change abruptly with $\bm{x}$. Formally, if $\bm{x}_j \approx \bm{x}_k$, then the residual distribution given $\bm{x}_j$ is similar to that given $\bm{x}_k$.

This assumption allows us to estimate the conditional expectations by averaging over a local neighborhood $\mathcal{N}(\bm{x})$ around the point $\bm x$. The estimable MARs are thus defined as:
\begin{align}
    \mathrm{MAR}(\tilde y|\bm{x})   &\approx \mathbb{E}\bigl[|r| \mid (\bm{x'}, r) \in \mathcal{R}, \bm{x'} \in \mathcal{N}(\bm{x}) \bigr], \nonumber\\
    \mathrm{MAR}^{+}(\tilde y|\bm{x}) &\approx \mathbb{E}\bigl[|r| \mid (\bm{x'}, r) \in \mathcal{R}^+, \bm{x'} \in \mathcal{N}(\bm{x}) \bigr], \nonumber\\
    \mathrm{MAR}^{-}(\tilde y|\bm{x}) &\approx \mathbb{E}\bigl[|r| \mid (\bm{x'}, r) \in \mathcal{R}^-, \bm{x'} \in \mathcal{N}(\bm{x}) \bigr]. \nonumber
\end{align}

By computing these quantities for different neighborhoods across the input space $\mathcal{X}$, one can obtain an empirical estimate of the conditional residual distribution. This can be achieved, for example, by using $k$-nearest neighbors or kernel-based methods to define $\mathcal{N}(\bm{x})$ \cite{chung2021beyond}, or following statistical decision theory~\cite[Section~2.4]{hastie2009elements}, by regressing the conditional mean with a nonlinear regressor trained via the mean squared loss. To adapt deep learning tasks, all MARs in our UQ framework for both regression and classification are estimated using nonlinear regressors grounded in statistical decision theory.

\subsubsection{Derivation of MARs in Classification}
\label{subsubsect:mar-derivation-class}

For each class \(k\), let the softmax output be \(\tilde y_k\in(0,1)\) and the one-hot label \(y_k\in\{0,1\}\), so the residual is
\[
r_k = y_k - \tilde y_k.
\]
Form the nonzero-residual set
\[
\mathcal R_k = \bigl\{(\bm x,r_k)\mid(\bm x,y_k)\in\mathcal D,\;r_k\neq0\bigr\},
\]
and partition by sign:
\begin{align}
\mathcal R_k^+ &= \bigl\{(\bm x,r_k)\in\mathcal R_k \mid r_k>0\bigr\}, \nonumber\\
\mathcal R_k^- &= \bigl\{(\bm x,r_k)\in\mathcal R_k \mid r_k<0\bigr\}. \nonumber
\end{align}

We need to derive the pointwise MARs for any prediction \(\tilde y_k\) on input \(\bm x \in \mathcal{D}\) as:
\begin{align}
\mathrm{MAR}(\tilde y_k|\bm x)
&= \mathbb{E}\bigl[\,|r_k|\mid(\bm x',r_k)\in\mathcal R_k,\;\bm x'=\bm x\bigr],\nonumber\\
\mathrm{MAR}^+(\tilde y_k|\bm x)
&= \mathbb{E}\bigl[\,|r_k|\mid(\bm x',r_k)\in\mathcal R_k^+,\;\bm x'=\bm x\bigr],\nonumber\\
\mathrm{MAR}^-(\tilde y_k|\bm x)
&= \mathbb{E}\bigl[\,|r_k|\mid(\bm x',r_k)\in\mathcal R_k^-,\;\bm x'=\bm x\bigr].\nonumber
\end{align}

Since \(y_k\in\{0,1\}\) and \(\tilde y_k\in(0,1)\), note that
\[
y_k>\tilde y_k\iff y_k=1,
\quad
y_k<\tilde y_k\iff y_k=0.
\]
Hence
\begin{align}
\mathrm{MAR}(\tilde y_k|\bm x)
&= \mathbb{E}\bigl[\,|y_k-\tilde y_k|\mid\bm x\bigr]\nonumber\\
&= P_k(\bm x)\,(1-\tilde y_k)+(1-P_k(\bm x))\,\tilde y_k,\nonumber\\
\mathrm{MAR}^+(\tilde y_k|\bm x)
&= \mathbb{E}\bigl[y_k-\tilde y_k\mid y_k>\tilde y_k,\bm x\bigr]\nonumber\\
&= 1-\tilde y_k,\nonumber\\
\mathrm{MAR}^-(\tilde y_k|\bm x)
&= \mathbb{E}\bigl[\tilde y_k-y_k\mid y_k<\tilde y_k,\bm x\bigr]\nonumber\\
&= \tilde y_k, \nonumber
\end{align}
where 
\(\displaystyle P_k(\bm x)=\Pr(y_k=1|~\bm x)\), which can be empirically estimated from a training dataset.

These align with the heteroscedastic classification form in (2) of the main text.  In the noise-interpretation view:
\begin{align}
\varepsilon(\bm x) &= y_k - \tilde y_k,\nonumber\\
\mathrm{MAR}(\tilde y_k|\bm x) &= \mathbb{E}\bigl[|\varepsilon(\bm x)|\mid\varepsilon(\bm x)\neq0\bigr],\nonumber\\
\mathrm{MAR}^+(\tilde y_k|\bm x) &= \mathbb{E}\bigl[|\varepsilon(\bm x)|\mid\varepsilon(\bm x)>0\bigr],\nonumber\\
\mathrm{MAR}^-(\tilde y_k|\bm x) &= \mathbb{E}\bigl[|\varepsilon(\bm x)|\mid\varepsilon(\bm x)<0\bigr].\nonumber
\end{align}

\subsubsection{Derivation of Zero-Included MARs}
\label{app:zero-mar-derivation}

Recall the definition in the main text, the zero-included residuals for class \( k \) in the calibration set \( \mathcal{D}_{\mathrm{C}} \) are:   
\[
r_k = y_{\mathrm{C},k} - \tilde{y}_{\mathrm{C},k}, \quad
r_k^+ = \max\{r_k, 0\}, \quad
r_k^- = \min\{r_k, 0\},
\]
where \( y_{\mathrm{C},k} \in \{0,1\} \) is the one-hot label and \( \tilde{y}_{\mathrm{C},k} \in (0,1) \) is the softmax prediction for class \( k \).

Based on these residuals, we need to derive the  \textit{zero-included Mean Absolute Residuals}, 
$\mathrm{MAR}_{\mathrm{C}}, \mathrm{MAR}_{\mathrm{C}}^+, \mathrm{MAR}_{\mathrm{C}}^-$, for any prediction \(\tilde y_{{\mathrm C},k}\) on input \(\bm x \in \mathcal{D}_{\mathrm{C}}\):
\begin{align*}
\mathrm{MAR}_{\mathrm{C}}(\tilde y_{{\mathrm C},k}|\bm x)   &= \mathbb{E}_{r_k \in  \{ r_k \mid (\bm{x}_{\mathrm{C}}, y_{\mathrm{C},k}) \in \mathcal{D}_{\mathrm{C}} \}} \left[ |r_k| \right], \\
\mathrm{MAR}_{\mathrm{C}}^+(\tilde y_{{\mathrm C},k}|\bm x) &= \mathbb{E}_{r_k \in \{ r_k^+ \mid (\bm{x}_{\mathrm{C}}, y_{\mathrm{C},k}) \in \mathcal{D}_{\mathrm{C}} \}} \left[ |r_k| \right], \\
\mathrm{MAR}_{\mathrm{C}}^-(\tilde y_{{\mathrm C},k}|\bm x) &= \mathbb{E}_{r_k \in \{ r_k^- \mid (\bm{x}_{\mathrm{C}}, y_{\mathrm{C},k}) \in \mathcal{D}_{\mathrm{C}} \}} \left[ |r_k| \right].
\end{align*}

Since \( y_{\mathrm{C},k} \in \{0,1\} \) and \( \tilde{y}_{\mathrm{C},k} \in (0,1) \), we derive:
\begin{align*}
\mathrm{MAR}_{\mathrm{C}}(\tilde y_{{\mathrm C},k}|\bm x)
&= \mathbb{E} \left[ |y_{\mathrm{C},k} - \tilde{y}_{\mathrm{C},k}| \right] \\
&= \mathbb{E} \left[ 
\mathbb{I}(y_{\mathrm{C},k} = 1)(1 - \tilde{y}_{\mathrm{C},k}) 
+ \mathbb{I}(y_{\mathrm{C},k} = 0)\tilde{y}_{\mathrm{C},k} 
\right] \\
&= P_{\mathrm{C},k}(\bm x)(1 - \tilde{y}_{\mathrm{C},k}) + (1 - P_{\mathrm{C},k}(\bm x))\tilde{y}_{\mathrm{C},k},
\end{align*}
\begin{align*}
\mathrm{MAR}_{\mathrm{C}}^+(\tilde y_{{\mathrm C},k}|\bm x) 
&= \mathbb{E} \left[ \max\{y_{\mathrm{C},k} - \tilde{y}_{\mathrm{C},k}, 0\} \right] \\
&= \mathbb{E} \left[ \mathbb{I}(y_{\mathrm{C},k} = 1)(1 - \tilde{y}_{\mathrm{C},k}) \right] \\
&= P_{\mathrm{C},k}(\bm x)(1 - \tilde{y}_{\mathrm{C},k}),
\end{align*}
\begin{align*}
\mathrm{MAR}_{\mathrm{C}}^-(\tilde y_{{\mathrm C},k}|\bm x)
&= \mathbb{E} \left[ |\min\{y_{\mathrm{C},k} - \tilde{y}_{\mathrm{C},k}, 0\}| \right] \\
&= \mathbb{E} \left[ \mathbb{I}(y_{\mathrm{C},k} = 0)\tilde{y}_{\mathrm{C},k} \right] \\
&= (1 - P_{\mathrm{C},k}(\bm x))\tilde{y}_{\mathrm{C},k}.
\end{align*}
where \( P_{\mathrm{C},k}(\bm x) \) denotes the conditional frequency of class \( k \) in \( \mathcal{D}_{\mathrm{C}} \).

\subsection{Proof of Theorem \ref{th:sc} (Self-Consistency Constraint)  }
\label{sec:SC-proof}

\begin{theorem}[Self-Consistency Constraint] 
\label{th:sc} 
Let \(Y\) be a real-valued random variable with \(|\mathbb{E}[Y]|<\infty\). For a split-point $t \in \mathbb{R}$, define the total Mean Absolute Deviation (MAD), upper-side $\mathrm{MAD}^+$, and lower-side 
$\mathrm{MAD}^-$ by 
\begin{align*}
\mathrm{MAD} &= \mathbb{E}\bigl[\lvert Y - t \rvert \mid Y \ne t \bigr], \\
\mathrm{MAD}^+ &= \mathbb{E}\bigl[ Y - t \mid Y > t \bigr], \\
\mathrm{MAD}^- &= \mathbb{E}\bigl[ t - Y \mid Y < t \bigr].
\end{align*}
When $t = \mathbb{E}[Y]$, assuming 
$P(Y > t) > 0$ and $P(Y < t) > 0$, the following identity holds: 
$$ \mathrm{MAD} = H\bigl(\mathrm{MAD}^+,\, \mathrm{MAD}^-\bigr) = \frac{2\, \mathrm{MAD}^+\, \mathrm{MAD}^-} {\mathrm{MAD}^+ + \mathrm{MAD}^-}, 
$$ 
 where $H(a,b) = 2ab/(a + b)$ denotes the harmonic mean. 
\end{theorem}
\begin{proof}
Let $p^+ = P(Y > t)$ and $p^- = P(Y < t)$. By the law of total expectation, the total MAD (conditioned on $Y \ne t$) can be expressed as a weighted average of $\mathrm{MAD}^+$ and $\mathrm{MAD}^-$:
{\small
\begin{align*}
&\mathbb{E}\bigl[\lvert Y - t \rvert \mid Y \ne t\bigr] \\
&= \mathbb{E}\bigl[\lvert Y - t \rvert \mid Y > t\bigr] P(Y > t \mid Y \ne t) \\
&\quad + \mathbb{E}\bigl[\lvert Y - t \rvert \mid Y < t\bigr] P(Y < t \mid Y \ne t) \\
&= \mathbb{E}\bigl[Y - t \mid Y > t\bigr] \frac{p^+}{p^+ + p^-} + \mathbb{E}\bigl[t - Y \mid Y < t\bigr] \frac{p^-}{p^+ + p^-} \\
&= \mathrm{MAD}^+ \cdot \frac{p^+}{p^+ + p^-} + \mathrm{MAD}^- \cdot \frac{p^-}{p^+ + p^-} 
\end{align*}
}

Next, we leverage the fundamental property of the mean, $\mathbb{E}[Y - t] = 0$. Applying the law of total expectation again:
\[
\begin{aligned}
\mathbb{E}[Y - t] &= \mathbb{E}[Y - t \mid Y > t]\,p^+ + \mathbb{E}[Y - t \mid Y < t]\,p^- \\
&\quad + \mathbb{E}[Y - t \mid Y = t]\,P(Y = t) \\
&= (\mathrm{MAD}^+)\,p^+ + (-\mathrm{MAD}^-)\,p^- + 0 \\
&= 0 \\
\implies\quad & p^+ \cdot \mathrm{MAD}^+ = p^- \cdot \mathrm{MAD}^-
\end{aligned}
\]
This equation implies a ratio of probabilities: $\frac{p^+}{p^-} = \frac{\mathrm{MAD}^-}{\mathrm{MAD}^+}$.

Substituting $p^+ = \frac{\mathrm{MAD}^-}{\mathrm{MAD}^+} p^-$ into the weighted average expression:
\begin{align*}
\mathrm{MAD} &= 
\frac{\frac{\mathrm{MAD}^-}{\mathrm{MAD}^+} p^-}{\frac{\mathrm{MAD}^-}{\mathrm{MAD}^+} p^- + p^-} \cdot \mathrm{MAD}^+ 
 + \frac{p^-}{\frac{\mathrm{MAD}^-}{\mathrm{MAD}^+} p^- + p^-} \cdot \mathrm{MAD}^- \\
&= \frac{\mathrm{MAD}^- \cdot \mathrm{MAD}^+}{\mathrm{MAD}^- + \mathrm{MAD}^+} 
 + \frac{\mathrm{MAD}^+ \cdot \mathrm{MAD}^-}{\mathrm{MAD}^- + \mathrm{MAD}^+}.
\end{align*}

Combining the two terms:
\[
\mathrm{MAD} 
= \frac{2 \cdot \mathrm{MAD}^+ \cdot \mathrm{MAD}^-}{\mathrm{MAD}^+ + \mathrm{MAD}^-}.
\]
\end{proof}

\textit{Remark}:
Under heteroscedasticity, choosing \(t(\bm{x}) = \mathbb{E}[Y \mid X = \bm{x}]\) yields a conditional self-consistency constraint, extending the global identity to each conditional distribution.

\subsection{Proof of Proposition \ref{prop-sc} (Minimum Discrepancy)}
\label{subsect:proof-prop}

\begin{proposition}[Minimum Discrepancy]
\label{prop-sc}
For any $t \in \mathbb{R}$ with $P(Y > t)>0$ and $P(Y < t)>0$, define the self-consistency discrepancy:
\[
\Delta(t) := \left| \mathrm{MAD} - H\bigl(\mathrm{MAD}^+,\, \mathrm{MAD}^-\bigr) \right|.
\]
Then $\Delta(t)$ attains its global minimum of zero when $t = \mathbb{E}[Y]$ and at any balance points where $\mathrm{MAD}^+ = \mathrm{MAD}^-$.
\end{proposition}

\begin{proof}
To simplify notations from Appendix~\ref{sec:SC-proof}, we denote
\[
a := \mathrm{MAD}^+, \quad b := \mathrm{MAD}^-, \quad p := p^+ + p^-.
\]

From the proof in Theorem~\ref{th:sc}, the total MAD can be written as
\[
\operatorname{MAD} = \frac{p^+}{p} \cdot a + \frac{p^-}{p} \cdot b,
\]
and the discrepancy becomes:

\begin{align*}
\Delta(t) &= \left| \frac{p^+}{p} a + \frac{p^-}{p} b - \frac{2ab}{a + b} \right| \\
&= \left| \frac{p^+ a + p^- b}{p} - \frac{2ab}{a + b} \right| \\
&= \frac{1}{p(a + b)} \cdot \left| (p^+ a + p^- b)(a + b) - 2pab \right|.
\end{align*}

Simplify the numerator:
\begin{align*}
(p^+ a + p^- b)(a + b)
&= p^+ a^2 + p^+ ab + p^- ab + p^- b^2 \\
&= p^+ a^2 + p^- b^2 + (p^+ + p^-) ab\\
&= p^+ a^2 + p^- b^2 + pab.
\end{align*}

Thus,
\begin{align*}
\Delta(t) &= \frac{1}{p(a + b)} \cdot \left| p^+ a^2 + p^- b^2 + pab - 2pab \right| \\
&= \frac{1}{p(a + b)} \cdot \left| p^+ a^2 + p^- b^2 - pab \right|.
\end{align*}

Now observe that:
\[
p^+ a^2 + p^- b^2 - pab = (p^+ a - p^- b)(a - b),
\]

Since \( \frac{1}{p(a + b)} > 0 \), we have \( \Delta(t) = 0 \) if and only if:
\[
(p^+ a - p^- b)(a - b) = 0,
\]
i.e., either \(a = b\) or \(p^+ a = p^- b\). Otherwise, \( \Delta(t) > 0 \).

If \(p^{+}a=p^{-}b\), then
\(
p^{+}a-p^{-}b=\mathbb{E}[Y-t]=0
\), 
so \(t=\mathbb{E}[Y]\).

If $a=b $, then $t$ is a balance point where $\mathrm{MAD}^+ = \mathrm{MAD}^-$. 
Since $\Delta(t) \geq 0$ everywhere and equals zero precisely at these points, they are the global minima.

\end{proof}

\textit{Remark.} Let \(Y\) be a real-valued random variable with \(\lvert\mathbb{E}[Y]\rvert<\infty\).  From the proof of Proposition~\ref{prop-sc}, we derive the following implications, followed by a practical note from a machine learning perspective.

\medskip
\noindent\textbf{Equal-MAD points.} Define
\[
\mu^+(t)=\mathbb{E}[Y\mid Y>t],
\qquad
\mu^-(t)=\mathbb{E}[Y\mid Y<t].
\]
Then
\[
\mathrm{MAD}^+(t)=\mu^+(t)-t,
\quad
\mathrm{MAD}^-(t)=t-\mu^-(t),
\]
so
\[
\mathrm{MAD}^+(t)=\mathrm{MAD}^-(t)
\;\iff\;
t=\tfrac12\bigl(\mu^+(t)+\mu^-(t)\bigr).
\]
Hence each solution \(t\) is a \emph{balance point} where the two directional mean deviations agree.  
\begin{itemize}
  \item \emph{Symmetric laws.} If \(Y\) has a distribution symmetric about \(c\), then \(t=c\) is an equal-MAD point. For common symmetric unimodal families (Gaussian, Laplace, Logistic, uniform), this point is unique and equals \(\mathbb{E}[Y]\).
  \item \emph{Asymmetric laws.} If the law of \(Y\) is skewed, equal-MAD points need not coincide with \(\mathbb{E}[Y]\): they shift toward the heavier tail, and multimodal densities can admit multiple balance points.
\end{itemize}

\noindent\textbf{Implications for the discrepancy.} By Proposition~\ref{prop-sc},
\[
\Delta(t)=0
\quad\Longleftrightarrow\quad
t=\mathbb{E}[Y]\quad\text{or}\quad\mathrm{MAD}^+(t)=\mathrm{MAD}^-(t).
\]
Thus minimizing \(\Delta(t)\) recovers the mean when it is the unique zero of \(\Delta\), as in symmetric unimodal cases.  If additional equal-MAD zeros exist, minimization alone cannot distinguish \(\mathbb{E}[Y]\) from other balance points.

\noindent\textbf{Practical note.} In ML applications where \(t\) targets \(\mathbb{E}[Y]\), any balance point far from \(\mathbb{E}[Y]\) has negligible effect on \(\Delta(t)\).  If a non-mean balance point lies close to \(\mathbb{E}[Y]\), it is effectively indistinguishable and \(\Delta(t)\) remains a useful error proxy.  However, if \(\Delta(t)\) vanishes at a non-mean point coinciding with \(\mathbb{E}[Y]\), the estimator is suboptimal; resolving this ambiguity is an open direction for future research.

\subsection{Proof of Proposition \ref{prop-sc-C} (Calibration Identity)}
\begin{proposition}[Calibration Identity]
\label{prop-sc-C}
Let \(\mathcal D_{\mathrm C}=\{(\bm{x}_{{\mathrm C},i},\bm{y}_{{\mathrm C},i})\}_{i=1}^{|\mathcal D_{\mathrm C}|}\) be a calibration set where \(\mathcal D_{\mathrm C}\neq\mathcal D\). Then the zero-included MARs defined in (5) of the main text satisfy:
\begin{align*}
\mathrm{MAR}_{\mathrm C}(\tilde y_{{\mathrm C},k}|\bm{x})
&= \mathrm{MAR}^+_{\mathrm C}(\tilde y_{{\mathrm C},k}|\bm{x}) + \mathrm{MAR}^-_{\mathrm C}(\tilde y_{{\mathrm C},k}|\bm{x}),\\
P_{\mathrm C,k}(\bm{x})
= \tilde y_{\mathrm C,k} &+ \mathrm{MAR}^+_{\mathrm C}(\tilde y_{{\mathrm C},k}|\bm{x}) - \mathrm{MAR}^-_{\mathrm C}(\tilde y_{{\mathrm C},k}|\bm{x}).
\end{align*}
\end{proposition}

\begin{proof}
Recall the zero-included MARs from (5) in the main text,
\begin{align*}
\mathrm{MAR}_{\mathrm C}(\tilde y_{{\mathrm C},k}|\bm{x})
&= P_{{\mathrm{C}}}(\tilde y_k|\bm{x}) (1 - \tilde{y}_{{\mathrm{C}},k}) + (1 - P_{{\mathrm{C}},k}(\bm{x})) \tilde{y}_{{\mathrm{C}},k}, \\
\mathrm{MAR}^+_{\mathrm{C}}(\tilde y_{{\mathrm C},k}|\bm{x})
&= P_{\mathrm{C},k}(\bm{x}) (1 - \tilde{y}_{\mathrm{C},k}), \\
\mathrm{MAR}^-_{\mathrm{C}}(\tilde y_{{\mathrm C},k}|\bm{x})
&= (1 - P_{\mathrm{C},k}(\bm{x})) \tilde{y}_{\mathrm{C},k}.
\end{align*}
where \(  P_{\mathrm{C},k}(\bm{x}) \) denotes the conditional class probability of class \(k\) given input \(\bm{x}\), from the empirical distribution in \( \mathcal{D}_{\mathrm{C}} \).

Adding the last two equations on $\mathrm{MAR}^+_{\mathrm{C}}(\tilde y_{{\mathrm C},k}|\bm{x})$ and $\mathrm{MAR}^-_{\mathrm{C}}(\tilde y_{{\mathrm C},k}|\bm{x})$ and comparing to the definition of $\mathrm{MAR}_{\mathrm{C}}(\tilde y_{{\mathrm C},k}|\bm{x})$:
\begin{align*}
&\mathrm{MAR}^+_{\mathrm{C}}(\tilde y_{{\mathrm C},k}|\bm{x}) + \mathrm{MAR}^-_{\mathrm{C}}(\tilde y_{{\mathrm C},k}|\bm{x}) \\
&= P_{\mathrm{C},k}(\bm{x})(1 - \tilde{y}_{\mathrm{C},k}) + (1 - P_{\mathrm{C},k}(\bm{x}))\tilde{y}_{\mathrm{C},k} \nonumber \\
&= \mathrm{MAR}_{\mathrm{C}}(\tilde y_{{\mathrm C},k}|\bm{x}),
\end{align*}
which proves the first identity in the proposition.

Next, we compute:
\begin{align*}
&\tilde{y}_{\mathrm{C},k} + \mathrm{MAR}^+_{\mathrm{C}}(\tilde y_{{\mathrm C},k}|\bm{x}) - \mathrm{MAR}^-_{\mathrm{C}}(\tilde y_{{\mathrm C},k}|\bm{x}) \\
&= \tilde{y}_{\mathrm{C},k} + P_{\mathrm{C},k}(\bm{x})(1 - \tilde{y}_{\mathrm{C},k}) - (1 - P_{\mathrm{C},k}(\bm{x}))\tilde{y}_{\mathrm{C},k} \\
&= \tilde{y}_{\mathrm{C},k}(1 - P_{\mathrm{C},k}(\bm{x})) + P_{\mathrm{C},k}(\bm{x}) - (1 - P_{\mathrm{C},k}(\bm{x}))\tilde{y}_{\mathrm{C},k} \\
&= P_{\mathrm{C},k}(\bm{x}),
\end{align*}
which proves the second identity in the proposition.
\end{proof}

\section{Algorithm and Complexity Analysis}
\label{sec:training-algorithm}

This appendix provides the loss function definitions, the pseudo-code of the learning algorithms and a computational complexity analysis.

\subsection{Loss Functions and Algorithmic Pseudocode}

\subsubsection{MAR Regression}  

To estimate the expected residual magnitude, we adopt the standard mean squared error (MSE) loss. Let \(\mathcal{D} = \{(\bm{h}_i, (r_{ki})_{k=1}^K)\}_{i=1}^{|\mathcal{D}|}\) denote the training dataset, where \(r_{ki}\) represents the ground-truth MAR for class \(k\) of the \(i\)-th data point. In scalar regression datasets, this reduces to \(K = 1\). Given the UQ network \(q\) with parameters \(\Phi\), the predicted output is \(\tilde{\bm{z}}_i = q(\bm{h}_i; \Phi)\). The MSE loss is defined as:
\begin{equation}
\mathcal{L}_{\mathrm{MSE}}(\mathcal{D}; \Phi) = 
\frac{1}{|\mathcal{D}| \cdot K} \sum_{i=1}^{|\mathcal{D}|} \sum_{k=1}^{K} \left( \tilde{z}_{ki} - r_{ki} \right)^2,
\label{eq:loss-MSE}
\end{equation}

When the UQ network \(q(\bm{h}_i; \Phi)\) produces multiple outputs \(\tilde{\bm z}_i^S\), each output head is trained on its corresponding training set,  
\(\mathcal{D}^S = \{(\bm{h}_i, r_i)\}_{i=1}^{|\mathcal{D}^S|}\), where $S\in\{\,, +, -\}$. The overall MSE loss is formulated as the sum of individual loss terms over three MAR heads:

\begin{align}
\label{eq:total-MSE-loss}
\mathcal{L}_{\mathrm{MSE}}\bigl(\mathcal{D}, \mathcal{D}^+, \mathcal{D}^-; \Phi\bigr)
= &\ \mathcal{L}_{\mathrm{MSE}}\bigl(\mathcal{D}; \Phi\bigr) 
+ \mathcal{L}_{\mathrm{MSE}}\bigl(\mathcal{D}^+; \Phi\bigr) \nonumber \\
& + \mathcal{L}_{\mathrm{MSE}}\bigl(\mathcal{D}^-; \Phi\bigr).
\end{align}

\begin{algorithm}
\caption{Training Procedure for Regression}
\label{alg:regression-train}
\begin{algorithmic}[1]
\REQUIRE 
Training data  $\mathcal{D} = \{(\bm{x}_i, y_i)\}_{i=1}^{|\mathcal{D}|}$; \\
Target PI coverage level $\tau^+$ and $\tau^-$; \\
Trained model $f(\bm{x};\Theta^*) = g\bigl(h(\bm{x};\Theta^*)\bigr)$; \\
A fully connected MLP regressor with parameters $\Phi$:
\[
q\bigl(\bm h;\Phi\bigr) = \bigl({q}^{+}, {q}^{-}, {z}, {z}^{+}, {z}^{-}\bigr).
\]
Here, ${q}^{+}$ and ${q}^{-}$ are two quantile regression (QR) heads to learn $Q_{\tau^+}$ and $Q_{\tau^-}$, while ${z}$, ${z}^{+}$, and ${z}^{-}$ are MAR heads to learn $\mathrm{MAR}$, $\mathrm{MAR}^{+}$, and $\mathrm{MAR}^{-}$.
\STATE Compute feature maps and residuals: 

\(
\bm {h}_i = h(\bm{x}_i;\Theta^*), \quad r_i = y_i - g(h(\bm{x}_i;\Theta^*)), \quad \forall i \in [|\mathcal{D}|]
\)
\STATE Construct the training sets: 
\[
\begin{aligned}
    \mathcal{D}_{\mathrm{MAR}} &= \{(\bm{h}, |r|) \mid r \ne 0 \} \\
    \mathcal{D}_{\mathrm{QR}}^+ = \mathcal{D}_{\mathrm{MAR}}^+ &= \{(\bm{h}, |r|) \mid r > 0 \} \\
    \mathcal{D}_{\mathrm{QR}}^- = \mathcal{D}_{\mathrm{MAR}}^- &= \{(\bm{h}, |r|) \mid r < 0 \}
\end{aligned}
\]
\FOR{each training iteration (batch or full set)}
    \STATE Compute QR loss:
    \(\mathcal{L}_{\mathrm{QR}}\bigl(\mathcal{D}_{\mathrm{QR}}^+, \mathcal{D}_{\mathrm{QR}}^-, \tau^+, \tau^-; \Phi\bigr)\) based on \eqref{eq:loss-QR} and \eqref{eq:total-QR-loss}
    \STATE Compute MSE loss:
    \(
    \mathcal{L}_{\mathrm{MSE}}\bigl(\mathcal{D}_{\mathrm{MAR}},\mathcal{D}_{\mathrm{MAR}}^+,\mathcal{D}_{\mathrm{MAR}}^-;\Phi\bigr)
    \) based on \eqref{eq:loss-MSE} and \eqref{eq:total-MSE-loss}
    \STATE Update \(\Phi\) via gradient descent on the total loss:
    {\small
    \[
    \begin{aligned}
    \Phi \leftarrow \Phi - \eta \cdot \nabla_\Phi \Big[\, 
    & \mathcal{L}_{\mathrm{QR}}\bigl(\mathcal{D}_{\mathrm{QR}}^+, \mathcal{D}_{\mathrm{QR}}^-, \tau^+, \tau^-; \Phi\bigr) \\
    & +\, \mathcal{L}_{\mathrm{MSE}}\bigl(\mathcal{D}_{\mathrm{MAR}}, \mathcal{D}_{\mathrm{MAR}}^+, \mathcal{D}_{\mathrm{MAR}}^-; \Phi\bigr) \Big]
    \end{aligned}
    \]
    }
\ENDFOR
\RETURN Trained UQ network $q\bigl(\bm h;\Phi^*\bigr)$
\end{algorithmic}
\end{algorithm}

\begin{algorithm}
\caption{Training Procedure for Classification}
\label{alg:classification-train}
\begin{algorithmic}[1]
\REQUIRE 
Training dataset $\mathcal{D} = \{(\bm{x}_i, \bm y_i)\}_{i=1}^{|\mathcal{D}|}$, and calibration dataset $\mathcal D_{\mathrm C} = \{(\bm{x}_{{\mathrm C},i}, \bm{y}_{{\mathrm C},i})\}_{i=1}^{|\mathcal D_{\mathrm C}|}~(\mathcal D_{\mathrm C} \ne \mathcal D)$, where $\bm{y}_i \in \{0,1\}^K$ is a one-hot label; \\
Trained model $f(\bm{x};\Theta^*) = g\bigl(h(\bm{x};\Theta^*)\bigr)$; \\
A fully connected MLP regressor with parameters $\Phi_{\mathrm{T}}$:
\[
q_{\mathrm{T}}\bigl(\bm h;\Phi_{\mathrm{T}}\bigr) = \bm z.
\]
A fully connected MLP regressor with parameters $\Phi_{\mathrm{C}}$:
\[
q_{\mathrm{C}}\bigl(\bm h;\Phi_{\mathrm{C}}\bigr)
= \bigl({\bm z}_{\mathrm{C}},{\bm z}_{\mathrm{C}}^{+}, {\bm z}_{\mathrm{C}}^{-}\bigr).
\]
Here, $\bm z$ is the MAR head to learn $\mathrm{MAR}$, while ${\bm z}_{\mathrm{C}}$, ${\bm z}_{\mathrm{C}}^{+}$, and ${\bm z}_{\mathrm{C}}^{-}$ are zero-included MAR heads to learn $\mathrm{MAR}_{\mathrm C}$, $\mathrm{MAR}_{\mathrm C}^+$, and $\mathrm{MAR}_{\mathrm C}^-$.

\vspace{0.5em}
\STATE \textbf{Epistemic Uncertainty Phase (on $\mathcal{D}$):}
\STATE Compute feature maps and residuals: 

\(
\bm {h}_i = h(\bm{x}_i;\Theta^*), \quad \bm{r}_i = \bm{y}_i - g(h(\bm{x}_i;\Theta^*)), \quad \forall i \in [|\mathcal{D}|]
\)
\STATE Construct the training set: 
\(
\mathcal{D}_{\mathrm{MAR}} = \{(\bm{h}, |\bm{r}|)\} 
\)
\FOR{each training iteration  (batch or full set)}
    \STATE Compute MSE loss:
        \(
        \mathcal{L}_{\mathrm{MSE}}\bigl(\mathcal{D}_{\mathrm{MAR}};\Phi_{\mathrm{T}}\bigr)
        \) based on \eqref{eq:loss-MSE} 
    \STATE Update \(\Phi_{\mathrm{T}}\) via gradient descent on \(\mathcal{L}_{\mathrm{MAR}}\):
    \[
    \Phi_{\mathrm{T}} \leftarrow \Phi_{\mathrm{T}} - \eta \cdot \nabla_{\Phi_{\mathrm{T}}} \Bigl[\mathcal{L}_{\mathrm{MSE}}\bigl(\mathcal{D}_{\mathrm{MAR}};\Phi_{\mathrm{T}}\bigr)\Bigr]
    \]

\ENDFOR

\vspace{0.5em}
\STATE \textbf{Calibration Phase (on $\mathcal{D_C}$):}
\STATE Compute feature maps and residuals: 

\(
\bm {h}_i = h(\bm{x}_{{\mathrm C},i};\Theta^*),  \bm{r}_i = \bm{y}_{{\mathrm C},i} - g(h(\bm{x}_{{\mathrm C},i};\Theta^*)),  \forall i \in [|\mathcal{D_C}|]
\)
\STATE Construct the training sets: 
{\small \[
\begin{aligned}
    \mathcal{D}_{\mathrm{MAR}_{\mathrm C}} &= \left\{ (\bm{h}, |\bm{r}|) \right\} \\
    \mathcal{D}_{\mathrm{MAR}_{\mathrm C}}^+ &= \left\{ (\bm{h},  \max\{r_k, 0\}) \mid k = 1, \dots, K \right\} \\
    \mathcal{D}_{\mathrm{MAR}_{\mathrm C}}^- &= \left\{ (\bm{h},  -\min\{r_k, 0\}) \mid k = 1, \dots, K \right\}
\end{aligned}
\] }
\FOR{each training iteration (batch or full set)}
    \STATE Compute loss:
    \(
    \mathcal{L}_{\mathrm{MSE}}\bigl(\mathcal{D}_{\mathrm{MAR}_{\mathrm C}}, \mathcal{D}_{\mathrm{MAR}_{\mathrm C}^+},\mathcal{D}_{\mathrm{MAR}_{\mathrm C}^-};\Phi_{\mathrm{C}}\bigr)
    \) based on \eqref{eq:loss-MSE} and \eqref{eq:total-MSE-loss}
    \STATE Update \(\Phi_{\mathrm{C}}\) via gradient descent on the total loss:
    {\small
    \[
    \Phi_{\mathrm{C}} \leftarrow \Phi_{\mathrm{C}} - \eta \cdot \nabla_{\Phi_{\mathrm{C}}} \Bigl[\mathcal{L}_{\mathrm{MSE}}(\mathcal{D}_{\mathrm{MAR}_{\mathrm C}}, \mathcal{D}_{\mathrm{MAR}_{\mathrm C}^+},\mathcal{D}_{\mathrm{MAR}_{\mathrm C}^-};\Phi_{\mathrm{C}})\Bigr]
    \]
    }

\ENDFOR

\RETURN Trained UQ networks $q_{\mathrm{T}}\bigl(\bm h;\Phi_{\mathrm{T}}^*\bigr)$ and $q_{\mathrm{C}}\bigl(\bm h;\Phi_{\mathrm{C}}^*\bigr)$
\end{algorithmic}
\end{algorithm}

\subsubsection{Quantile Regression}  
  
To regress quantile bounds for prediction intervals, we adopt the calibration-aware quantile regression (QR) loss proposed in~\cite{chung2021beyond}. Let \(D = \{(\bm{h}_i, r_i)\}_{i=1}^{|D|}\) be the training data, and \(\tau \in (0,1)\) be the target quantile level. Given a regression model \(q\) with parameters \(\Phi\) that predicts \(\tilde{q}_i = q(\bm{h}_i; \Phi)\), the calibration-aware QR loss is defined as:
{\small
\begin{align}
&\mathcal{L}_{\mathrm{QR}}(D, \tau; \Phi) =\ 
\ \mathbb{I}\left\{ \hat{p}_{D} < \tau \right\} \cdot \frac{1}{|D|} \sum_{i=1}^{|D|} \left[ (y_i - \tilde{q}_i) \cdot \mathbb{I}\left\{ y_i > \tilde{q}_i \right\} \right] \notag \\
& \quad \quad + \mathbb{I}\left\{ \hat{p}_{D} > \tau \right\} \cdot \frac{1}{|D|} \sum_{i=1}^{|D|} \left[ (\tilde{q}_i - y_i) \cdot \mathbb{I}\left\{ y_i < \tilde{q}_i \right\} \right],
\label{eq:loss-QR}
\end{align}
}
where \(\tilde{p}_D\) is the  empirical coverage in $D$:
\[
\tilde{p}_D = \frac{1}{|D|} \sum_{i=1}^{|D|} \mathbb{I} \left\{ y_i \le \tilde{q}_i \right\}.
\]

This loss encourages the estimated quantile \(\tilde{q}_i\) to match the target coverage level \(\tau\) by penalizing over- or under-coverage symmetrically. Compared to traditional quantile regression losses such as the pinball loss \cite{koenker2001quantile}, the calibration-aware quantile regression loss explicitly penalizes the miscalibration of predicted quantiles, leading to improved calibration performance in practice.

In our split-point quantile regression, the UQ network \(q(\bm{h}_i; \Phi)\) produces two outputs \(\tilde{q}_i^S\), where $S \in \{+,- \}$, and each output head is trained on their corresponding training set, \(\mathcal{D}^S = \{(\bm{h}_i, r_i)\}_{i=1}^{|\mathcal{D}^S|}\), the overall QR loss is formulated as the sum of individual loss terms over each training set:
\begin{align}
&\mathcal{L}_{\mathrm{QR}}\bigl(\mathcal{D}^+, \mathcal{D}^-, \tau^+, \tau^-; \Phi\bigr) 
= \mathcal{L}_{\mathrm{QR}}\bigl(\mathcal{D}^+, \tau^+; \Phi\bigr) \notag \\
& \quad \quad \quad \quad \quad \quad + \mathcal{L}_{\mathrm{QR}}\bigl(\mathcal{D}^-, \tau^-; \Phi\bigr).
\label{eq:total-QR-loss}
\end{align}

Algorithms~\ref{alg:regression-train} and \ref{alg:classification-train} detail the training procedures for the regression and classification settings, respectively.

\subsection{Computational Complexity Analysis}

We adopt an \(L\)-layer MLP regressor as our UQ network, using the feature map \(h(\bm{x};\Theta^*)\) extracted by the base model \(f(\bm{x};\Theta^*)\). Training complexity comprises the forward–backward passes through the UQ MLP and the forward pass through the base model:
\[
\mathcal{O}\Bigl(B\sum_{i=0}^{L}h_i h_{i+1}\Bigr)
+
\mathcal{O}\Bigl(B\cdot\mathrm{Cost}\bigl[f(\bm{x};\Theta^*)\bigr]\Bigr),
\]
where \(B\) is the batch size, \(h_i\) and \(h_{i+1}\) are the input and output dimensions of layer \(i\), and \(\mathrm{Cost}[f(\bm{x};\Theta^*)]\) is the forward-pass cost of the base model. 

During inference, the cost reduces to
\[
\mathcal{O}\Bigl(B\sum_{i=0}^{L}h_i h_{i+1}\Bigr).
\]

Notably, our implementation requires only standard MLP based UQ heads, structures that are widely supported in modern deep learning frameworks, benefit from optimized low level implementations, and leverage hardware acceleration. By comparison, some deterministic single-forward-pass methods, e.g., EDL-based internal methods,  modify model layers or introduce specialized loss functions, which may lack broad hardware support and incur additional overhead. 

Moreover, our UQ heads integrate seamlessly into standard mini-batch stochastic gradient training: each step incurs a single forward pass through the shared encoder and small regression heads, an MSE evaluation against the MAR targets, and one back-propagation update. In contrast, Bayesian methods demand expensive sampling or variational approximations at each iteration, while ensemble methods multiply cost by training and storing many separate models and performing multi-pass inference.

\section{Details of Experimental Settings}
\label{sec:experimental-settings}

In this appendix, we detail the experimental settings referenced in Section VI.A of the main text to ensure completeness and facilitate reproducibility. While Table~\ref{tab:experiment-setting} summarizes the overall configuration, the following sections describe each specific setting and implementation detail.

\begin{table*}
\caption{Overall Experimental Configuration}
\label{tab:experiment-setting}
\centering
\scalebox{1}{
\begin{tabular}{|l|l|l|l|l|}
\hline
Settings                                                                  & Datasets                            & Baseline Methods                                                                                                                                                      & Metrics                                                                            & Model Architectures                                                 \\ \hline
\multirow{3}{*}{Regression}                                               & Cubic Regression                         & DE, EDL, SQR-OC                                                                                                                                                       & \begin{tabular}[c]{@{}l@{}}Predicted vs. True Plot,\\ RMSE, PIECE\end{tabular} & MLP                                                         \\ \cline{2-5} 
                                                                          & UCI Regression Datasets             & MD, DE, EDL, SQR-OC                                                                                                                                                   & \begin{tabular}[c]{@{}l@{}}RMSE, Winkler Score,\\ PIECE, Correlation\end{tabular} & MLP                                                         \\ \cline{2-5} 
                                                                          & Monocular Depth Estimation Datasets & MD, DE, EDL, SQR-OC                                                                                                                                                   & \begin{tabular}[c]{@{}l@{}}RMSE, Winkler Score,\\ PIECE, AUROC\end{tabular}       & CNN                                                         \\ \hline
\multirow{2}{*}{Classification}                                           & Image Classification Datasets       & LA, DE, EDL, OC, DDU, TS                                                                                                                                              & \begin{tabular}[c]{@{}l@{}}Accuracy, ECE,\\ AUROC\end{tabular}                     & CNN                                                         \\ \cline{2-5} 
                                                                          & Multimodal Classification Datasets  & MD, DE, EDL                                                                                                                                                           & Accuracy, AUROC                                                                    & \begin{tabular}[c]{@{}l@{}}CNN, \\ Transformer\end{tabular} \\ \hline
\end{tabular}
}
\end{table*}

\subsection{Datasets}
\label{sec:datasets}
This section presents comprehensive information regarding the benchmark datasets used in our experiments.

\subsubsection{Cubic Regression}
\label{sec:datasets-cubic}
For intuitive illustration of regression setting, we follow the setup in~\cite{hernandez2015probabilistic, Lakshminarayanan2016SimpleAS, amini2020deep} to construct a synthetic cubic regression task with zero-mean, asymmetric log-normal noise. The training samples are drawn from the function:
\[
y = x^3 + \epsilon(x) - \mathbb{E}[\epsilon(x)], \quad \epsilon(x) \sim \mathrm{LogNormal}(1.5, 1).
\]
Where the input \( x \) is sampled uniformly from the range \( [-4, 4] \). The test set is similarly constructed, with inputs sampled from a broader range \( [-6, 6] \). The training set contains 2,000 data points, while the test set consists of 1,000 samples. We define the interval \( [-4, 4] \) as the in-distribution (iD) region, while the regions outside this interval, i.e., \( [-6, -4) \cup (4, 6] \), are considered as out-of-distribution (OOD) region. 

In addition, we also include two alternative noise distributions to assess the robustness of UQ methods in capturing uncertainty under various label noise :
\begin{itemize}
    \item A skewed trimodal Gaussian mixture:
    \[
    \epsilon(x) \sim 0.4 \cdot \mathcal{N}(0, 1) + 0.3 \cdot \mathcal{N}(40, 1) + 0.3 \cdot \mathcal{N}(-10, 1),
    \]

    \item A high-variance Gaussian distribution:
    \(
    \epsilon(x) \sim \mathcal{N}(0, 8).
    \)
\end{itemize}

\subsubsection{UCI Regression Benchmarks}
\label{sec:datasets-uci}
For standard scalar regression tasks, we follow the settings in~\cite{hernandez2015probabilistic, Lakshminarayanan2016SimpleAS, amini2020deep} and evaluate our method on nine widely used UCI regression benchmarks~\cite{UCI2025}. Since no official data splits are provided, each dataset is randomly divided into training and testing sets using a 9:1 ratio over 20 independent trials to ensure statistical robustness. Key dataset statistics, including the number of samples (\( N \)), input dimensionality (\( d \)), train/test split ratio, and the number of trials, are summarized in Table~\ref{tab:UCI-dataset}.

To further evaluate robustness under label noise, we inject synthetic noise into the regression targets of the UCI benchmarks. Specifically, target values in each dataset are first normalized, after which two types of asymmetric noise distributions are introduced. Consistent with the cubic regression setup, we adopt the following noise distributions:
\begin{itemize}
    \item Asymmetric log-normal noise:
    \[
    \epsilon(x) \sim \mathrm{LogNormal}(1, 0.5),
    \]
    
    \item Skewed trimodal Gaussian mixture:
    \[
    \epsilon(x) \sim 0.3 \cdot \mathcal{N}(-1, 0.1) + 0.4 \cdot \mathcal{N}(0, 0.1) + 0.3 \cdot \mathcal{N}(3, 0.1),
    \]
\end{itemize}
These additional settings simulate highly non-Gaussian noise distributions and provide a challenging testbed for evaluating the quality and robustness of uncertainty estimates.

\begin{table}[h]
\caption{Characteristics of UCI Regression Datasets}
\label{tab:UCI-dataset}
\centering
\scalebox{1}{
\begin{tabular}{lllcc}
\hline
Dataset                       & $N$   & $d$ & Split Ratio & Trails \\ \hline
Boston Housing                & 506   & 13  & 9:1         & 20     \\
Concrete Compression Strength & 1030  & 8   & 9:1         & 20     \\
Energy Efficiency             & 768   & 8   & 9:1         & 20     \\
Kin8nm                        & 8192  & 8   & 9:1         & 20     \\
Naval Propulsion              & 11934 & 16  & 9:1         & 20     \\
Combined Cycle Power Plant    & 9568  & 4   & 9:1         & 20     \\
Protein Structure             & 45730 & 9   & 9:1         & 20     \\
Wine Quality Red              & 1599  & 11  & 9:1         & 20     \\
Yacht Hydrodynamics           & 308   & 6   & 9:1         & 20     \\ \hline
\end{tabular}
}
\end{table}

\subsubsection{Monocular Depth Estimation Datasets}
\label{sec:datasets-depth}
To evaluate our method on high-dimensional, multi-output regression tasks, we follow the setting in~\cite{amini2020deep} and adopt monocular image-based end-to-end depth estimation as a benchmark. Specifically, we train our model on the NYU Depth V2 dataset~\cite{silberman2012indoor}, which consists of over 27,000 RGB-to-depth image pairs (\(128 \times 160\)) captured in indoor environments. The dataset is randomly split into training, validation, and test subsets with an 80-10-10 ratio, ensuring no overlap in scene scans.

To assess OOD detection, we use the ApolloScape dataset \cite{huang2018apolloscape}, which contains outdoor driving scenes. We randomly sample 1,000 images from ApolloScape as the OOD data set.

To evaluate the model’s fine-grained uncertainty estimation, we also generate adversarial variants using the Fast Gradient Sign Method (FGSM)~\cite{goodfellow2014explaining}. FGSM perturbs the input in the direction of the gradient of the loss function:
\[
\bm{x}^{\text{adv}} = \bm{x} + \epsilon \cdot \text{sign}(\nabla_{\bm{x}} \mathcal{L}(\bm{x}, y)),
\]
where \( \mathcal{L} \) denotes the loss function, \( (\bm{x}, y) \) is the input-label pair, and \( \epsilon \) controls the perturbation strength. The function \( \text{sign}(\cdot) \) denotes the element-wise sign operation, indicating the direction of the input gradient.

In our experiments, we incrementally vary \( \epsilon \) from 0 to 0.2 with a step size of 0.025 to simulate adversarial perturbations of increasing strength.

\subsubsection{Image Classification Datasets}
\label{sec:datasets-cls}
For classification tasks, we first follow the evaluation protocol of~\cite{mukhoti2023deep, schweighofer2023quantification} and assess model performance across a variety of iD vs. adversarial and iD vs. OOD dataset pairs. We conduct experiments on CIFAR-10, CIFAR-100~\cite{Krizhevsky2009}, and ImageNet-1K~\cite{Deng2009}, which represent classification benchmarks of increasing scale and complexity.

For the small-scale datasets, CIFAR-10 and CIFAR-100, we generate adversarial examples from the test set using FGSM with \( \epsilon=0.02 \), and randomly collect 10,000 samples from SVHN~\cite{Netzer2011} and Tiny ImageNet~\cite{le2015tiny} as the OOD evaluation set. For the large-scale ImageNet-1K dataset, we adopt ImageNet-A and ImageNet-O~\cite{Hendrycks2019} as sources of adversarial and OOD examples. Table~\ref{tab:cls-dataset} summarizes the dataset configurations, including the number of samples (\( N \)), input dimensionality (\( d \)), and number of classes (\( K \)) in each setting.

Additionally, for CIFAR-10 and CIFAR-100, we also design dedicated test sets to evaluate robustness under increasing levels of adversarial perturbation, by varying the FGSM strength \( \epsilon \) from 0 to 0.4 with a step size of 0.04.

\begin{table}
\caption{Characteristics of Image Classification Datasets}
\label{tab:cls-dataset}
\centering
\scalebox{0.8}{
\begin{tabular}{ccc|c|c}
\hline
\multicolumn{3}{c|}{iD   datasets}                            & Adversarial datasets & OOD datasets         \\ \hline
\multicolumn{3}{c|}{CIFAR-10}                                 & FGSM on CIFAR-10     & SVHN + Tiny ImageNet \\
\(N_{\text{train}/\text{test}}\) & \(d\)              & \(K\) & \(N\)                & \(N\)                \\
60000 / 10000                    & \(32 \times 32\)   & 10    & 10000                & 10000                \\ \hline
\multicolumn{3}{c|}{CIFAR-100}                                & FGSM on CIFAR-100    & SVHN + Tiny ImageNet \\
\(N_{\text{train}/\text{test}}\) & \(d\)              & \(K\) & \(N\)                & \(N\)                \\
60000 / 10000                    & \(32 \times 32\)   & 100   & 10000                & 10000                \\ \hline
\multicolumn{3}{c|}{ImageNet-1K}                              & ImageNet-A           & ImageNet-O           \\
\(N_{\text{train}/\text{test}}\) & \(d\)              & \(K\) & \(N\)                & \(N\)                \\
1.28M / 50000                    & \(224 \times 224\) & 1000  & 7500                 & 2000                 \\ \hline
\end{tabular}
}
\end{table}

\subsubsection{Multimodal Classification Dataset}
For multimodal classification, we utilize the LUMA benchmark~\cite{luma_dataset2025}, which comprises audio, image, and textual modalities spanning 50 distinct classes. The image modality is sourced from the CIFAR-10 and CIFAR-100 datasets~\cite{Krizhevsky2009}, the audio samples are collected from three diverse audio corpora, and the textual modality is generated using a large language model.

The dataset contains 600 examples per class (500 for training and 100 for testing) for 42 in-distribution classes, along with 3,859 OOD samples drawn from the remaining 8 classes. In addition, the LUMA benchmark provides a Python toolkit for generating datasets with controllable levels of noise and uncertainty. This uncertainty generator enables the systematic manipulation of aleatoric uncertainty in the input data and epistemic uncertainty in the model predictions.

\subsection{Baselines}
We compare against representative UQ baseline methods spanning four categories:  

\subsubsection{Bayesian-based methods} 
In our experiments, we include two widely adopted Bayesian methods to quantify epistemic uncertainty: \textit{MC-Dropout} (MD)~\cite{gal2016dropout} and \textit{Laplace Approximation} (LA)~\cite{daxberger2021laplace}.

\textbf{MC-Dropout (MD)} performs approximate Bayesian inference by applying dropout at both training and test time. Let \( f_{\theta}(\bm{x}) \) denote the network output given input \( \bm{x} \) and weights \( \theta \). At inference time, the model performs \( T \) stochastic forward passes, producing \( \{ f_{\theta_t}(\bm{x}) \}_{t=1}^T \), where each \( \theta_t \sim q(\theta) \) corresponds to a different dropout mask. The predictive mean and epistemic uncertainty can be estimated as:
\begin{align*}
\mathbb{E}[f(\bm{x})] &\approx \frac{1}{T} \sum_{t=1}^T f_{\theta_t}(\bm{x}), \\
\text{Var}[f(\bm{x})] &\approx \frac{1}{T} \sum_{t=1}^T f_{\theta_t}(\bm{x})^2 - \left( \mathbb{E}[f(\bm{x})] \right)^2.
\end{align*}

\textbf{Laplace Approximation (LA)} approximates the posterior distribution \( p(\theta \mid \mathcal{D}) \) with a Gaussian centered at the maximum a posteriori (MAP) estimate \( \theta_{\text{MAP}} \). Specifically, it expands the negative log-likelihood \( \mathcal{L} \) around \( \theta_{\text{MAP}} \) and uses the inverse Hessian as the covariance:
\[
p(\theta \mid \mathcal{D}) \approx \mathcal{N}(\theta_{\text{MAP}}, \Sigma), \quad \text{where } \Sigma^{-1} = \nabla^2 \mathcal{L}(\theta_{\text{MAP}}).
\]
This local Gaussian approximation enables efficient epistemic uncertainty estimation through \( p(\theta \mid \mathcal{D}) \). In deep learning, LA provides a tractable way to perform approximate Bayesian inference with minimal changes to the training pipeline.

\subsubsection{Ensemble-based Methods}
In our experiments, we adopt \textit{Deep Ensemble} (DE)~\cite{Lakshminarayanan2016SimpleAS}, a widely-used and effective baseline.

\textbf{Deep Ensemble (DE)} constructs an ensemble of \( M \) neural networks \( \{ f_{\theta_m} \}_{m=1}^M \), each trained independently with different random initializations and data shuffling. Given an input \( \bm{x} \), the ensemble prediction and its uncertainty are estimated as:
\begin{align*}
\mathbb{E}[f(\bm{x})] &\approx \frac{1}{M} \sum_{m=1}^M f_{\theta_m}(\bm{x}), \label{eq:ens-mean} \\
\text{Var}[f(\bm{x})] &\approx \frac{1}{M} \sum_{m=1}^M f_{\theta_m}(\bm{x})^2 - \left( \mathbb{E}[f(\bm{x})] \right)^2.
\end{align*}

\subsubsection{Internal deterministic single forward-pass methods}
This category primarily includes evidential methods that aim to quantify both aleatoric and epistemic uncertainty within a single deterministic forward pass. 
As each method is tailored to a specific task type, we adopt the following representatives in our experiments: \textit{Evidential Regression} (EDL-R)~\cite{amini2020deep}, \textit{Evidential Quantile Regression} (EDL-QR)~\cite{huttel2023deep}, and \textit{Evidential Classification} (EDL-C)~\cite{sensoy2018evidential}. 
In addition, for regression baselines that do not explicitly model aleatoric uncertainty, we apply \textit{Gaussian likelihood regression} to approximate the data distribution and construct prediction intervals accordingly.

\textbf{Evidential Regression} (EDL-R)~\cite{amini2020deep} models the target \( y \) using a Normal-Inverse-Gamma (NIG) distribution over the Gaussian parameters:
\[
p(y \mid \bm{x}) = \int \mathcal{N}(y \mid \mu, \sigma^2) \cdot \text{NIG}(\mu, \sigma^2 \mid \gamma, \nu, \alpha, \beta) \, d\mu \, d\sigma^2.
\]
Here, the network outputs the NIG parameters \( \gamma \) (mean), \( \nu \) (strength of belief in \( \gamma \)), \( \alpha \), and \( \beta \) (shape and scale of inverse-Gamma for \( \sigma^2 \)).

From this distribution, the predictive mean and total variance are:
\[
\mathbb{E}[y] = \gamma, \quad
\text{Var}[y] = \underbrace{\frac{\beta}{\alpha - 1}}_{\text{Aleatoric}} + \underbrace{\frac{\beta}{\nu (\alpha - 1)}}_{\text{Epistemic}}, \quad \text{for } \alpha > 1.
\]

\textbf{Evidential Quantile Regression} (EDL-QR) extends evidential learning to the quantile regression setting by modeling uncertainty in estimating a specific quantile \( \tau \in (0, 1) \) of the target distribution. The predictive likelihood is assumed to follow an asymmetric Laplace distribution (ALD), parameterized by location \( \mu \), scale \( \sigma \), and asymmetry \( \tau \), i.e.,
\[
p(y \mid \mu, \sigma, \tau) = \frac{\tau(1 - \tau)}{\sigma} \exp\left(-\rho_{\tau}\left(\frac{y - \mu}{\sigma}\right)\right),
\]
where \( \rho_{\tau}(u) = u (\tau - \mathbb{I}_{\{u < 0\}}) \) is the check loss function.

Analogous to EDL-R. EDL-QR learns evidential parameters \( \gamma, \nu, \alpha, \beta \) for each target quantile \( \tau \) by modeling the quantile prediction using a Student's \( t \)-distribution. the predictive quantile \( \hat{q}_{\tau} \) is given by \( \gamma \), and the total predictive uncertainty is:
\[
\text{Aleatoric} = \frac{\beta}{\alpha - 1}, \quad
\text{Epistemic} = \frac{\beta}{\nu(\alpha - 1)}, \quad \text{for } \alpha > 1.
\]

\textbf{Evidential Classification} (EDL-C) models class probabilities via a Dirichlet distribution:
\[
p(\mathbf{p} \mid \bm{x}) = \text{Dir}(\mathbf{p} \mid \boldsymbol{\alpha}), \quad \boldsymbol{\alpha} = \mathbf{e} + 1,
\]
where \( \mathbf{e} \in \mathbb{R}_+^K \) is the evidence output for each of the \( K \) classes. The expected predictive probability is:
\[
\mathbb{E}[p_k] = \frac{\alpha_k}{S}, \quad \text{where } S = \sum_{k=1}^{K} \alpha_k.
\]

Aleatoric uncertainty is captured by the entropy of the Dirichlet distribution, while epistemic uncertainty is inversely proportional to the total evidence mass \( S \), and is typically quantified as \( K/S \).

\textbf{Gaussian Likelihood Regression} is a common method for modeling both the predictive mean and data uncertainty in regression tasks. 
It assumes the target variable follows a Gaussian distribution with input-dependent mean \( \mu(\bm{x}) \) and variance \( \sigma^2(\bm{x}) \), and trains the model by maximizing the Gaussian log-likelihood:

\[
\mathcal{L}_{\text{Gaussian}} = \sum_{i} \left[ \frac{1}{2} \log \sigma^2(\bm{x}_i) + \frac{(y_i - \mu(\bm{x}_i))^2}{2\sigma^2(\bm{x}_i)} \right],
\]

where \( \mu(\bm{x}_i) \) is the predicted mean and \( \sigma^2(\bm{x}_i) \) is the predicted variance for input \( \bm{x}_i \).
The learned variance \( \sigma^2(\bm{x}) \) captures the aleatoric uncertainty, and prediction intervals can be constructed under the Gaussian assumption, e.g., \( \mu(\bm{x}) \pm 2\sigma(\bm{x}) \) for 95\% confidence.

\subsubsection{External Deterministic Single-Forward-Pass Methods}
We include two representative methods: \textit{SQR-OC}~\cite{tagasovska2019single}, and \textit{DDU}~\cite{mukhoti2023deep}, which estimates epistemic uncertainty by modeling feature space via a deterministic deep model.
In addition, for classification baseline methods that do not explicitly quantify aleatoric uncertainty, we apply \textit{Temperature Scaling} (TS)~\cite{guo2017calibration} as a post-hoc calibration technique to adjust the softmax logits and improve confidence reliability.

\textbf{SQR-OC} estimates aleatoric uncertainty in regression by Simultaneous Quantile Regression (SQR), a loss function to learn all the conditional quantiles of a given target variable. 
Given a set of quantile levels \( \tau \in (0, 1) \), the model learns to predict the corresponding quantile values \( q_\tau(x) \) by minimizing the quantile regression loss:
\[
\begin{aligned}
\mathcal{L}_{\text{SQR}} &= \sum_{i} \sum_{\tau \in \mathcal{T}} \rho_\tau \left( y_i - q_\tau(x_i) \right), \\
\text{where} \quad \rho_\tau(r) &= \max(\tau r, (\tau - 1) r).
\end{aligned}
\]
and \( \mathcal{T} \) is the set of quantile levels (e.g., \( \{0.025, 0.5, 0.975\} \)). 
The predicted quantiles can be used to construct prediction intervas (PIs), such as the 95\% PI: \( [q_{0.025}(x), q_{0.975}(x)] \). 

To capture epistemic uncertainty in regression and classification, SQR-OC introduces Orthonormal Certificates (OC), which uses a lightweight auxiliary module that maps penultimate-layer features \(h(\bm{x})\) to a low-dimensional subspace spanned by orthonormal vectors \( \{\mathbf{v}_i\}_{i=1}^m \), trained to produce near-zero output on the training data:
\[
C(h(\bm{x})) = V^\top h(\bm{x}), \quad \text{where } V = [\mathbf{v}_1, \dots, \mathbf{v}_m], \quad V^\top V = I.
\]
During inference, the certificate score \( C(h(\bm{x}))\) is used as an epistemic uncertainty measure as larger values indicate deviation from the training feature manifold.

\textbf{Deep Deterministic Uncertainty} (DDU) estimates epistemic uncertainty by modeling the distribution of penultimate-layer features using a Gaussian Mixture Model (GMM). During training, feature vectors \( h(\bm{x}) \) are extracted from the penultimate layer and used to fit a GMM with \( K \) components, one for each class:
\[
p(h(\bm{x})) = \sum_{k=1}^{K} \pi_k \, \mathcal{N}(h(\bm{x}) \mid \boldsymbol{\mu}_k, \Sigma_k),
\]
where \( \pi_k \), \( \boldsymbol{\mu}_k \), and \( \Sigma_k \) are the mixture weight, mean, and covariance of class \( k \), respectively.

At inference time, the model computes the log-likelihood of a test sample’s feature under the fitted GMM. The epistemic uncertainty is defined as the negative log-likelihood:
\[
\text{Uncertainty}(\bm{x}) = -\log p(h(\bm{x})).
\]
Lower likelihood indicates the feature lies far from the learned feature distribution, suggesting high epistemic uncertainty.

\textbf{Temperature Scaling} (TS) is a simple and widely used post-hoc calibration method for classification models. It adjusts the softmax logits by a scalar temperature parameter \( T > 0 \) to smooth or sharpen predicted probabilities:
\[
\hat{p}_k = \frac{\exp(z_k / T)}{\sum_{j=1}^K \exp(z_j / T)},
\]
where \( \mathbf{z} = (z_1, \dots, z_K) \) is the uncalibrated logit vector for an input, and \( \hat{p}_k \) is the calibrated probability for class \( k \).

The optimal temperature \( T^* \) is obtained by minimizing the negative log-likelihood (NLL) on a held-out validation set:
\[
T^* = \arg\min_T \sum_{i} -\log \hat{p}_{y_i}^{(i)}.
\]

\subsection{Base Models}
For cubic regression, we train a multilayer perceptron (MLP) with two hidden layers of 64 neurons each.

For UCI regression datasets, we train a smaller MLP architecture with one hidden layer of 50 neurons to reflect standard experimental settings in prior work.

For monocular depth estimation, we train a U-Net~\cite{ronneberger2015u} to extract spatial features.

In image classification tasks, we evaluate performance using two convolutional architectures: VGG-16~\cite{Simonyan2015} and Wide-ResNet~\cite{zagoruyko2016wide}. On CIFAR-10 and CIFAR-100, all base models are trained from scratch, whereas for ImageNet-1K, we adopt pretrained models from the \texttt{torchvision} library~\cite{torchvision2024models}.

For the multimodal classification task, we follow the LUMA benchmark~\cite{luma_dataset2025} and adopt the customized convolutional neural networks (CNNs) defined therein to encode the visual and audio modalities, while employing a Transformer-based encoder (BERT~\cite{devlin2019bert}) for the text modality.

For external UQ methods such as OC, DDU, and our proposed method, the base models are trained solely using the original task loss, and their output structures remain unchanged. In contrast, other UQ methods reuse the base model's backbone but modify the loss function or output layers to suit their specific designs.

As shown in Table~\ref{tab:basemodel-summary}, we summarize the base model settings adopted for each experimental task.

\begin{table}
\caption{Summary of Base Model Settings}
\label{tab:basemodel-summary}
\centering
\scalebox{1}{
\begin{tabular}{l|l}
\hline
Task                                                                          & Base Model                                                                            \\ \hline
Cubic Regression                                                              & \begin{tabular}[c]{@{}l@{}}MLP\\ (2 hidden layers, 64 neurons each)\end{tabular}      \\ \hline
UCI Benchmarks                                                                & \begin{tabular}[c]{@{}l@{}}MLP\\ (1 hidden layer, 50 neurons)\end{tabular}            \\ \hline
Monocular Depth Estimation                                                    & U-Net (trained from scratch)                                                                      \\ \hline
\begin{tabular}[c]{@{}l@{}}Image Classification\\ (CIFAR-10/100)\end{tabular} & \begin{tabular}[c]{@{}l@{}}VGG-16 / Wide-ResNet\\ (trained from scratch)\end{tabular} \\ \hline
\begin{tabular}[c]{@{}l@{}}Image Classification\\ (ImageNet-1K)\end{tabular}  & \begin{tabular}[c]{@{}l@{}}VGG-16 / Wide-ResNet\\ (pretrained)\end{tabular}           \\ \hline
Multimodal Classification                                                     & \begin{tabular}[c]{@{}l@{}}CNNs (visual/audio modality)\\ BERT (text modality)\end{tabular}             \\ \hline
\end{tabular}
}
\end{table}

\subsection{Experimental Protocol}
\label{tab:experimental-protocol}
We evaluate all models under identical settings, including the same training, validation and test splits, and a consistent hyperparameter search.

\subsubsection{Aleatoric Uncertainty Protocol}
For regression, MD and DE use Gaussian-likelihood regression, while other methods rely on their own evidential or quantile-based distributions to construct 95\% prediction intervals (PIs). Specifically, PIs are defined as \( \mu \pm 2\sigma \) for Gaussian models, the 2.5th to 97.5th percentiles for QR-based models, and symmetric intervals covering 95\% of samples for the split-point analysis. Point predictions are defined as the predictive mean in Gaussianbased models, the 50th percentile in quantile regression, and the MSE-optimal output in our framework. Since all compared methods produce both point predictions and PIs, they can be jointly evaluated from a split-point perspective. 

For classification tasks, we apply post-hoc softmax calibration to baseline methods that retain softmax outputs, such as DDU, OC, and DE. Calibration is performed using a held-out calibration set, sampled from the training data with a proportion of 10\%. The optimal temperature is selected via grid search over the range \([0, 10]\) with a step size of 0.1.

\subsubsection{Epistemic Uncertainty Protocol}
Although different UQ methods quantify epistemic uncertainty through fundamentally different mechanisms, and their uncertainty scores may vary in scale or range, we do not normalize or rescale the outputs. Instead, we focus on the trend and relative correlation of epistemic uncertainty with other uncertainty signals (e.g., error or OOD samples), making the raw scores directly comparable in our evaluations.

\subsubsection{Hyperparameter Tuning}
We randomly reserve 10\% of the training data as a validation set and perform $k$-fold cross validation, with $k = 20$ for synthetic, UCI datasets and CIFAR-10/CIFAR-100, and $k = 5$ for the remaining datasets due to computational constraints. To simulate calibration data, we further sample 10\% of the training set without data leakage. All models are evaluated on the predefined test sets.

\subsubsection{Other Training Protocols}
\label{tab:other-protocol}
For our method, we further evaluate four different training protocols:  
(1) joint training of the base model and the UQ network from scratch;  
(2) stagewise training with varying amounts of training data to examine robustness to data volume, using subsets ranging from 20\% to 100\% with a step size of 20\%;  
(3) evaluation of our proposed calibration under different training/calibration set splits, using a held-out calibration set ranging from 10\% to 50\% of the training data;
(4) performance of the UQ network under MLP architectures with 1, 2, and 3 hidden layers.

\subsection{Evaluation Criteria}
To comprehensively evaluate the performance, efficiency, and practicality of our UQ framework, we conduct evaluation from four perspectives.

\subsubsection{Learning Task Performance}
We use the \textit{Root Mean Squared Error (RMSE)} for point estimation in regression, and \textit{accuracy} for classification. They are computed as:
\begin{align*}
\text{RMSE} &= \sqrt{\frac{1}{N} \sum_{i=1}^{N} (y_i - \tilde{y}_i)^2}, \\
\text{Accuracy} &= \frac{1}{N} \sum_{i=1}^{N} \mathbb{I}(\tilde{y}_i = y_i),
\end{align*}
where \( \tilde{y}_i \) is the prediction, and \( y_i \) is the ground truth.

\subsubsection{Aleatoric Uncertainty}
In regression, we first adopt the \textit{Prediction Interval Expected Calibration Error} (PIECE) \cite{levi2022evaluating, zadrozny2001obtaining}, which evaluates calibration error across different ranges of PI width.
Formally, we partition the prediction set into \( M \) disjoint bins based on the PI width.
Given a PI \( [l_i, u_i] \) at confidence level \( 1 - \alpha \), and let \( \mathcal{B}_m \) denote the set of indices in the \( m \)-th bin. 
The PIECE is defined as:
\[
\text{PIECE} = \sum_{m=1}^{M} \frac{|\mathcal{B}_m|}{N} \cdot \left| \frac{1}{|\mathcal{B}_m|} \sum_{i \in \mathcal{B}_m} \mathbb{I} \left[ y_i \in [l_i, u_i] \right] - (1 - \alpha) \right|,
\]
where \( N \) is the total number of samples. 

Furthermore, motivated by our SPA, we also adopt fine-grained split-point metrics $\text{PIECE}^{+}$ and $\text{PIECE}^{-}$ on the upper and lower split point intervals, respectively. This decomposition measures overestimation and underestimation separately and applies to any model yielding point predictions:
\[
\begin{aligned}
\text{PIECE}^{+} &= \left| \frac{1}{|\mathcal{R}^{+}|} \sum_{r_i \in \mathcal{R}^{+}} \mathbb{I}\left( |r_i| \le \left(u_i - \hat{y}_i\right) \right) - \tau^{+} \right|, \\
\text{PIECE}^{-} &= \left| \frac{1}{|\mathcal{R}^{-}|} \sum_{r_i \in \mathcal{R}^{-}} \mathbb{I}\left( |r_i| \le \left(\hat{y}_i - l_i\right) \right) - \tau^{-} \right|,
\end{aligned}
\]
where \(u_i\) and \(l_i\) denote the predicted upper and lower quantile bounds for sample \(i\), and \(\hat{y}_i\) is the point prediction. \(\mathcal{R}^+\) and \(\mathcal{R}^-\) represent the subsets of residuals drfined in Section III-B1 of the main text, respectively. \(\tau^+\) and \(\tau^-\) are the target one-sided coverage levels for the upper and lower bounds.

In addition, to jointly evaluate the calibration and sharpness of PIs, we adopt the \textit{Winkler Score}~\cite{winkler1972decision}. Given a PI \( [l_i, u_i] \) at confidence level \( 1 - \alpha \), the Winkler score is computed as:
\begin{align*}
\text{Winkler}_i =
\begin{cases}
u_i - l_i, & \text{if } y_i \in [l_i, u_i], \\
(u_i - l_i) + \frac{2}{\alpha}(l_i - y_i), & \text{if } y_i < l_i, \\
(u_i - l_i) + \frac{2}{\alpha}(y_i - u_i), & \text{if } y_i > u_i,
\end{cases}
\end{align*}
and the average Winkler score is \( \frac{1}{N} \sum_{i=1}^N \text{Winkler}_i \). Lower scores indicate better PI quality.

For classification, we use \textit{Expected Calibration Error (ECE)}
\cite{naeini2015obtaining} to assess the alignment between predicted confidence and empirical accuracy:
\[
\text{ECE} = \sum_{b=1}^{M} \frac{|\mathcal{B}_m|}{N} \left| \text{Acc}(\mathcal{B}_m) - \text{Conf}(\mathcal{B}_m) \right|,
\]
where the predictions are grouped into \( M \) bins, \(|\mathcal{B}_m| \) is the number of samples in bin \( \mathcal{B}_m \), \( \text{Acc}(\mathcal{B}_m) \) is the accuracy, and \( \text{Conf}(\mathcal{B}_m) \) is the average confidence in that bin.

\subsubsection{Epistemic Uncertainty}
Since epistemic uncertainty is expected to correlate with model errors, we evaluate it in regression by computing the \textit{Spearman rank correlation coefficient}~\cite{spearman1961proof} between absolute prediction errors \( |y_i - \tilde{y}_i| \) and epistemic uncertainty scores \( u_i \), defined as:
\[
\rho = 1 - \frac{6 \sum_{i=1}^{N} (r_i - s_i)^2}{N(N^2 - 1)},
\]
where \( r_i \) and \( s_i \) are the ranks of \( |y_i - \tilde{y}_i| \) and \( u_i \), respectively.

In monocular depth estimation and classification tasks, we introduce adversarial and OOD samples, which are inherently uncertain from the model's perspective. Therefore, the ability to detect such samples using epistemic uncertainty estimates can serve as an indirect measure of epistemic uncertainty quality.
To evaluate this uncertainty-based detection, we use the \textit{Area Under the Receiver Operating Characteristic Curve (AUROC)}. Given uncertainty scores \( u_i \) and binary labels \( z_i \in \{0,1\} \) (e.g., iD vs. OOD), AUROC measures the probability that a randomly chosen positive sample has a higher uncertainty score than a randomly chosen negative one:
\[
\text{AUROC} = \mathbb{P}(u^+ > u^-), \quad u^+ \sim \mathcal{U}_1, \, u^- \sim \mathcal{U}_0.
\]
Higher AUROC values indicate better discrimination ability.

\subsubsection{Efficiency}
We compare the training-time per epoch and inference-time per batch during both training and evaluation to assess the computational efficiency of each method. Since the trends are consistent across all experiments, we report the results only on the depth estimation task for brevity.

\subsection{Implementation}
\subsubsection{Software and Hardware}
All benchmark implementations (except LUMA) are developed in Python 3.8 with PyTorch 2.1.2. For the LUMA benchmark, we follow the official requirements and use Python 3.9 with PyTorch 2.3.0. 

Most evaluations can be conducted on an NVIDIA V100 GPU with 16~GB of memory. For ImageNet-1K, which requires higher memory capacity, evaluations are performed on an NVIDIA A100 GPU with 80~GB of memory.

\subsubsection{Hyperparameters and Optimization}
This section outlines the hyperparameter settings and optimization configurations used for all methods across different experimental setups.

\paragraph{Implementation on Cubic Regression}
For the synthetic regression function, we do not use mini-batch training due to the small dataset size. Instead, the entire training set is used for full-batch updates. All models are trained using the Adam optimizer with a learning rate of 0.001 for 5,000 epochs.

For the UQ-related hyperparameters, \textit{Deep Ensemble} is constructed by training 5 independently initialized models. The \textit{Evidential} regression model is trained with a regularization coefficient of \( \lambda = 0.01 \). \textit{SQR-OC} is implemented with a certificate layer of size \( k = 20 \) and trained for 10 epochs. 
Our method attaches an MLP head with a hidden layer of 64 units to the final hidden layer of the base model, and is trained using the same optimizer and learning rate as the base model.

\paragraph{Implementation on UCI Regression Datasets}
For all UCI regression datasets, base models are trained using the Adam optimizer with a learning rate of \( 1 \times 10^{-4} \) and a batch size of 64 for 400 epochs.

Regarding UQ-related hyperparameters, \textit{MC-Dropout} uses a dropout rate of 0.2 and performs 5 stochastic forward passes at inference. \textit{Deep Ensemble} consists of 5 independently trained models. The evidential regression model is trained with a regularization coefficient of \( \lambda = 0.01 \). \textit{SQR-OC} employs a certificate layer of size \( k = 100 \) and is trained for 10 epochs.
Our method appends an MLP head with a hidden layer of 50 units to the final hidden layer of the base model and is trained using the same optimizer and learning rate as the base model.

\paragraph{Implementation on Monocular Depth Estimation}
For the monocular depth regression task, all models are trained using the Adam optimizer with a learning rate of \( 5 \times 10^{-5} \), a batch size of 32, and for 60,000 iterations. Each model is independently trained 5 times from random initialization to ensure robustness and to report averaged results.

Regarding UQ-related hyperparameters, \textit{MC-Dropout} uses a dropout rate of 0.1 and performs 5 stochastic forward passes at inference. \textit{Deep Ensemble} consists of 5 independently trained models. The evidential regression model is trained with a regularization coefficient of \( \lambda = 0.1 \). \textit{SQR-OC} employs a certificate layer of size \( k = 100 \) and is trained for 10 epochs.
Our method appends an MLP head with a hidden layer of 32 units to the final hidden layer of the base model and is trained using the same optimizer and learning rate as the base model.

\paragraph{Implementation on Image Classification}
For CIFAR-10/100, models are trained for 350 epochs using stochastic gradient descent (SGD) with a momentum of 0.9 and an initial learning rate of 0.1. The learning rate is decayed by a factor of 10 at epochs 150 and 250. Each model is independently trained 25 times from different initializations. For ImageNet-1K, we adopt pretrained models from the \texttt{torchvision} library~\cite{torchvision2024models} and perform calibration without updating the feature extractor.

Regarding UQ-related hyperparameters, \textit{Laplace Approximation} is applied using the default settings from~\cite{daxberger2021laplace}. \textit{Deep Ensemble} consists of 5 independently trained models. The evidential classification model is trained with a regularization coefficient of \( \lambda = 0.0001 \). Although evidential classification is not inherently post-hoc, we simulate a post-hoc setup on ImageNet-1K by freezing the feature extractor and training the fully connected layers with evidential loss. \textit{SQR-OC} employs a certificate layer of size \( k = 100 \) and is trained for 20 epochs. 
Our method appends a 3-layer MLP head whose hidden layer matches the size of the final hidden layer of the base model. It is trained using the Adam optimizer with a learning rate of \( 1 \times 10^{-4} \) for 300 epochs on CIFAR-10/100, and a learning rate of \( 1 \times 10^{-5} \) for 100 epochs on ImageNet-1K.

\paragraph{Implementation on Multimodal Classification}
For the multimodal benchmark LUMA, we follow the official training protocol from~\cite{luma_dataset2025}. All models are trained for up to 300 epochs, with early stopping applied if the validation loss does not improve for 10 consecutive epochs. The initial learning rate is 0.001 and is reduced by a factor of 0.33 if no improvement is observed in the validation loss after 5 epochs.

Regarding UQ-related hyperparameters, our method appends a 3-layer MLP head, with hidden dimensions matching the size of the final hidden layer of the base model. It is trained using the same optimizer and learning rate as the base model.

\textbf{Our source code is available at \url{https://github.com/zzz0527/SPC-UQ}.}

\section{Additional Experimental Results}
\label{sec:additional-results}

Beyond the experimental results reported in Sections VI.B and VI.C of the main text, we further evaluate our UQ framework’s robustness by injecting synthetic label noise and adversarial input perturbations into existing datasets. These experiments measure the stability of various UQ methods under different types of uncertainty.

\subsection{Robustness to Label Noise}
To assess the robustness of UQ methods in capturing uncertainty under various forms of label noise, we construct synthetic datasets by injecting controlled noise into the regression targets. The details of the noise injection procedure are provided in Appendix~\ref{sec:datasets-cubic} and Appendix~\ref{sec:datasets-uci}.

\subsubsection{Illustration and Results on Cubic Regression}

\begin{figure*}
  \centering
  \includegraphics[width=1\linewidth]{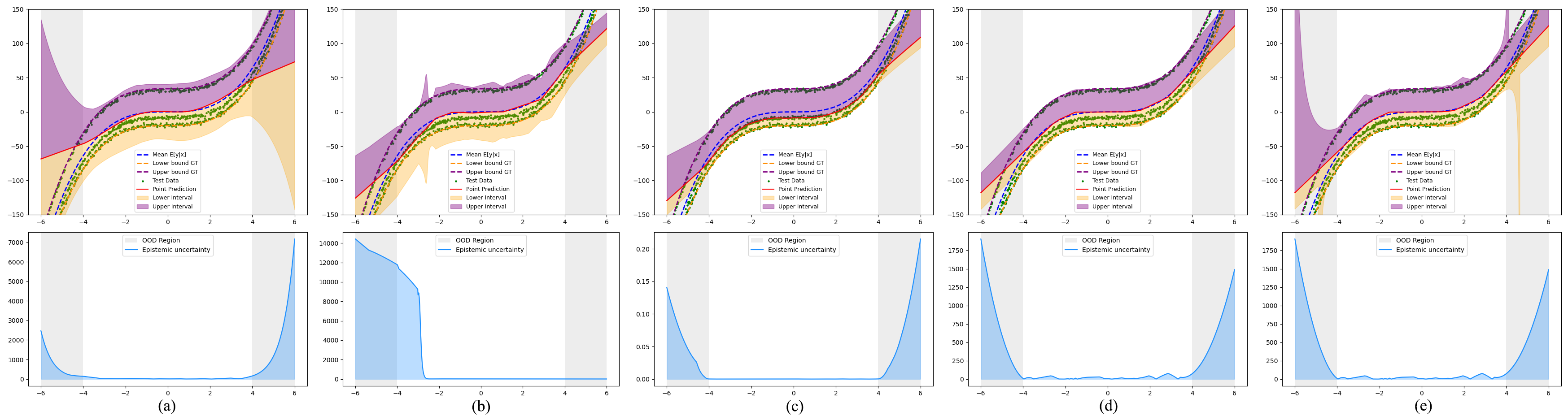}
  \caption{Uncertainties quantified for the cubic regression with trimodal noise using (a) Deep Ensemble (DE), (b) Evidential Regression (EDL-R), (c) SQR-OC, (d) our method without calibration, and (e) our method with calibration. \textbf{Top row} shows aleatoric uncertainty estimates, \textbf{bottom row} shows epistemic uncertainty estimates. Ground truth and true PI boundaries are shown as dashed lines.}
  \label{fig:unc-decomposition2}
\end{figure*}

\begin{figure*}
  \centering
  \includegraphics[width=1\linewidth]{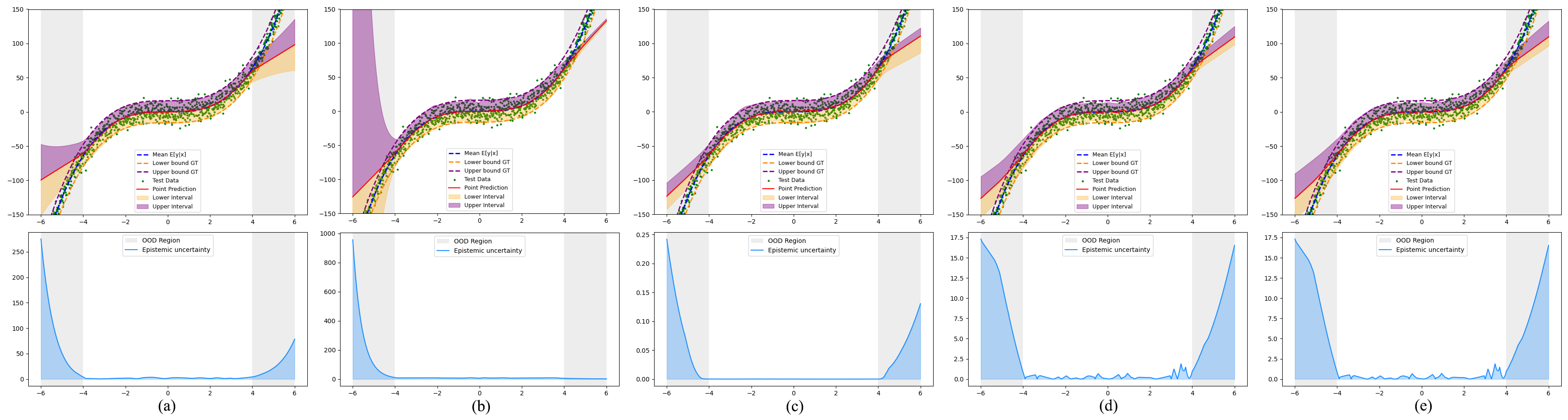}
  \caption{Uncertainties quantified for the cubic regression with high-variance Gaussian noise using (a) Deep Ensemble (DE), (b) Evidential Regression (EDL-R), (c) SQR-OC, (d) our method without calibration, and (e) our method with calibration. \textbf{Top row} shows aleatoric uncertainty estimates, \textbf{bottom row} shows epistemic uncertainty estimates. Ground truth and true PI boundaries are shown as dashed lines.}
  \label{fig:unc-decomposition3}
\end{figure*}

Figure~\ref{fig:unc-decomposition2} presents the results under the skewed trimodal Gaussian mixture noise setting. Although DE and EDL-R, both based on Gaussian assumptions, accurately fit the predictive mean, their PIs exhibit clear mismatches with the true distribution. Moreover, the multimodal nature of the noise severely impairs EDL-R’s ability to estimate epistemic uncertainty. Quantile regression-based methods show noticeable deviation between the predicted median and the true distribution mean. In contrast, our method accurately captures both the predictive mean and interval boundaries. Furthermore, the epistemic uncertainty score derived from self-consistency discrepancy score (SDS) effectively distinguishes the iD and OOD regions.

Finally, we compare the calibration results. The initial PIs are already well-aligned with the empirical distribution. After applying the calibration procedure, the PI boundaries are slightly expanded in some iD regions, while the overall sharpness of the intervals remains preserved.

Figure~\ref{fig:unc-decomposition3} illustrates the results under the high-variance Gaussian noise setting. Under this symmetric and unimodal distribution, all UQ methods demonstrate accurate point predictions and well-calibrated PIs. In terms of epistemic uncertainty, most methods except for EDL-R show clear capability in distinguishing iD and OOD regions.

These results confirm that many UQ methods are inherently well-suited for settings aligned with Gaussian assumptions, yielding strong performance under such conditions. However, their performance may degrade when applied to non-Gaussian distributions.

\begin{table}[h]
\caption{Quantitative Analysis of Cubic Regression under Different Noise Settings}
\label{tab:cubic-results}
\centering
\scalebox{0.8}{
\begin{tabular}{l|llll|llll}
\hline
\multirow{2}{*}{Method} & \multicolumn{4}{c|}{Trimodal Noise}                                                    & \multicolumn{4}{c}{Gaussian Noise}                                                    \\ \cline{2-9} 
                        & RMSE                 & PICP                & $\text{PICP}^+$     & $\text{PICP}^-$     & RMSE                & PICP                & $\text{PICP}^+$     & $\text{PICP}^-$     \\ \hline
DE                      & \underline{\textit{21.08}} & 0.05                & 0.05                & 0.05                & 8.50                & \underline{\textbf{0.00}} & \underline{\textbf{0.01}} & \underline{\textbf{0.00}} \\
EDL-R                   & 21.39                & 0.03                & \underline{\textit{0.02}} & 0.05                & \underline{\textbf{8.48}} & 0.01                & 0.02                & \underline{\textit{0.01}} \\
SQR-OC                  & 22.73                & \underline{\textit{0.02}} & \underline{\textit{0.02}} & \underline{\textbf{0.01}} & 8.52                & 0.02                & \underline{\textbf{0.01}} & 0.02                \\
Ours                    & \underline{\textbf{20.86}} & \underline{\textbf{0.01}} & \underline{\textit{0.01}} & \underline{\textit{0.02}} & \underline{\textbf{8.48}} & 0.02                & 0.02                & 0.02                \\
Ours-Calib              & \underline{\textbf{20.86}} & \underline{\textbf{0.01}} & \underline{\textbf{0.00}} & \underline{\textit{0.02}} & \underline{\textbf{8.48}} & \underline{\textit{0.01}} & \underline{\textbf{0.01}} & \underline{\textit{0.01}} \\ \hline
\end{tabular}
}
\end{table}

To provide a quantitative analysis of the cubic regression results, Table~\ref{tab:cubic-results} reports the RMSE and split-point PIECEs for each method. Our method achieves the best point prediction accuracy under both noise settings and ranks among the top two in terms of PI quality after applying our calibration procedure. Notably, this experiment also highlights the value of our split-point metrics \(\text{PIECE}^{+}\) and \(\text{PIECE}^{-}\), which offer a more fine-grained evaluation of PI quality. For instance, under asymmetric noise, distribution mismatch can lead to overly wide intervals on one side, which may not heavily affect the total PIECE score but significantly increases the our proposed directional PIECE scores. This underscores the necessity of split-point PIECEs in practical settings.

\subsubsection{Results on UCI Regression Datasets}

As shown in Table~\ref{tab:UCI-results2} and Table~\ref{tab:UCI-results3}, most methods achieve comparable scores under the total \textsc{PIECE} metric. However, the presence of asymmetric noise substantially disrupts calibration balance for methods that rely on symmetric Gaussian assumptions when modeling data uncertainty. In particular, MD, DE, and EDL-R exhibit noticeably higher \(\text{PIECE}^{+}\) or \(\text{PIECE}^{-}\) values compared to quantile regression-based methods.

In contrast, our method consistently delivers stable and reliable uncertainty estimates across both noise scenarios. It achieves the best point prediction performance (measured by RMSE) and ranks among the top two methods in PI calibration metrics, including Winkler Score and split-point PIECEs, across all datasets. Furthermore, on half of the benchmarks, our method demonstrates the strongest correlation between prediction error and estimated uncertainty. This suggests that while our epistemic uncertainty estimates may be less precise than those of evidential methods under strong distributional assumptions, the distribution-agnostic nature of our method enables superior robustness under non-standard or complex noise distributions.

\begin{table*}
\caption{Results on UCI Regression Benchmarks with Log-normal Noise}
\label{tab:UCI-results2}
\centering
\scalebox{0.88}{
\begin{tabular}{cllllllllll}
\hline
\multicolumn{1}{l}{\multirow{2}{*}{Metric}}                              & \multirow{2}{*}{Method}                                                      & \multicolumn{9}{c}{Dataset}                                                                                                                                                                                                                                              \\ \cline{3-11} 
\multicolumn{1}{l}{}                                                     &                                                                              & Boston                      & Concrete                    & Energy                      & Kin8nm                     & Naval                      & Power                       & Protein                     & Wine                       & Yacht                       \\ \hline
\multirow{6}{*}{RMSE}                                                    & MD                                                                           & 3.68 ± 0.10                 & 6.56 ± 0.07                 & 3.18 ± 0.04                 & 0.13 ± 0.00                & \underline{\textbf{0.00 ± 0.00}} & 4.83 ± 0.03                 & 4.76 ± 0.01                 & 0.63 ± 0.00                & 3.01 ± 0.15                 \\
                                                                         & DE                                                                           & \underline{\textit{3.39 ± 0.10}}  & \underline{\textit{6.02 ± 0.07}}  & \underline{\textit{2.78 ± 0.04}}  & \underline{\textbf{0.09 ± 0.00}} & \underline{\textbf{0.00 ± 0.00}} & \underline{\textbf{4.60 ± 0.03}}  & \underline{\textit{4.50 ± 0.01}}  & \underline{\textit{0.62 ± 0.00}} & \underline{\textit{2.54 ± 0.08}}  \\
                                                                         & EDL-R                                                                        & 3.55 ± 0.11                 & 6.13 ± 0.07                 & 2.85 ± 0.05                 & \underline{\textit{0.10 ± 0.00}} & \underline{\textbf{0.00 ± 0.00}} & 4.61 ± 0.03                 & 4.67 ± 0.01                 & 0.63 ± 0.00                & 2.59 ± 0.10                 \\
                                                                         & EDL-QR                                                                       & 3.63 ± 0.11                 & 6.33 ± 0.08                 & 2.96 ± 0.04                 & \underline{\textit{0.10 ± 0.00}} & \underline{\textbf{0.00 ± 0.00}} & 4.64 ± 0.03                 & 4.62 ± 0.01                 & 0.63 ± 0.00                & 2.88 ± 0.12                 \\
                                                                         & SQR-OC                                                                       & 3.52 ± 0.10                 & 6.21 ± 0.07                 & 2.73 ± 0.04                 & \underline{\textbf{0.09 ± 0.00}} & \underline{\textbf{0.00 ± 0.00}} & 4.64 ± 0.03                 & 4.59 ± 0.01                 & 0.63 ± 0.00                & 2.66 ± 0.11                 \\
                                                                         & Ours                                                                         & \underline{\textbf{3.06 ± 0.06}}  & \underline{\textbf{5.69 ± 0.06}}  & \underline{\textbf{2.25 ± 0.04}}  & \underline{\textbf{0.09 ± 0.00}} & \underline{\textbf{0.00 ± 0.00}} & \underline{\textbf{4.60 ± 0.02}}  & \underline{\textbf{4.39 ± 0.01}}  & \underline{\textbf{0.61 ± 0.00}} & \underline{\textbf{2.52 ± 0.08}}  \\ \hline
\multirow{7}{*}{\begin{tabular}[c]{@{}c@{}}Winkler\\ Score\end{tabular}} & MD                                                                           & 16.66 ± 0.42                & 30.00 ± 0.30                & 12.77 ± 0.15                & 0.57 ± 0.00                & \underline{\textbf{0.01 ± 0.00}} & 24.42 ± 0.18                & 21.94 ± 0.06                & 2.99 ± 0.02                & 16.60 ± 0.62                \\
                                                                         & DE                                                                           & 15.06 ± 0.48                & \underline{\textit{26.57 ± 0.41}} & 11.36 ± 0.22                & \underline{\textit{0.40 ± 0.00}} & \underline{\textbf{0.01 ± 0.00}} & 23.26 ± 0.18                & 20.78 ± 0.06                & \underline{\textit{2.89 ± 0.03}} & \underline{\textit{14.21 ± 0.75}} \\
                                                                         & EDL-R                                                                        & 16.50 ± 0.62                & 27.93 ± 0.52                & 12.02 ± 0.28                & 0.43 ± 0.00                & \underline{\textbf{0.01 ± 0.00}} & 23.44 ± 0.18                & 22.57 ± 0.12                & 3.12 ± 0.02                & 14.83 ± 0.76                \\
                                                                         & EDL-QR                                                                       & 17.24 ± 0.62                & 29.46 ± 0.48                & 11.62 ± 0.19                & 0.45 ± 0.00                & \underline{\textbf{0.01 ± 0.00}} & 23.39 ± 0.16                & 18.34 ± 0.04                & 2.98 ± 0.02                & 16.28 ± 0.66                \\
                                                                         & SQR-OC                                                                       & 16.86 ± 0.43                & 28.83 ± 0.37                & \underline{\textit{10.97 ± 0.19}} & 0.42 ± 0.00                & \underline{\textbf{0.01 ± 0.00}} & \underline{\textit{23.12 ± 0.17}} & \underline{\textit{17.95 ± 0.04}} & 2.89 ± 0.02                & 15.76 ± 0.61                \\
                                                                         & \multirow{2}{*}{\begin{tabular}[c]{@{}l@{}}Ours\\ Ours-Calib\end{tabular}} & \underline{\textbf{14.27 ± 0.31}} & \underline{\textbf{24.90 ± 0.41}} & \underline{\textbf{9.44 ± 0.16}}  & \underline{\textbf{0.38 ± 0.00}} & \underline{\textbf{0.01 ± 0.00}} & \underline{\textbf{22.61 ± 0.17}} & \underline{\textbf{16.69 ± 0.03}} & \underline{\textbf{2.75 ± 0.02}} & \underline{\textbf{13.98 ± 0.65}} \\
                                                                         &                                                                              & \underline{\textit{14.93 ± 0.47}} & \underline{\textit{25.07 ± 0.31}} & \underline{\textit{10.35 ± 0.53}} & \underline{\textit{0.39 ± 0.00}} & \underline{\textbf{0.01 ± 0.00}} & \underline{\textbf{22.61 ± 0.16}} & 19.47 ± 0.47                & \underline{\textit{2.88 ± 0.04}} & 14.36 ± 0.64                \\ \hline
\multirow{7}{*}{PIECE}                                                   & MD                                                                           & \underline{\textbf{0.03 ± 0.00}}  & \underline{\textit{0.03 ± 0.00}}  & \underline{\textit{0.04 ± 0.00}}  & 0.03 ± 0.00                & 0.03 ± 0.00                & 0.02 ± 0.00                 & 0.02 ± 0.00                 & \underline{\textbf{0.01 ± 0.00}} & \underline{\textbf{0.04 ± 0.00}}  \\
                                                                         & DE                                                                           & 0.04 ± 0.00                 & \underline{\textit{0.03 ± 0.00}}  & \underline{\textit{0.04 ± 0.00}}  & \underline{\textbf{0.01 ± 0.00}} & \underline{\textbf{0.01 ± 0.00}} & \underline{\textbf{0.01 ± 0.00}}  & 0.02 ± 0.00                 & 0.02 ± 0.00                & 0.06 ± 0.00                 \\
                                                                         & EDL-R                                                                        & 0.04 ± 0.00                 & \underline{\textit{0.03 ± 0.00}}  & \underline{\textbf{0.03 ± 0.00}}  & \underline{\textbf{0.01 ± 0.00}} & \underline{\textbf{0.01 ± 0.00}} & \underline{\textbf{0.01 ± 0.00}}  & \underline{\textbf{0.01 ± 0.00}}  & 0.02 ± 0.00                & 0.06 ± 0.01                 \\
                                                                         & EDL-QR                                                                       & 0.04 ± 0.00                 & \underline{\textit{0.03 ± 0.00}}  & \underline{\textit{0.04 ± 0.00}}  & \underline{\textbf{0.01 ± 0.00}} & 0.02 ± 0.00                & \underline{\textbf{0.01 ± 0.00}}  & \underline{\textbf{0.01 ± 0.00}}  & 0.02 ± 0.00                & 0.07 ± 0.00                 \\
                                                                         & SQR-OC                                                                       & 0.04 ± 0.00                 & \underline{\textit{0.03 ± 0.00}}  & \underline{\textit{0.04 ± 0.00}}  & \underline{\textbf{0.01 ± 0.00}} & 0.02 ± 0.00                & \underline{\textbf{0.01 ± 0.00}}  & \underline{\textbf{0.01 ± 0.00}}  & 0.02 ± 0.00                & 0.07 ± 0.00                 \\
                                                                         & \multirow{2}{*}{\begin{tabular}[c]{@{}l@{}}Ours\\ Ours-Calib\end{tabular}} & \underline{\textbf{0.03 ± 0.00}}  & \underline{\textit{0.03 ± 0.00}}  & \underline{\textbf{0.03 ± 0.00}}  & \underline{\textbf{0.01 ± 0.00}} & \underline{\textbf{0.01 ± 0.00}} & \underline{\textbf{0.01 ± 0.00}}  & \underline{\textbf{0.01 ± 0.00}}  & 0.02 ± 0.00                & 0.06 ± 0.00                 \\
                                                                         &                                                                              & \underline{\textbf{0.03 ± 0.00}}  & \underline{\textbf{0.02 ± 0.00}}  & \underline{\textbf{0.03 ± 0.00}}  & \underline{\textbf{0.01 ± 0.00}} & \underline{\textbf{0.01 ± 0.00}} & \underline{\textbf{0.01 ± 0.00}}  & \underline{\textbf{0.01 ± 0.00}}  & \underline{\textbf{0.01 ± 0.00}} & \underline{\textit{0.05 ± 0.00}}  \\ \hline
\multirow{7}{*}{$\text{PIECE}^+$}                                        & MD                                                                           & \underline{\textbf{0.02 ± 0.00}}  & 0.02 ± 0.00                 & \underline{\textbf{0.01 ± 0.00}}  & 0.03 ± 0.00                & \underline{\textbf{0.01 ± 0.00}} & \underline{\textit{0.01 ± 0.00}}  & 0.02 ± 0.00                 & \underline{\textbf{0.01 ± 0.00}} & \underline{\textbf{0.03 ± 0.01}}  \\
                                                                         & DE                                                                           & 0.05 ± 0.01                 & \underline{\textbf{0.01 ± 0.00}}  & 0.03 ± 0.00                 & \underline{\textbf{0.00 ± 0.00}} & 0.04 ± 0.00                & \underline{\textbf{0.00 ± 0.00}}  & 0.01 ± 0.00                 & 0.02 ± 0.00                & 0.08 ± 0.01                 \\
                                                                         & EDL-R                                                                        & 0.05 ± 0.01                 & 0.02 ± 0.00                 & 0.03 ± 0.00                 & \underline{\textit{0.01 ± 0.00}} & 0.04 ± 0.00                & \underline{\textit{0.01 ± 0.00}}  & 0.02 ± 0.00                 & 0.03 ± 0.00                & 0.08 ± 0.01                 \\
                                                                         & EDL-QR                                                                       & \underline{\textbf{0.02 ± 0.00}}  & \underline{\textbf{0.01 ± 0.00}}  & \underline{\textit{0.02 ± 0.00}}  & \underline{\textit{0.01 ± 0.00}} & \underline{\textbf{0.01 ± 0.00}} & \underline{\textit{0.01 ± 0.00}}  & \underline{\textbf{0.00 ± 0.00}}  & \underline{\textbf{0.01 ± 0.00}} & 0.04 ± 0.01                 \\
                                                                         & SQR-OC                                                                       & \underline{\textbf{0.02 ± 0.00}}  & 0.02 ± 0.00                 & \underline{\textit{0.02 ± 0.00}}  & \underline{\textit{0.01 ± 0.00}} & \underline{\textbf{0.01 ± 0.00}} & \underline{\textit{0.01 ± 0.00}}  & \underline{\textbf{0.00 ± 0.00}}  & 0.02 ± 0.00                & \underline{\textbf{0.03 ± 0.00}}  \\
                                                                         & \multirow{2}{*}{\begin{tabular}[c]{@{}l@{}}Ours\\ Ours-Calib\end{tabular}} & \underline{\textbf{0.02 ± 0.00}}  & \underline{\textbf{0.01 ± 0.00}}  & \underline{\textit{0.02 ± 0.00}}  & \underline{\textit{0.01 ± 0.00}} & \underline{\textbf{0.01 ± 0.00}} & \underline{\textit{0.01 ± 0.00}}  & 0.01 ± 0.00                 & \underline{\textbf{0.01 ± 0.00}} & 0.04 ± 0.00                 \\
                                                                         &                                                                              & \underline{\textbf{0.02 ± 0.00}}  & 0.02 ± 0.00                 & \underline{\textit{0.02 ± 0.00}}  & \underline{\textit{0.01 ± 0.00}} & \underline{\textbf{0.01 ± 0.00}} & \underline{\textit{0.01 ± 0.00}}  & 0.01 ± 0.00                 & \underline{\textbf{0.01 ± 0.00}} & \underline{\textbf{0.03 ± 0.00}}  \\ \hline
\multirow{7}{*}{$\text{PIECE}^-$}                                        & MD                                                                           & 0.03 ± 0.00                 & 0.03 ± 0.00                 & 0.05 ± 0.00                 & 0.03 ± 0.00                & 0.05 ± 0.00                & 0.02 ± 0.00                 & 0.04 ± 0.00                 & \underline{\textbf{0.01 ± 0.00}} & 0.05 ± 0.00                 \\
                                                                         & DE                                                                           & \underline{\textbf{0.02 ± 0.00}}  & \underline{\textbf{0.01 ± 0.00}}  & 0.04 ± 0.00                 & 0.02 ± 0.00                & 0.05 ± 0.00                & 0.02 ± 0.00                 & 0.04 ± 0.00                 & \underline{\textbf{0.01 ± 0.00}} & 0.03 ± 0.00                 \\
                                                                         & EDL-R                                                                        & \underline{\textbf{0.02 ± 0.00}}  & \underline{\textbf{0.01 ± 0.00}}  & 0.04 ± 0.00                 & \underline{\textbf{0.00 ± 0.00}} & 0.04 ± 0.00                & \underline{\textit{0.01 ± 0.00}}  & 0.04 ± 0.00                 & \underline{\textbf{0.01 ± 0.00}} & 0.03 ± 0.00                 \\
                                                                         & EDL-QR                                                                       & 0.03 ± 0.01                 & \underline{\textbf{0.01 ± 0.00}}  & \underline{\textbf{0.02 ± 0.00}}  & \underline{\textit{0.01 ± 0.00}} & 0.03 ± 0.00                & 0.02 ± 0.00                 & 0.02 ± 0.00                 & \underline{\textbf{0.01 ± 0.00}} & \underline{\textbf{0.02 ± 0.00}}  \\
                                                                         & SQR-OC                                                                       & 0.03 ± 0.01                 & 0.02 ± 0.00                 & \underline{\textbf{0.02 ± 0.00}}  & \underline{\textit{0.01 ± 0.00}} & \underline{\textit{0.02 ± 0.00}} & \underline{\textit{0.01 ± 0.00}}  & \underline{\textbf{0.00 ± 0.00}}  & 0.02 ± 0.00                & 0.04 ± 0.01                 \\
                                                                         & \multirow{2}{*}{\begin{tabular}[c]{@{}l@{}}Ours\\ Ours-Calib\end{tabular}} & \underline{\textbf{0.02 ± 0.01}}  & 0.02 ± 0.00                 & \underline{\textbf{0.02 ± 0.00}}  & \underline{\textit{0.01 ± 0.00}} & \underline{\textbf{0.01 ± 0.00}} & \underline{\textbf{0.00 ± 0.00}}  & \underline{\textit{0.01 ± 0.00}}  & \underline{\textbf{0.01 ± 0.00}} & 0.03 ± 0.00                 \\
                                                                         &                                                                              & \underline{\textbf{0.02 ± 0.00}}  & 0.02 ± 0.00                 & \underline{\textbf{0.02 ± 0.00}}  & \underline{\textit{0.01 ± 0.00}} & \underline{\textbf{0.01 ± 0.00}} & \underline{\textbf{0.00 ± 0.00}}  & \underline{\textit{0.01 ± 0.00}}  & \underline{\textbf{0.01 ± 0.00}} & \underline{\textbf{0.02 ± 0.00}}  \\ \hline
\multirow{6}{*}{Correlation}                                             & MD                                                                           & \underline{\textbf{0.34 ± 0.01}}  & 0.31 ± 0.01                 & 0.45 ± 0.01                 & \underline{\textit{0.32 ± 0.01}} & \underline{\textbf{0.22 ± 0.01}} & 0.09 ± 0.00                 & 0.42 ± 0.00                 & 0.25 ± 0.01                & \underline{\textbf{0.29 ± 0.03}}  \\
                                                                         & DE                                                                           & 0.31 ± 0.01                 & \underline{\textbf{0.43 ± 0.01}}  & \underline{\textbf{0.55 ± 0.01}}  & 0.30 ± 0.00                & 0.14 ± 0.01                & 0.16 ± 0.00                 & \underline{\textit{0.51 ± 0.00}}  & 0.33 ± 0.01                & 0.19 ± 0.03                 \\
                                                                         & EDL-R                                                                        & 0.30 ± 0.01                 & \underline{\textit{0.42 ± 0.01}}  & \underline{\textbf{0.55 ± 0.01}}  & \underline{\textbf{0.33 ± 0.01}} & -0.06 ± 0.01               & \underline{\textit{0.18 ± 0.00}}  & -0.13 ± 0.01                & \underline{\textit{0.35 ± 0.01}} & 0.14 ± 0.03                 \\
                                                                         & EDL-QR                                                                       & 0.31 ± 0.01                 & 0.40 ± 0.01                 & \underline{\textit{0.54 ± 0.01}}  & 0.31 ± 0.01                & \underline{\textbf{0.22 ± 0.02}} & \underline{\textit{0.18 ± 0.00}}  & 0.50 ± 0.00                 & 0.30 ± 0.01                & 0.21 ± 0.03                 \\
                                                                         & SQR-OC                                                                       & 0.26 ± 0.01                 & 0.30 ± 0.01                 & 0.48 ± 0.01                 & 0.24 ± 0.01                & 0.11 ± 0.01                & 0.13 ± 0.00                 & 0.32 ± 0.01                 & 0.22 ± 0.01                & \underline{\textit{0.22 ± 0.03}}  \\
                                                                         & Ours                                                                         & \underline{\textit{0.32 ± 0.01}}  & 0.39 ± 0.02                 & 0.50 ± 0.01                 & 0.25 ± 0.01                & \underline{\textit{0.17 ± 0.01}} & \underline{\textbf{0.23 ± 0.01}}  & \underline{\textbf{0.59 ± 0.00}}  & \underline{\textbf{0.42 ± 0.01}} & 0.18 ± 0.03                 \\ \hline
\end{tabular}
}
\end{table*}

\begin{table*}
\caption{Results on UCI Regression Benchmarks with Trimodal Noise}
\label{tab:UCI-results3}
\centering
\scalebox{0.88}{
\begin{tabular}{cllllllllll}
\hline
\multicolumn{1}{l}{\multirow{2}{*}{Metric}}                              & \multirow{2}{*}{Method}                                                      & \multicolumn{9}{c}{Dataset}                                                                                                                                                                                                                                              \\ \cline{3-11} 
\multicolumn{1}{l}{}                                                     &                                                                              & Boston                      & Concrete                    & Energy                      & Kin8nm                     & Naval                      & Power                       & Protein                     & Wine                       & Yacht                       \\ \hline
\multirow{6}{*}{RMSE}                                                    & MD                                                                           & 4.31 ± 0.09                 & 7.66 ± 0.09                 & 3.91 ± 0.04                 & 0.14 ± 0.00                & \underline{\textbf{0.00 ± 0.00}} & 6.26 ± 0.02                 & 4.96 ± 0.01                 & 0.65 ± 0.00                & 4.72 ± 0.08                 \\
                                                                         & DE                                                                           & \underline{\textit{4.10 ± 0.08}}  & \underline{\textit{7.18 ± 0.08}}  & \underline{\textit{3.61 ± 0.04}}  & \underline{\textbf{0.11 ± 0.00}} & \underline{\textbf{0.00 ± 0.00}} & \underline{\textbf{6.11 ± 0.02}}  & \underline{\textit{4.69 ± 0.01}}  & \underline{\textit{0.64 ± 0.00}} & \underline{\textbf{4.46 ± 0.07}}  \\
                                                                         & EDL-R                                                                        & 4.21 ± 0.09                 & 7.29 ± 0.08                 & 3.66 ± 0.04                 & \underline{\textbf{0.11 ± 0.00}} & \underline{\textbf{0.00 ± 0.00}} & 6.12 ± 0.02                 & 4.82 ± 0.01                 & \underline{\textit{0.64 ± 0.00}} & 4.50 ± 0.07                 \\
                                                                         & EDL-QR                                                                       & 4.32 ± 0.10                 & 7.50 ± 0.08                 & 3.76 ± 0.04                 & \underline{\textbf{0.11 ± 0.00}} & \underline{\textbf{0.00 ± 0.00}} & 6.16 ± 0.02                 & 4.79 ± 0.01                 & 0.65 ± 0.00                & 4.72 ± 0.08                 \\
                                                                         & SQR-OC                                                                       & 4.22 ± 0.09                 & 7.40 ± 0.08                 & 3.63 ± 0.04                 & \underline{\textbf{0.11 ± 0.00}} & \underline{\textbf{0.00 ± 0.00}} & 6.16 ± 0.02                 & 4.78 ± 0.01                 & 0.65 ± 0.00                & 4.63 ± 0.08                 \\
                                                                         & Ours                                                                         & \underline{\textbf{3.91 ± 0.06}}  & \underline{\textbf{7.08 ± 0.08}}  & \underline{\textbf{3.40 ± 0.03}}  & \underline{\textbf{0.11 ± 0.00}} & \underline{\textbf{0.00 ± 0.00}} & \underline{\textbf{6.11 ± 0.02}}  & \underline{\textbf{4.64 ± 0.01}}  & \underline{\textbf{0.63 ± 0.00}} & \underline{\textbf{4.46 ± 0.07}}  \\ \hline
\multirow{7}{*}{\begin{tabular}[c]{@{}c@{}}Winkler\\ Score\end{tabular}} & MD                                                                           & 20.48 ± 0.51                & 36.27 ± 0.41                & 17.03 ± 0.13                & 0.64 ± 0.00                & \underline{\textbf{0.01 ± 0.00}} & 30.82 ± 0.13                & 22.66 ± 0.04                & 3.08 ± 0.02                & 23.30 ± 0.31                \\
                                                                         & DE                                                                           & \underline{\textit{19.55 ± 0.50}} & \underline{\textit{33.92 ± 0.47}} & 16.16 ± 0.14                & \underline{\textit{0.50 ± 0.00}} & \underline{\textbf{0.01 ± 0.00}} & 30.11 ± 0.15                & 21.56 ± 0.05                & \underline{\textit{2.98 ± 0.03}} & 23.39 ± 0.43                \\
                                                                         & EDL-R                                                                        & 20.58 ± 0.62                & 35.20 ± 0.50                & 16.77 ± 0.17                & 0.52 ± 0.00                & \underline{\textbf{0.01 ± 0.00}} & 30.38 ± 0.17                & 22.70 ± 0.07                & 3.10 ± 0.02                & 25.67 ± 0.52                \\
                                                                         & EDL-QR                                                                       & 20.34 ± 0.53                & 36.60 ± 0.45                & 15.87 ± 0.16                & 0.54 ± 0.00                & \underline{\textbf{0.01 ± 0.00}} & 29.64 ± 0.14                & 19.82 ± 0.03                & 3.09 ± 0.02                & 23.06 ± 0.45                \\
                                                                         & SQR-OC                                                                       & 20.32 ± 0.48                & 35.73 ± 0.50                & \underline{\textit{15.59 ± 0.18}} & 0.51 ± 0.00                & \underline{\textbf{0.01 ± 0.00}} & \underline{\textit{29.40 ± 0.17}} & \underline{\textit{19.63 ± 0.02}} & 3.04 ± 0.03                & \underline{\textit{21.11 ± 0.32}} \\
                                                                         & \multirow{2}{*}{\begin{tabular}[c]{@{}l@{}}Ours\\ Ours-Calib\end{tabular}} & \underline{\textbf{18.25 ± 0.35}} & \underline{\textit{33.10 ± 0.43}} & \underline{\textbf{14.35 ± 0.15}} & \underline{\textbf{0.49 ± 0.00}} & \underline{\textbf{0.01 ± 0.00}} & \underline{\textbf{29.10 ± 0.14}} & \underline{\textbf{18.54 ± 0.03}} & \underline{\textbf{2.89 ± 0.03}} & \underline{\textbf{20.08 ± 0.35}} \\
                                                                         &                                                                              & \underline{\textit{18.30 ± 0.31}} & \underline{\textbf{33.08 ± 0.41}} & 15.62 ± 0.42                & \underline{\textit{0.50 ± 0.00}} & 0.02 ± 0.00                & \underline{\textit{29.19 ± 0.13}} & 20.16 ± 0.40                & 3.00 ± 0.07                & \underline{\textit{20.45 ± 0.37}} \\ \hline
\multirow{7}{*}{PIECE}                                                   & MD                                                                           & \underline{\textit{0.03 ± 0.00}}  & \underline{\textit{0.03 ± 0.00}}  & \underline{\textit{0.04 ± 0.00}}  & 0.02 ± 0.00                & 0.02 ± 0.00                & \underline{\textbf{0.01 ± 0.00}}  & 0.02 ± 0.00                 & \underline{\textbf{0.01 ± 0.00}} & \underline{\textbf{0.05 ± 0.00}}  \\
                                                                         & DE                                                                           & 0.04 ± 0.00                 & \underline{\textit{0.03 ± 0.00}}  & \underline{\textit{0.04 ± 0.00}}  & \underline{\textbf{0.01 ± 0.00}} & \underline{\textbf{0.01 ± 0.00}} & \underline{\textbf{0.01 ± 0.00}}  & 0.02 ± 0.00                 & 0.02 ± 0.00                & 0.07 ± 0.00                 \\
                                                                         & EDL-R                                                                        & 0.04 ± 0.00                 & \underline{\textit{0.03 ± 0.00}}  & \underline{\textit{0.04 ± 0.00}}  & \underline{\textbf{0.01 ± 0.00}} & 0.02 ± 0.00                & \underline{\textbf{0.01 ± 0.00}}  & \underline{\textbf{0.01 ± 0.00}}  & 0.02 ± 0.00                & 0.08 ± 0.01                 \\
                                                                         & EDL-QR                                                                       & 0.04 ± 0.00                 & \underline{\textit{0.03 ± 0.00}}  & \underline{\textit{0.04 ± 0.00}}  & \underline{\textbf{0.01 ± 0.00}} & \underline{\textbf{0.01 ± 0.00}} & \underline{\textbf{0.01 ± 0.00}}  & \underline{\textbf{0.01 ± 0.00}}  & 0.02 ± 0.00                & 0.06 ± 0.00                 \\
                                                                         & SQR-OC                                                                       & 0.04 ± 0.00                 & \underline{\textit{0.03 ± 0.00}}  & 0.05 ± 0.00                 & \underline{\textbf{0.01 ± 0.00}} & 0.02 ± 0.00                & \underline{\textbf{0.01 ± 0.00}}  & \underline{\textbf{0.01 ± 0.00}}  & 0.02 ± 0.00                & 0.07 ± 0.01                 \\
                                                                         & \multirow{2}{*}{\begin{tabular}[c]{@{}l@{}}Ours\\ Ours-Calib\end{tabular}} & 0.04 ± 0.00                 & \underline{\textit{0.03 ± 0.00}}  & \underline{\textit{0.04 ± 0.00}}  & \underline{\textbf{0.01 ± 0.00}} & \underline{\textbf{0.01 ± 0.00}} & \underline{\textbf{0.01 ± 0.00}}  & \underline{\textbf{0.01 ± 0.00}}  & 0.02 ± 0.00                & 0.06 ± 0.00                 \\
                                                                         &                                                                              & \underline{\textbf{0.02 ± 0.00}}  & \underline{\textbf{0.02 ± 0.00}}  & \underline{\textbf{0.03 ± 0.00}}  & \underline{\textbf{0.01 ± 0.00}} & 0.03 ± 0.00                & \underline{\textbf{0.01 ± 0.00}}  & 0.02 ± 0.00                 & \underline{\textbf{0.01 ± 0.00}} & \underline{\textbf{0.05 ± 0.00}}  \\ \hline
\multirow{7}{*}{$\text{PIECE}^+$}                                        & MD                                                                           & \underline{\textit{0.03 ± 0.00}}  & \underline{\textbf{0.01 ± 0.00}}  & 0.03 ± 0.00                 & \underline{\textit{0.01 ± 0.00}} & 0.02 ± 0.00                & 0.02 ± 0.00                 & 0.02 ± 0.00                 & \underline{\textbf{0.01 ± 0.00}} & 0.07 ± 0.01                 \\
                                                                         & DE                                                                           & 0.07 ± 0.01                 & 0.04 ± 0.01                 & 0.09 ± 0.01                 & 0.03 ± 0.00                & 0.11 ± 0.00                & 0.04 ± 0.00                 & 0.01 ± 0.00                 & 0.02 ± 0.00                & 0.15 ± 0.01                 \\
                                                                         & EDL-R                                                                        & 0.07 ± 0.01                 & 0.05 ± 0.00                 & 0.08 ± 0.01                 & 0.03 ± 0.00                & 0.05 ± 0.00                & 0.04 ± 0.00                 & 0.02 ± 0.00                 & 0.03 ± 0.00                & 0.15 ± 0.01                 \\
                                                                         & EDL-QR                                                                       & \underline{\textbf{0.02 ± 0.00}}  & 0.02 ± 0.00                 & \underline{\textbf{0.01 ± 0.00}}  & \underline{\textit{0.01 ± 0.00}} & \underline{\textbf{0.01 ± 0.00}} & \underline{\textit{0.01 ± 0.00}}  & \underline{\textbf{0.00 ± 0.00}}  & \underline{\textbf{0.01 ± 0.00}} & \underline{\textbf{0.02 ± 0.00}}  \\
                                                                         & SQR-OC                                                                       & \underline{\textit{0.03 ± 0.01}}  & \underline{\textbf{0.01 ± 0.00}}  & \underline{\textit{0.02 ± 0.00}}  & \underline{\textit{0.01 ± 0.00}} & \underline{\textbf{0.01 ± 0.00}} & \underline{\textbf{0.00 ± 0.00}}  & \underline{\textbf{0.00 ± 0.00}}  & 0.02 ± 0.00                & \underline{\textit{0.04 ± 0.01}}  \\
                                                                         & \multirow{2}{*}{\begin{tabular}[c]{@{}l@{}}Ours\\ Ours-Calib\end{tabular}} & 0.04 ± 0.01                 & \underline{\textbf{0.01 ± 0.00}}  & \underline{\textit{0.02 ± 0.00}}  & \underline{\textbf{0.00 ± 0.00}} & \underline{\textbf{0.01 ± 0.00}} & \underline{\textit{0.01 ± 0.00}}  & 0.01 ± 0.00                 & \underline{\textbf{0.01 ± 0.00}} & 0.05 ± 0.01                 \\
                                                                         &                                                                              & \underline{\textit{0.03 ± 0.00}}  & 0.02 ± 0.00                 & \underline{\textit{0.02 ± 0.00}}  & \underline{\textit{0.01 ± 0.00}} & 0.03 ± 0.00                & \underline{\textit{0.01 ± 0.00}}  & 0.02 ± 0.00                 & \underline{\textbf{0.01 ± 0.00}} & \underline{\textit{0.04 ± 0.01}}  \\ \hline
\multirow{7}{*}{$\text{PIECE}^-$}                                        & MD                                                                           & 0.03 ± 0.00                 & 0.03 ± 0.00                 & 0.05 ± 0.00                 & 0.03 ± 0.00                & 0.05 ± 0.00                & 0.03 ± 0.00                 & 0.04 ± 0.00                 & \underline{\textbf{0.01 ± 0.00}} & 0.05 ± 0.00                 \\
                                                                         & DE                                                                           & 0.03 ± 0.00                 & 0.02 ± 0.00                 & 0.04 ± 0.00                 & 0.03 ± 0.00                & 0.05 ± 0.00                & 0.03 ± 0.00                 & 0.04 ± 0.00                 & \underline{\textbf{0.01 ± 0.00}} & 0.04 ± 0.00                 \\
                                                                         & EDL-R                                                                        & 0.03 ± 0.00                 & 0.02 ± 0.00                 & 0.04 ± 0.00                 & 0.02 ± 0.00                & 0.05 ± 0.00                & 0.03 ± 0.00                 & 0.04 ± 0.00                 & \underline{\textbf{0.01 ± 0.00}} & 0.04 ± 0.00                 \\
                                                                         & EDL-QR                                                                       & \underline{\textbf{0.02 ± 0.00}}  & \underline{\textbf{0.01 ± 0.00}}  & \underline{\textbf{0.02 ± 0.00}}  & \underline{\textit{0.01 ± 0.00}} & \underline{\textit{0.02 ± 0.00}} & 0.02 ± 0.00                 & 0.02 ± 0.00                 & \underline{\textbf{0.01 ± 0.00}} & \underline{\textbf{0.03 ± 0.00}}  \\
                                                                         & SQR-OC                                                                       & 0.03 ± 0.00                 & 0.02 ± 0.00                 & \underline{\textbf{0.02 ± 0.00}}  & \underline{\textit{0.01 ± 0.00}} & 0.03 ± 0.00                & \underline{\textbf{0.00 ± 0.00}}  & \underline{\textbf{0.00 ± 0.00}}  & 0.02 ± 0.00                & 0.05 ± 0.01                 \\
                                                                         & \multirow{2}{*}{\begin{tabular}[c]{@{}l@{}}Ours\\ Ours-Calib\end{tabular}} & \underline{\textbf{0.02 ± 0.00}}  & \underline{\textbf{0.01 ± 0.00}}  & \underline{\textbf{0.02 ± 0.00}}  & \underline{\textbf{0.00 ± 0.00}} & \underline{\textbf{0.01 ± 0.00}} & \underline{\textit{0.01 ± 0.00}}  & \underline{\textit{0.01 ± 0.00}}  & \underline{\textbf{0.01 ± 0.00}} & \underline{\textbf{0.03 ± 0.00}}  \\
                                                                         &                                                                              & \underline{\textbf{0.02 ± 0.00}}  & \underline{\textbf{0.01 ± 0.00}}  & \underline{\textbf{0.02 ± 0.00}}  & \underline{\textit{0.01 ± 0.00}} & \underline{\textit{0.02 ± 0.00}} & \underline{\textit{0.01 ± 0.00}}  & 0.02 ± 0.00                 & \underline{\textbf{0.01 ± 0.00}} & \underline{\textbf{0.03 ± 0.00}}  \\ \hline
\multirow{6}{*}{Correlation}                                             & MD                                                                           & \underline{\textbf{0.18 ± 0.02}}  & 0.22 ± 0.01                 & 0.26 ± 0.01                 & \underline{\textbf{0.24 ± 0.01}} & 0.10 ± 0.01                & 0.04 ± 0.00                 & 0.34 ± 0.00                 & 0.21 ± 0.01                & \underline{\textbf{0.19 ± 0.02}}  \\
                                                                         & DE                                                                           & \underline{\textit{0.16 ± 0.01}}  & \underline{\textbf{0.26 ± 0.01}}  & 0.30 ± 0.02                 & 0.18 ± 0.00                & 0.07 ± 0.01                & 0.07 ± 0.00                 & \underline{\textit{0.41 ± 0.00}}  & \underline{\textit{0.27 ± 0.01}} & 0.07 ± 0.03                 \\
                                                                         & EDL-R                                                                        & \underline{\textit{0.16 ± 0.01}}  & 0.24 ± 0.01                 & 0.30 ± 0.01                 & \underline{\textit{0.21 ± 0.01}} & -0.06 ± 0.01               & 0.08 ± 0.00                 & 0.15 ± 0.01                 & \underline{\textit{0.27 ± 0.01}} & 0.04 ± 0.02                 \\
                                                                         & EDL-QR                                                                       & \underline{\textit{0.16 ± 0.01}}  & \underline{\textit{0.25 ± 0.01}}  & \underline{\textbf{0.32 ± 0.01}}  & 0.20 ± 0.00                & \underline{\textit{0.11 ± 0.01}} & \underline{\textit{0.09 ± 0.00}}  & 0.40 ± 0.00                 & 0.24 ± 0.01                & \underline{\textit{0.13 ± 0.02}}  \\
                                                                         & SQR-OC                                                                       & 0.15 ± 0.01                 & 0.21 ± 0.01                 & \underline{\textit{0.31 ± 0.01}}  & 0.17 ± 0.00                & 0.08 ± 0.01                & 0.07 ± 0.00                 & 0.29 ± 0.00                 & 0.19 ± 0.01                & 0.15 ± 0.02                 \\
                                                                         & Ours                                                                         & \underline{\textit{0.16 ± 0.01}}  & 0.24 ± 0.01                 & 0.27 ± 0.01                 & 0.15 ± 0.00                & \underline{\textbf{0.14 ± 0.02}} & \underline{\textbf{0.12 ± 0.00}}  & \underline{\textbf{0.49 ± 0.00}}  & \underline{\textbf{0.33 ± 0.01}} & 0.10 ± 0.02                 \\ \hline
\end{tabular}
}
\end{table*}

\subsection{Robustness to Adversarial Samples}
To further demonstrate that our proposed SDS serves as a fine-grained estimator of epistemic uncertainty, we introduce adversarial perturbations into image inputs and evaluate how uncertainty responds to changes in prediction accuracy. 

Since adversarial noise is generated along the gradient of the model with respect to the input, base models are typically highly sensitive to such perturbations. However, compared to OOD samples, adversarial examples remain close to the iD data in the input space, making their detection more challenging.  As a result, they serve as a more stringent test for evaluating the consistency between epistemic uncertainty estimation and model predictions.

Experiments are conducted on the monocular depth estimation dataset (regression) and CIFAR-10/100 (classification). Details of the perturbation strengths used in each dataset are provided in Appendix~\ref{sec:datasets-depth} and Appendix~\ref{sec:datasets-cls}.

\subsubsection{Results on Monocular Depth Estimation}
\begin{figure*}
  \centering
  \includegraphics[width=1\linewidth]{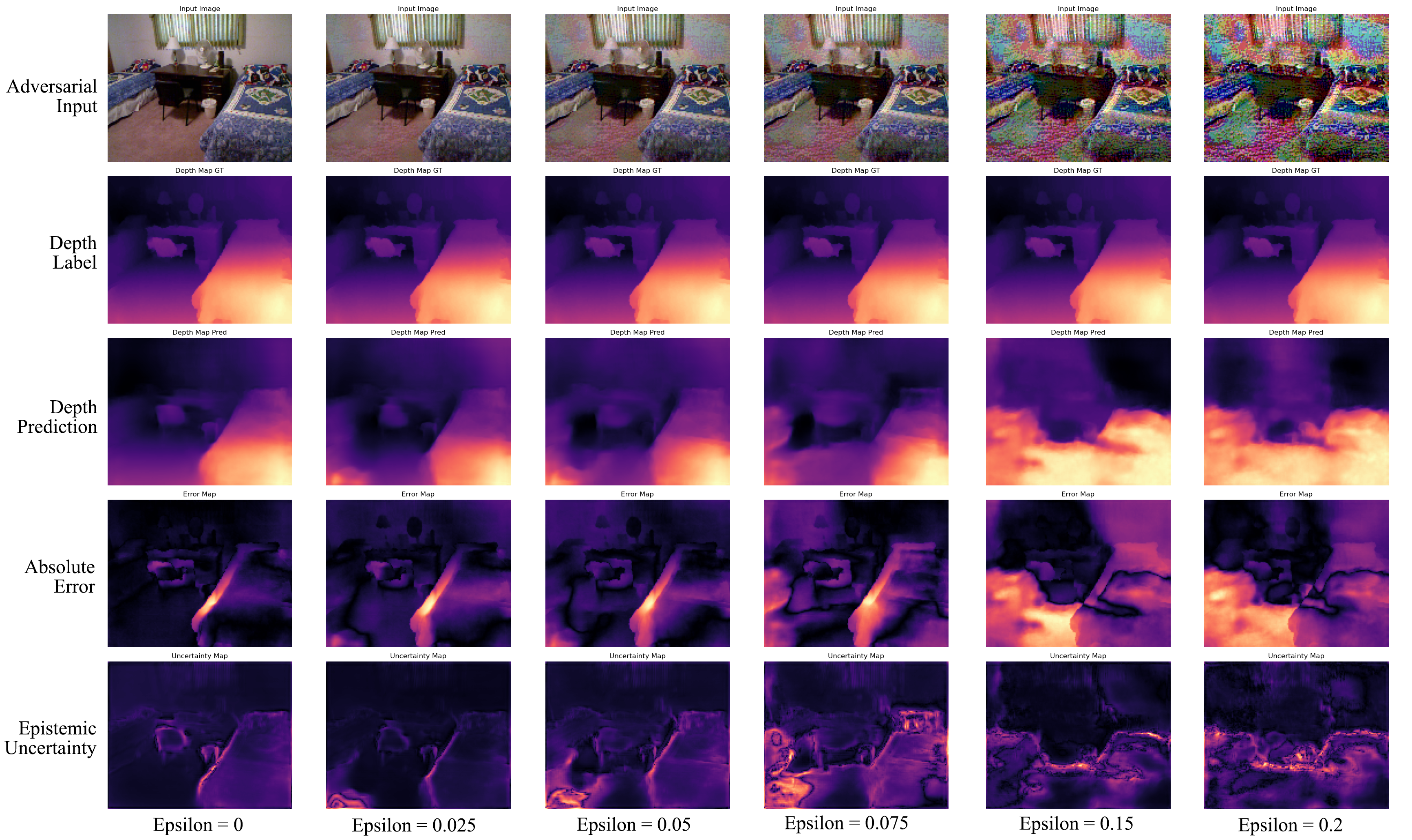}
\caption{
Visualization of the effect of increasing adversarial perturbations on monocular depth predictions, prediction error, and epistemic uncertainty produced by our method (SDS). As the perturbation strength increases, the corrupted regions in the depth maps become more apparent. The uncertainty maps consistently highlight regions of prediction error, demonstrating the model’s ability to capture model uncertainty under adversarial attack.
}
  \label{fig:Monocular-Depth-noise-Visualization}
\end{figure*}

\begin{figure*}
  \centering
  \includegraphics[width=1\linewidth]{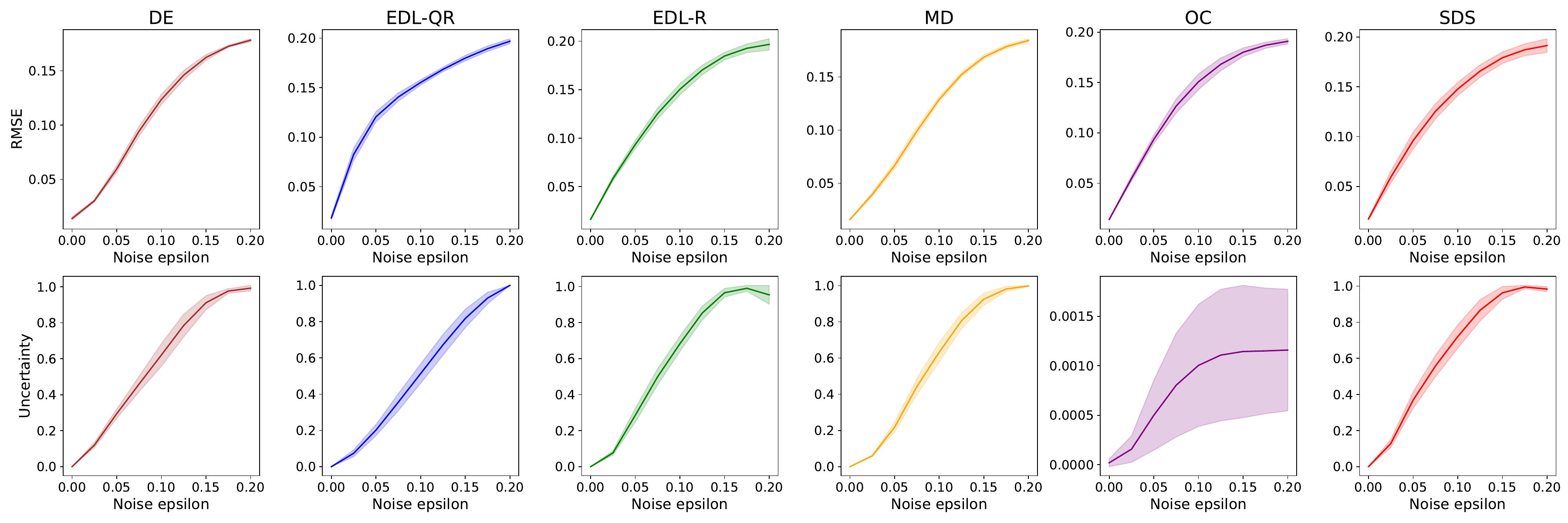}
\caption{
Comparison of RMSE and epistemic uncertainty under increasing adversarial perturbations in the monocular depth estimation task. A fine-grained UQ method is expected to produce uncertainty curves that closely track the upward trend of RMSE.
}
  \label{fig:rmse-uncertainty-grid}
\end{figure*}

We first analyze the consistency between the average RMSE of depth predictions and the corresponding epistemic uncertainty scores. As shown in Figure~\ref{fig:Monocular-Depth-noise-Visualization}, adversarial perturbations do not change the underlying depth structure of the scene but significantly degrade model performance, as evidenced by the rising RMSE curve in Figure~\ref{fig:rmse-uncertainty-grid}. To evaluate the quality of epistemic uncertainty estimation, a fine-grained uncertainty score is expected to exhibit strong consistency with the degradation of model performance reflected by RMSE.

Figure~\ref{fig:rmse-uncertainty-grid} illustrates the behavior of prediction error and estimated epistemic uncertainty as adversarial noise intensifies. Among all methods, the OC method exhibits minimal changes in uncertainty, indicating its limited capability in detecting adversarial samples in monocular depth estimation. EDL-based methods also fail to produce consistent trends between uncertainty and RMSE. In particular, the blue curve (EDL-QR) shows a rapid increase in RMSE, while the corresponding uncertainty grows slowly. Similarly, the green curve (EDL-R) shows a counter-intuitive decrease in uncertainty under strong perturbations.

In contrast, the epistemic uncertainty estimated by our method (SDS) closely aligns with the RMSE escalation trend, demonstrating its robustness in capturing uncertainty under adversarial perturbations in regression tasks.

\subsubsection{Results on Image Classification}

We further evaluate the alignment between epistemic uncertainty and classification accuracy using Wide-ResNet as the base model.

\begin{figure*}
  \centering
  \includegraphics[width=1\linewidth]{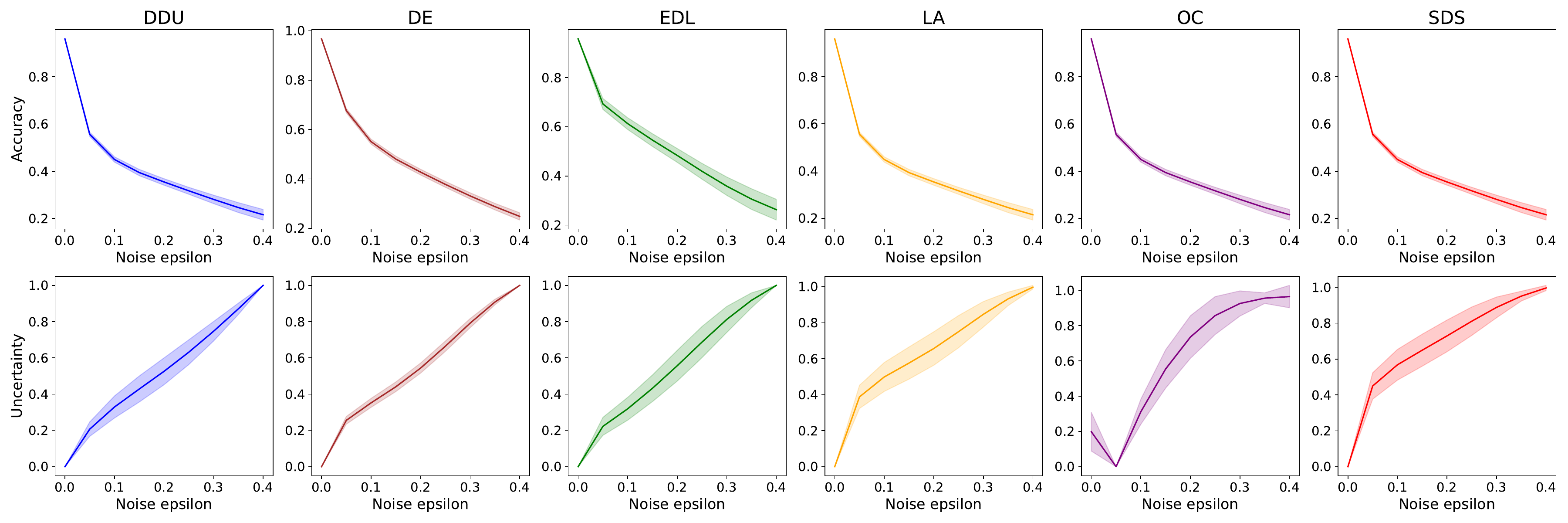}
  \caption{Comparison of classification accuracy and epistemic uncertainty under increasing adversarial perturbations on CIFAR-10. A fine-grained UQ method is expected to produce uncertainty curves that closely track the downward trend of accuracy.}
  \label{fig:acc-unc-grid-cifar10}
\end{figure*}

\begin{figure*}
  \centering
  \includegraphics[width=1\linewidth]{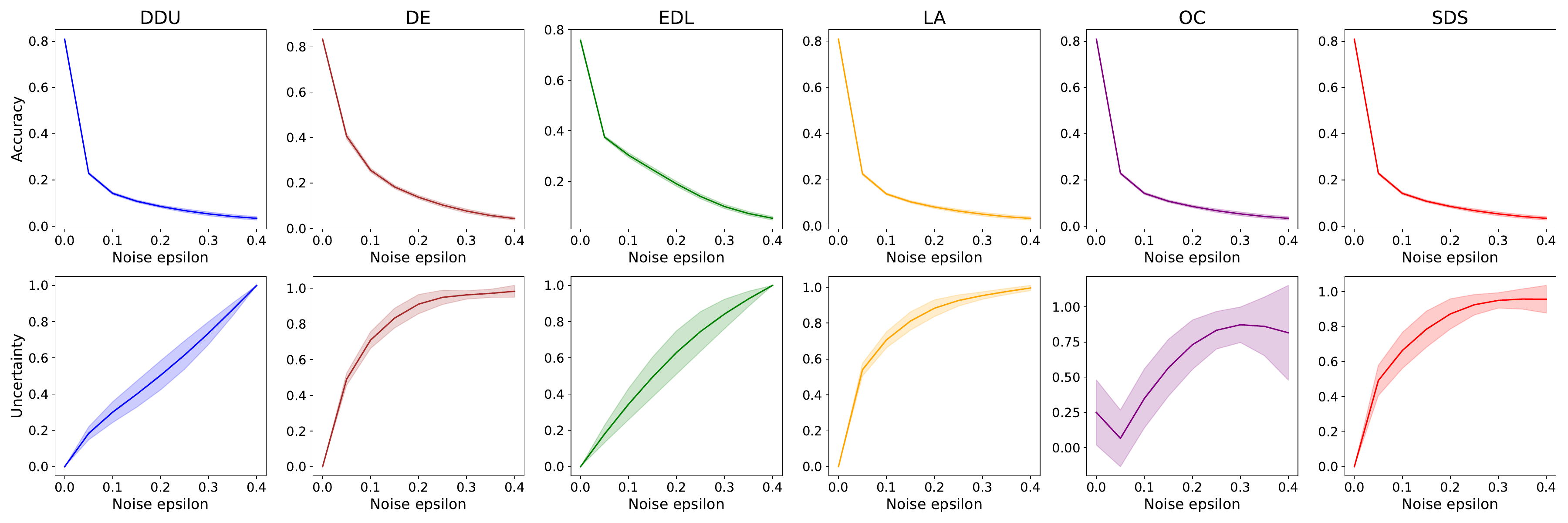}
  \caption{Comparison of classification accuracy and epistemic uncertainty under increasing adversarial perturbations on CIFAR-100. A fine-grained UQ method is expected to produce uncertainty curves that closely track the downward trend of accuracy.}
  \label{fig:acc-unc-grid-cifar100}
\end{figure*}

As shown in Figure~\ref{fig:acc-unc-grid-cifar10} and Figure~\ref{fig:acc-unc-grid-cifar100}, with increasing adversarial noise strength (\(\epsilon\)), our method (SDS), along with LA and DE, exhibits a consistent trend between the rise in epistemic uncertainty and the degradation in classification accuracy. In contrast, DDU and EDL, which are representative distribution-based methods, fail to preserve this alignment. Their uncertainty scores increase almost linearly with the perturbation strength, regardless of the actual prediction accuracy. Even more notably, OC displays a counterintuitive decrease in uncertainty under stronger adversarial noise.
This mismatch between uncertainty and accuracy trends suggests that these methods primarily capture shifts in the input distribution rather than epistemic uncertainty stemming from the model’s predictive limitations. As a result, they fail to provide reliable uncertainty estimates under adversarial conditions.

In summary, for both regression and classification, our method incorporates the original task prediction into the uncertainty model, enabling it to capture nonlinear epistemic uncertainty induced by adversarial perturbations that directly target the task loss. Moreover, even as the strength of adversarial noise varies, the marked sensitivity to near-OOD inputs underscores the fine-grained resolution of our uncertainty estimates.

\section{Extended Experimental Findings}  
\label{sec:extend-study}

In our UQ framework, training the UQ network represents a distinct subtask within the overall learning pipeline.  To identify the best way to integrate this subtask with a DL base model and to assess its training dynamics and practical utility in different scenarios, we conduct experiments following the protocol in Appendix~\ref{tab:other-protocol}.  These studies evaluate different integration workflows, measure the stability and performance of the UQ networks, and offer guidance for robust, efficient deployment in real-world settings.

\subsection{Results on Training Strategies}
\label{subsect:res-trainging-strategy}

As noted in the main text, when no pretrained base model is available, our framework supports two training strategies: \emph{stagewise} or \emph{joint}.  Here, we compare the joint training against stagewise (post-hoc) training of the UQ network using different fractions of the original training set.  This evaluation mimics real-world scenarios where full access to the training data is limited or expensive.

Specifically, we select several UCI regression datasets containing more than 700 samples to form a representative regression benchmark. For classification, we adopt CIFAR-10, CIFAR-100 and ImageNet-1K as the benchmark datasets.

\subsubsection{Results on Regression}
Table~\ref{tab:training-reg-as} reports the results on the regression task under different training strategies. We first compare models trained jointly (denoted as Jointly) with those trained in a stagewise manner using the full training set (denoted as \(\mathcal{D}_{\mathrm{T}} \times 1.0\)). Since the UQ heads introduce additional loss terms and propagate gradients back into the base model during joint training, both the point prediction performance (measured by RMSE) and the quality of PIs (quantified by Winkler Score and PIECEs) tend to deteriorate. This suggests that multi-task joint optimization may introduce interference between the UQ heads and the primary task objective.
Interestingly, joint training yields a higher correlation between predicted uncertainty and regression error. This likely results from the UQ heads shaping the base model parameters during training, leading to a tighter alignment between the estimated uncertainty and the model’s epistemic limitations.

\begin{figure*}
  \centering
  \includegraphics[width=1\linewidth]{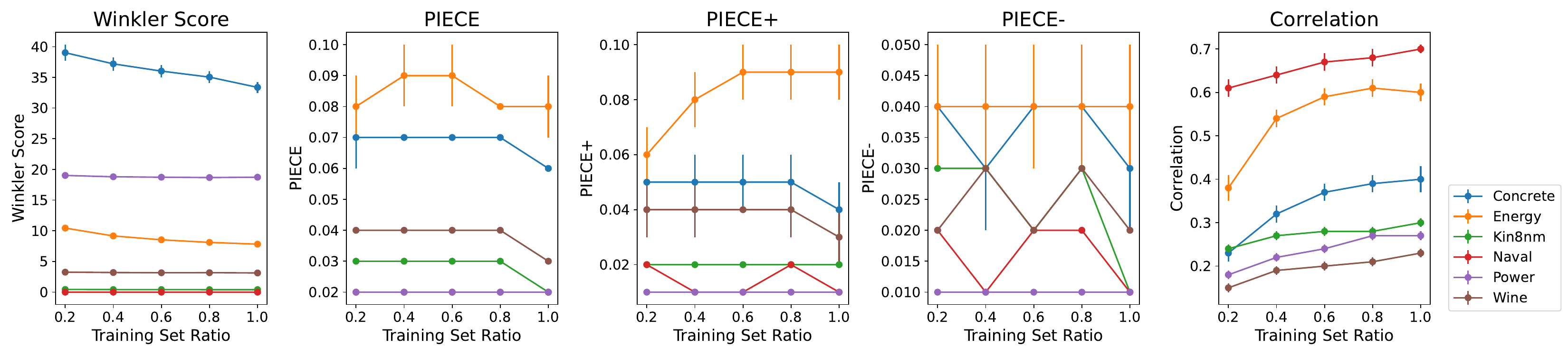}
  \caption{Effect of Data Subsampling on Post-hoc Uncertainty Estimation in Regression Benchmark}
  \label{fig:post-training-reg}
\end{figure*}

To further examine the effect of training UQ network with partial training data, we visualize the results from Table~\ref{tab:training-reg-as} as plots in Figure~\ref{fig:post-training-reg}, showing how the performance of post-hoc trained UQ networks varies across different training set proportions.
As observed, the quality of PIs generally improves with an increasing proportion of training data, as reflected by the decreasing Winkler Score and PIECEs. The improvement is particularly notable on the \textit{Concrete} and \textit{Energy} datasets, which can be attributed to their small sizes, where downsampling further reduces the diversity and coverage of the data distribution. In contrast, other datasets exhibit more stable PI quality across varying training ratios, as their larger sample sizes retain sufficient distributional information even when only a subset is used.

Regarding the correlation between uncertainty and RMSE, all datasets exhibit a clear increasing trend as the training set size is raised. This indicates that the distribution mismatch between the UQ head and the base model caused by training on different data subsets hampers the UQ head’s ability to accurately capture the base model’s knowledge limitations.

In summary, our experiments demonstrate that aleatoric uncertainty quantification chiefly depends on accurately modeling the data distribution and remains reliable provided the training set offers sufficient coverage.  By contrast, epistemic uncertainty quantification relies on capturing the base model’s learned distribution, which is best preserved when the UQ network is trained on the same dataset as the base model.

\begin{table*}
\caption{Results on UCI Regression Benchmarks under Different Training Strategies}
\label{tab:training-reg-as}
\centering
\scalebox{1}{
\begin{tabular}{ccllllll}
\hline
\multicolumn{1}{l}{}                                                      &                                           & \multicolumn{6}{c}{Datasets}                                                                                                                                                                                                                                                                                                  \\ \cline{3-8} 
\multicolumn{1}{l}{\multirow{-2}{*}{Metrics}}                             & \multirow{-2}{*}{Training}                & Concrete                                            & Energy                                             & Kin8nm                                             & Naval                                              & Power                                               & Wine                                               \\ \hline
                                                                          & \(|\mathcal{D}_{\mathrm{T}}| \times 0.2\) & \underline{\textbf{7.13 ± 0.13}}                          & \underline{\textbf{2.47 ± 0.07}}                         & \underline{\textbf{0.09 ± 0.00}}                         & \underline{\textbf{0.00 ± 0.00}}                         & \underline{\textbf{3.97 ± 0.03}}                          & \underline{\textbf{0.64 ± 0.01}}                         \\
                                                                          & \(|\mathcal{D}_{\mathrm{T}}| \times 0.4\) & \underline{\textbf{7.13 ± 0.13}}                          & \underline{\textbf{2.47 ± 0.07}}                         & \underline{\textbf{0.09 ± 0.00}}                         & \underline{\textbf{0.00 ± 0.00}}                         & \underline{\textbf{3.97 ± 0.03}}                          & \underline{\textbf{0.64 ± 0.01}}                         \\
                                                                          & \(|\mathcal{D}_{\mathrm{T}}| \times 0.6\) & \underline{\textbf{7.13 ± 0.13}}                          & \underline{\textbf{2.47 ± 0.07}}                         & \underline{\textbf{0.09 ± 0.00}}                         & \underline{\textbf{0.00 ± 0.00}}                         & \underline{\textbf{3.97 ± 0.03}}                          & \underline{\textbf{0.64 ± 0.01}}                         \\
                                                                          & \(|\mathcal{D}_{\mathrm{T}}| \times 0.8\) & \underline{\textbf{7.13 ± 0.13}}                          & \underline{\textbf{2.47 ± 0.07}}                         & \underline{\textbf{0.09 ± 0.00}}                         & \underline{\textbf{0.00 ± 0.00}}                         & \underline{\textbf{3.97 ± 0.03}}                          & \underline{\textbf{0.64 ± 0.01}}                         \\
                                                                          & \(|\mathcal{D}_{\mathrm{T}}| \times 1.0\) & \underline{\textbf{7.13 ± 0.13}}                          & \underline{\textbf{2.47 ± 0.07}}                         & \underline{\textbf{0.09 ± 0.00}}                         & \underline{\textbf{0.00 ± 0.00}}                         & \underline{\textbf{3.97 ± 0.03}}                          & \underline{\textbf{0.64 ± 0.01}}                         \\
\multirow{-6}{*}{RMSE}                                                    & \cellcolor[HTML]{EFEFEF}Jointly           & \cellcolor[HTML]{EFEFEF}7.65 ± 0.12                 & \cellcolor[HTML]{EFEFEF}2.72 ± 0.08                & \cellcolor[HTML]{EFEFEF}0.11 ± 0.00                & \cellcolor[HTML]{EFEFEF}\underline{\textbf{0.00 ± 0.00}} & \cellcolor[HTML]{EFEFEF}4.06 ± 0.03                 & \cellcolor[HTML]{EFEFEF}\underline{\textbf{0.64 ± 0.01}} \\ \hline
                                                                          & \(|\mathcal{D}_{\mathrm{T}}| \times 0.2\) & 39.02 ± 1.31                                        & 10.44 ± 0.42                                       & 0.44 ± 0.01                                        & \underline{\textbf{0.00 ± 0.00}}                         & 19.02 ± 0.25                                        & 3.26 ± 0.06                                        \\
                                                                          & \(|\mathcal{D}_{\mathrm{T}}| \times 0.4\) & 37.20 ± 1.08                                        & 9.16 ± 0.29                                        & 0.42 ± 0.00                                        & \underline{\textbf{0.00 ± 0.00}}                         & 18.80 ± 0.24                                        & 3.20 ± 0.06                                        \\
                                                                          & \(|\mathcal{D}_{\mathrm{T}}| \times 0.6\) & 36.02 ± 1.00                                        & 8.53 ± 0.22                                        & 0.42 ± 0.01                                        & \underline{\textbf{0.00 ± 0.00}}                         & 18.73 ± 0.24                                        & \underline{\textit{3.16 ± 0.06}}                         \\
                                                                          & \(|\mathcal{D}_{\mathrm{T}}| \times 0.8\) & 35.04 ± 0.94                                        & \underline{\textit{8.11 ± 0.16}}                         & \underline{\textbf{0.41 ± 0.00}}                         & \underline{\textbf{0.00 ± 0.00}}                         & \underline{\textbf{18.67 ± 0.24}}                         & 3.17 ± 0.06                                        \\
                                                                          & \(|\mathcal{D}_{\mathrm{T}}| \times 1.0\) & \underline{\textbf{33.37 ± 0.89}}                         & \underline{\textbf{7.82 ± 0.15}}                         & \underline{\textbf{0.41 ± 0.01}}                         & \underline{\textbf{0.00 ± 0.00}}                         & 18.73 ± 0.24                                        & \underline{\textbf{3.13 ± 0.06}}                         \\
\multirow{-6}{*}{\begin{tabular}[c]{@{}c@{}}Winkler\\ Score\end{tabular}} & \cellcolor[HTML]{EFEFEF}Jointly           & \cellcolor[HTML]{EFEFEF}\underline{\textit{34.52 ± 0.91}} & \cellcolor[HTML]{EFEFEF}8.54 ± 0.17                & \cellcolor[HTML]{EFEFEF}0.42 ± 0.00                & \cellcolor[HTML]{EFEFEF}\underline{\textbf{0.00 ± 0.00}} & \cellcolor[HTML]{EFEFEF}\underline{\textit{18.69 ± 0.25}} & \cellcolor[HTML]{EFEFEF}3.20 ± 0.06                \\ \hline
                                                                          & \(|\mathcal{D}_{\mathrm{T}}| \times 0.2\) & 0.07 ± 0.01                                         & \underline{\textbf{0.08 ± 0.01}}                         & \underline{\textit{0.03 ± 0.00}}                         & \underline{\textbf{0.02 ± 0.00}}                         & \underline{\textbf{0.02 ± 0.00}}                          & \underline{\textit{0.04 ± 0.00}}                         \\
                                                                          & \(|\mathcal{D}_{\mathrm{T}}| \times 0.4\) & 0.07 ± 0.00                                         & 0.09 ± 0.01                                        & \underline{\textit{0.03 ± 0.00}}                         & \underline{\textbf{0.02 ± 0.00}}                         & \underline{\textbf{0.02 ± 0.00}}                          & \underline{\textit{0.04 ± 0.00}}                         \\
                                                                          & \(|\mathcal{D}_{\mathrm{T}}| \times 0.6\) & 0.07 ± 0.00                                         & 0.09 ± 0.01                                        & \underline{\textit{0.03 ± 0.00}}                         & \underline{\textbf{0.02 ± 0.00}}                         & \underline{\textbf{0.02 ± 0.00}}                          & \underline{\textit{0.04 ± 0.00}}                         \\
                                                                          & \(|\mathcal{D}_{\mathrm{T}}| \times 0.8\) & 0.07 ± 0.00                                         & \underline{\textbf{0.08 ± 0.00}}                         & \underline{\textit{0.03 ± 0.00}}                         & \underline{\textbf{0.02 ± 0.00}}                         & \underline{\textbf{0.02 ± 0.00}}                          & \underline{\textit{0.04 ± 0.00}}                         \\
                                                                          & \(|\mathcal{D}_{\mathrm{T}}| \times 1.0\) & \underline{\textbf{0.06 ± 0.00}}                          & \underline{\textbf{0.08 ± 0.01}}                         & \underline{\textbf{0.02 ± 0.00}}                         & \underline{\textbf{0.02 ± 0.00}}                         & \underline{\textbf{0.02 ± 0.00}}                          & \underline{\textbf{0.03 ± 0.00}}                         \\
\multirow{-6}{*}{PIECE}                                                   & \cellcolor[HTML]{EFEFEF}Jointly           & \cellcolor[HTML]{EFEFEF}\underline{\textbf{0.06 ± 0.00}}  & \cellcolor[HTML]{EFEFEF}\underline{\textbf{0.08 ± 0.00}} & \cellcolor[HTML]{EFEFEF}\underline{\textit{0.03 ± 0.00}} & \cellcolor[HTML]{EFEFEF}0.07 ± 0.02                & \cellcolor[HTML]{EFEFEF}\underline{\textbf{0.02 ± 0.00}}  & \cellcolor[HTML]{EFEFEF}\underline{\textit{0.04 ± 0.00}} \\ \hline
                                                                          & \(|\mathcal{D}_{\mathrm{T}}| \times 0.2\) & 0.05 ± 0.01                                         & \underline{\textbf{0.06 ± 0.01}}                         & \underline{\textbf{0.02 ± 0.00}}                         & 0.02 ± 0.00                                        & \underline{\textbf{0.01 ± 0.00}}                          & \underline{\textit{0.04 ± 0.01}}                         \\
                                                                          & \(|\mathcal{D}_{\mathrm{T}}| \times 0.4\) & 0.05 ± 0.01                                         & \underline{\textit{0.08 ± 0.01}}                         & \underline{\textbf{0.02 ± 0.00}}                         & \underline{\textbf{0.01 ± 0.00}}                         & \underline{\textbf{0.01 ± 0.00}}                          & \underline{\textit{0.04 ± 0.01}}                         \\
                                                                          & \(|\mathcal{D}_{\mathrm{T}}| \times 0.6\) & 0.05 ± 0.01                                         & 0.09 ± 0.01                                        & \underline{\textbf{0.02 ± 0.00}}                         & \underline{\textbf{0.01 ± 0.00}}                         & \underline{\textbf{0.01 ± 0.00}}                          & \underline{\textit{0.04 ± 0.00}}                         \\
                                                                          & \(|\mathcal{D}_{\mathrm{T}}| \times 0.8\) & 0.05 ± 0.01                                         & 0.09 ± 0.01                                        & \underline{\textbf{0.02 ± 0.00}}                         & 0.02 ± 0.00                                        & \underline{\textbf{0.01 ± 0.00}}                          & \underline{\textit{0.04 ± 0.01}}                         \\
                                                                          & \(|\mathcal{D}_{\mathrm{T}}| \times 1.0\) & \underline{\textbf{0.04 ± 0.01}}                          & 0.09 ± 0.01                                        & \underline{\textbf{0.02 ± 0.00}}                         & \underline{\textbf{0.01 ± 0.00}}                         & \underline{\textbf{0.01 ± 0.00}}                          & \underline{\textbf{0.03 ± 0.01}}                         \\
\multirow{-6}{*}{$\text{PIECE}^+$}                                        & \cellcolor[HTML]{EFEFEF}Jointly           & \cellcolor[HTML]{EFEFEF}\underline{\textbf{0.04 ± 0.01}}  & \cellcolor[HTML]{EFEFEF}0.09 ± 0.01                & \cellcolor[HTML]{EFEFEF}\underline{\textbf{0.02 ± 0.00}} & \cellcolor[HTML]{EFEFEF}0.06 ± 0.02                & \cellcolor[HTML]{EFEFEF}\underline{\textbf{0.01 ± 0.00}}  & \cellcolor[HTML]{EFEFEF}\underline{\textit{0.04 ± 0.01}} \\ \hline
                                                                          & \(|\mathcal{D}_{\mathrm{T}}| \times 0.2\) & 0.04 ± 0.01                                         & \underline{\textit{0.04 ± 0.01}}                         & 0.03 ± 0.00                                        & 0.02 ± 0.00                                        & \underline{\textbf{0.01 ± 0.00}}                          & \underline{\textbf{0.02 ± 0.00}}                         \\
                                                                          & \(|\mathcal{D}_{\mathrm{T}}| \times 0.4\) & \underline{\textbf{0.03 ± 0.01}}                          & \underline{\textit{0.04 ± 0.01}}                         & 0.03 ± 0.00                                        & \underline{\textbf{0.01 ± 0.00}}                         & \underline{\textbf{0.01 ± 0.00}}                          & 0.03 ± 0.00                                        \\
                                                                          & \(|\mathcal{D}_{\mathrm{T}}| \times 0.6\) & 0.04 ± 0.01                                         & \underline{\textit{0.04 ± 0.01}}                         & \underline{\textit{0.02 ± 0.00}}                         & 0.02 ± 0.00                                        & \underline{\textbf{0.01 ± 0.00}}                          & \underline{\textbf{0.02 ± 0.00}}                         \\
                                                                          & \(|\mathcal{D}_{\mathrm{T}}| \times 0.8\) & 0.04 ± 0.01                                         & \underline{\textit{0.04 ± 0.01}}                         & 0.03 ± 0.00                                        & 0.02 ± 0.00                                        & \underline{\textbf{0.01 ± 0.00}}                          & 0.03 ± 0.00                                        \\
                                                                          & \(|\mathcal{D}_{\mathrm{T}}| \times 1.0\) & \underline{\textbf{0.03 ± 0.01}}                          & \underline{\textit{0.04 ± 0.01}}                         & \underline{\textbf{0.01 ± 0.00}}                         & \underline{\textbf{0.01 ± 0.00}}                         & \underline{\textbf{0.01 ± 0.00}}                          & \underline{\textbf{0.02 ± 0.00}}                         \\
\multirow{-6}{*}{$\text{PIECE}^-$}                                        & \cellcolor[HTML]{EFEFEF}Jointly           & \cellcolor[HTML]{EFEFEF}0.04 ± 0.01                 & \cellcolor[HTML]{EFEFEF}\underline{\textbf{0.03 ± 0.01}} & \cellcolor[HTML]{EFEFEF}\underline{\textit{0.02 ± 0.00}} & \cellcolor[HTML]{EFEFEF}0.08 ± 0.02                & \cellcolor[HTML]{EFEFEF}\underline{\textbf{0.01 ± 0.00}}  & \cellcolor[HTML]{EFEFEF}\underline{\textbf{0.02 ± 0.00}} \\ \hline
                                                                          & \(|\mathcal{D}_{\mathrm{T}}| \times 0.2\) & 0.23 ± 0.02                                         & 0.38 ± 0.03                                        & 0.24 ± 0.01                                        & 0.61 ± 0.02                                        & 0.18 ± 0.01                                         & 0.15 ± 0.01                                        \\
                                                                          & \(|\mathcal{D}_{\mathrm{T}}| \times 0.4\) & 0.32 ± 0.02                                         & 0.54 ± 0.02                                        & 0.27 ± 0.01                                        & 0.64 ± 0.02                                        & 0.22 ± 0.01                                         & 0.19 ± 0.01                                        \\
                                                                          & \(|\mathcal{D}_{\mathrm{T}}| \times 0.6\) & 0.37 ± 0.02                                         & 0.59 ± 0.02                                        & 0.28 ± 0.01                                        & 0.67 ± 0.02                                        & 0.24 ± 0.01                                         & 0.20 ± 0.01                                        \\
                                                                          & \(|\mathcal{D}_{\mathrm{T}}| \times 0.8\) & 0.39 ± 0.02                                         & \underline{\textit{0.61 ± 0.02}}                         & 0.28 ± 0.01                                        & \underline{\textit{0.68 ± 0.02}}                         & \underline{\textbf{0.27 ± 0.01}}                          & \underline{\textit{0.21 ± 0.01}}                         \\
                                                                          & \(|\mathcal{D}_{\mathrm{T}}| \times 1.0\) & \underline{\textit{0.40 ± 0.03}}                          & 0.60 ± 0.02                                        & \underline{\textit{0.30 ± 0.01}}                         & \underline{\textbf{0.70 ± 0.01}}                         & \underline{\textbf{0.27 ± 0.01}}                          & \underline{\textbf{0.23 ± 0.01}}                         \\
\multirow{-6}{*}{Correlation}                                             & \cellcolor[HTML]{EFEFEF}Jointly           & \cellcolor[HTML]{EFEFEF}\underline{\textbf{0.42 ± 0.02}}  & \cellcolor[HTML]{EFEFEF}\underline{\textbf{0.70 ± 0.01}} & \cellcolor[HTML]{EFEFEF}\underline{\textbf{0.37 ± 0.01}} & \cellcolor[HTML]{EFEFEF}\underline{\textit{0.68 ± 0.02}} & \cellcolor[HTML]{EFEFEF}\underline{\textbf{0.27 ± 0.01}}  & \cellcolor[HTML]{EFEFEF}0.19 ± 0.01                \\ \hline
\end{tabular}
}
\end{table*}

\subsubsection{Results on Image Classification}
Table~\ref{tab:training-cls-as} summarizes the results on the classification benchmarks under different training strategies. We first compare models trained jointly with those trained in a stagewise manner using the full training set. Unlike in the regression setting, classification accuracy and calibration quality (measured by ECE) show no significant differences between the two strategies. This may be because classification models already possess strong feature extraction capabilities, and the addition of UQ heads does not noticeably interfere with the optimization of the task head.
However, joint training leads to a clear drop in adversarial detection performance (AUROC (adv)). This can be attributed to the altered classification loss during joint training, which reduces the effectiveness of detecting adversarial perturbations generated using standard cross-entropy gradients.
For OOD detection and misclassification detection, both strategies yield comparable results, with no consistent performance gap.

\begin{figure*}
  \centering
  \includegraphics[width=0.85\linewidth]{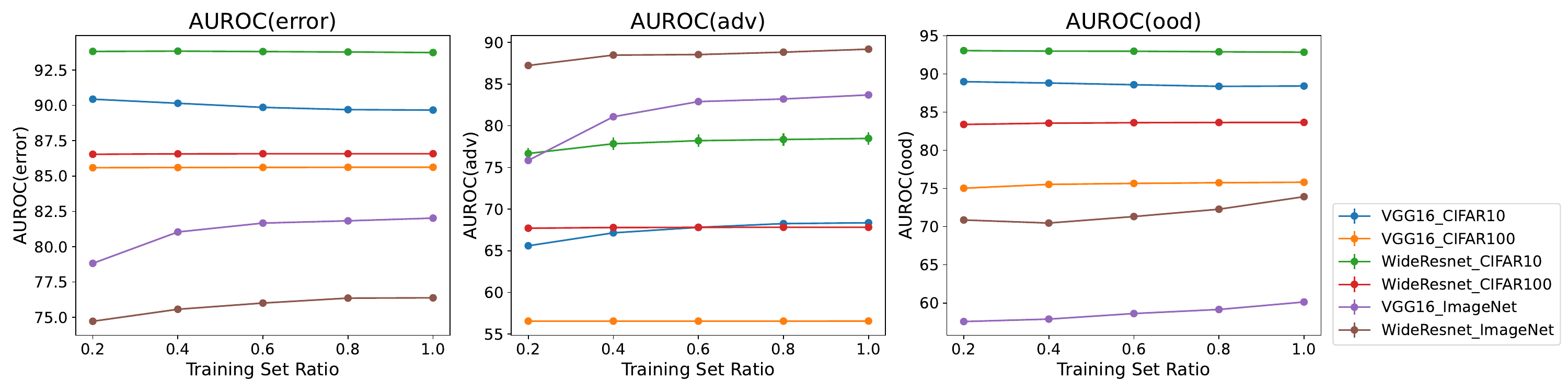}
  \caption{Effect of Data Subsampling on Post-hoc Uncertainty Estimation in Classification Benchmark}
  \label{fig:post-training-cls}
\end{figure*}

We also visualize the results from Table~\ref{tab:training-cls-as} in Figure~\ref{fig:post-training-cls}, illustrating how the performance of post-hoc trained UQ networks varies with different proportions of training data. Since these experiments are conducted on the training set, we do not analyze calibration performance here, which will be discussed in the next section.
As the training set size increases, the three AUROC metrics on CIFAR-10 and CIFAR-100 remain relatively stable. In contrast, for ImageNet, the increased complexity and diversity of the dataset result in greater aleatoric uncertainty. The accuracy of epistemic uncertainty estimates improves with more training data, indicating that access to the full dataset enables the UQ network to better capture the model's knowledge boundaries.

In summary, our experiments indicate that, on simple benchmark datasets, MAR values are typically very small, since \(\mathrm{MAR}^+\) and \(\mathrm{MAR}^-\) derive directly from the softmax outputs. Therefore, their effect on self-consistency verification is minimal and the SDS framework instead relies on the base model’s raw predictions.  Conversely, on complex datasets the MAR values are larger and contribute significantly to uncertainty estimation.  In these cases, training the UQ network on a larger fraction of the data improves performance by enabling it to more accurately capture the base model’s knowledge distribution.

\begin{table*}
\caption{Results on Classification Benchmark under Different Training Strategies}
\label{tab:training-cls-as}
\centering
\scalebox{0.85}{
\begin{tabular}{lc|llll|llll}
\hline
\multicolumn{1}{c}{}                          &                                           & \multicolumn{4}{c|}{VGG16}                                                                                                                                                              & \multicolumn{4}{c}{Wide-ResNet}                                                                                                                                                   \\ \cline{3-10} 
\multicolumn{1}{c}{\multirow{-2}{*}{Dataset}} & \multirow{-2}{*}{Training}                & Accuracy                                            & AUROC(error)                                        & AUROC(adv)                           & AUROC(ood)                           & Accuracy                                            & AUROC(error)                         & AUROC(adv)                                    & AUROC(ood)                           \\ \hline
                                              & \(|\mathcal{D}_{\mathrm{T}}| \times 0.2\) & \underline{\textbf{93.62 ± 0.03}}                         & \underline{\textbf{90.44 ± 0.15}}                         & 65.60 ± 0.17                         & \underline{\textbf{88.98 ± 0.35}}          & \underline{\textit{96.01 ± 0.03}}                         & \underline{\textit{93.82 ± 0.16}}          & 76.67 ± 0.67                                  & \underline{\textbf{93.05 ± 0.19}}          \\
                                              & \(|\mathcal{D}_{\mathrm{T}}| \times 0.4\) & \underline{\textbf{93.62 ± 0.03}}                         & \underline{\textit{90.15 ± 0.17}}                         & 67.15 ± 0.18                         & \underline{\textit{88.81 ± 0.37}}          & \underline{\textit{96.01 ± 0.03}}                         & \underline{\textbf{93.84 ± 0.14}}          & 77.84 ± 0.75                                  & \underline{\textit{92.98 ± 0.20}}          \\
                                              & \(|\mathcal{D}_{\mathrm{T}}| \times 0.6\) & \underline{\textbf{93.62 ± 0.03}}                         & 89.86 ± 0.15                                        & 67.81 ± 0.17                         & 88.58 ± 0.39                         & \underline{\textit{96.01 ± 0.03}}                         & 93.81 ± 0.16                         & 78.21 ± 0.76                                  & 92.97 ± 0.21                         \\
                                              & \(|\mathcal{D}_{\mathrm{T}}| \times 0.8\) & \underline{\textbf{93.62 ± 0.03}}                         & 89.70 ± 0.14                                        & \textit{68.26 ± 0.17}                & 88.37 ± 0.43                         & \underline{\textit{96.01 ± 0.03}}                         & 93.78 ± 0.16                         & \textit{78.35 ± 0.77}                         & 92.90 ± 0.22                         \\
                                              & \(|\mathcal{D}_{\mathrm{T}}| \times 1.0\) & \underline{\textbf{93.62 ± 0.03}}                         & 89.67 ± 0.15                                        & \textbf{68.36 ± 0.17}                & 88.41 ± 0.43                         & \underline{\textit{96.01 ± 0.03}}                         & 93.74 ± 0.18                         & \textbf{78.48 ± 0.77}                         & 92.84 ± 0.23                         \\
\multirow{-6}{*}{CIFAR10}                     & \cellcolor[HTML]{EFEFEF}Jointly           & \cellcolor[HTML]{EFEFEF}93.50 ± 0.05                & \cellcolor[HTML]{EFEFEF}89.60 ± 0.18                & \cellcolor[HTML]{EFEFEF}60.77 ± 0.30 & \cellcolor[HTML]{EFEFEF}87.48 ± 0.48 & \cellcolor[HTML]{EFEFEF}\underline{\textbf{96.03 ± 0.03}} & \cellcolor[HTML]{EFEFEF}92.22 ± 0.19 & \cellcolor[HTML]{EFEFEF}64.23 ± 0.33          & \cellcolor[HTML]{EFEFEF}90.97 ± 0.43 \\ \hline
                                              & \(|\mathcal{D}_{\mathrm{T}}| \times 0.2\) & \underline{\textit{73.51 ± 0.05}}                         & 85.59 ± 0.09                                        & \textit{56.56 ± 0.07}                & 75.03 ± 0.40                         & \underline{\textit{80.88 ± 0.05}}                         & 86.54 ± 0.07                         & 67.71 ± 0.06                                  & 83.38 ± 0.26                         \\
                                              & \(|\mathcal{D}_{\mathrm{T}}| \times 0.4\) & \underline{\textit{73.51 ± 0.06}}                         & 85.60 ± 0.09                                        & \textit{56.56 ± 0.07}                & 75.53 ± 0.39                         & \underline{\textit{80.88 ± 0.05}}                         & \underline{\textit{86.57 ± 0.07}}          & 67.79 ± 0.06                                  & 83.55 ± 0.26                         \\
                                              & \(|\mathcal{D}_{\mathrm{T}}| \times 0.6\) & \underline{\textit{73.51 ± 0.06}}                         & 85.61 ± 0.09                                        & \textit{56.56 ± 0.07}                & 75.66 ± 0.38                         & \underline{\textit{80.88 ± 0.05}}                         & \underline{\textbf{86.58 ± 0.07}}          & 67.81 ± 0.05                                  & 83.61 ± 0.26                         \\
                                              & \(|\mathcal{D}_{\mathrm{T}}| \times 0.8\) & \underline{\textit{73.51 ± 0.06}}                         & \underline{\textit{85.62 ± 0.09}}                         & \textit{56.56 ± 0.07}                & \underline{\textit{75.75 ± 0.38}}          & \underline{\textit{80.88 ± 0.05}}                         & \underline{\textbf{86.58 ± 0.07}}          & \textit{67.82 ± 0.05}                         & \underline{\textit{83.64 ± 0.26}}          \\
                                              & \(|\mathcal{D}_{\mathrm{T}}| \times 1.0\) & \underline{\textit{73.51 ± 0.06}}                         & \underline{\textit{85.62 ± 0.09}}                         & \textbf{56.57 ± 0.07}                & \underline{\textbf{75.81 ± 0.37}}          & \underline{\textit{80.88 ± 0.05}}                         & \underline{\textbf{86.58 ± 0.07}}          & \textit{67.82 ± 0.05}                         & \underline{\textbf{83.65 ± 0.26}}          \\
\multirow{-6}{*}{CIFAR100}                    & \cellcolor[HTML]{EFEFEF}Jointly           & \cellcolor[HTML]{EFEFEF}\underline{\textbf{73.53 ± 0.08}} & \cellcolor[HTML]{EFEFEF}\underline{\textbf{86.84 ± 0.07}} & \cellcolor[HTML]{EFEFEF}53.29 ± 0.11 & \cellcolor[HTML]{EFEFEF}73.98 ± 0.88 & \cellcolor[HTML]{EFEFEF}\underline{\textbf{81.01 ± 0.07}} & \cellcolor[HTML]{EFEFEF}86.31 ± 0.09 & \cellcolor[HTML]{EFEFEF}\textbf{68.02 ± 0.13} & \cellcolor[HTML]{EFEFEF}82.89 ± 0.60 \\ \hline
                                              & \(|\mathcal{D}_{\mathrm{T}}| \times 0.2\) & 71.59 ± 0.00                                        & 78.82 ± 0.03                                        & 75.84 ± 0.17                         & 57.57 ± 0.02                         & 81.30 ± 0.00                                        & 74.72 ± 0.03                         & 87.23 ± 0.02                                  & 70.87 ± 0.04                         \\
                                              & \(|\mathcal{D}_{\mathrm{T}}| \times 0.4\) & 71.59 ± 0.00                                        & 81.04 ± 0.02                                        & 81.08 ± 0.11                         & 57.89 ± 0.02                         & 81.30 ± 0.00                                        & 75.57 ± 0.00                         & 88.48 ± 0.01                                  & 70.47 ± 0.03                         \\
                                              & \(|\mathcal{D}_{\mathrm{T}}| \times 0.6\) & 71.59 ± 0.00                                        & 81.67 ± 0.02                                        & 82.90 ± 0.07                         & 58.62 ± 0.01                         & 81.30 ± 0.00                                        & 76.01 ± 0.02                         & 88.54 ± 0.02                                  & 71.32 ± 0.03                         \\
                                              & \(|\mathcal{D}_{\mathrm{T}}| \times 0.8\) & 71.59 ± 0.00                                        & \underline{\textit{81.83 ± 0.02}}                         & \textit{83.21 ± 0.12}                & \underline{\textit{59.15 ± 0.02}}          & 81.30 ± 0.00                                        & \underline{\textit{76.36 ± 0.02}}          & \textit{88.83 ± 0.01}                         & \underline{\textit{72.28 ± 0.05}}          \\
\multirow{-5}{*}{ImageNet}                    & \(|\mathcal{D}_{\mathrm{T}}| \times 1.0\) & 71.59 ± 0.00                                        & \underline{\textbf{82.02 ± 0.02}}                         & \textbf{83.70 ± 0.18}                & \underline{\textbf{60.13 ± 0.02}}          & 81.30 ± 0.00                                        & \underline{\textbf{76.38 ± 0.02}}          & \textbf{89.19 ± 0.00}                         & \underline{\textbf{73.92 ± 0.02}}          \\ \hline
\end{tabular}
}
\end{table*}

\subsection{Sensitivity to Calibration Set Selection}
As discussed earlier, in classification tasks, the predicted class probabilities often result in overconfident predictions, particularly when the model overfits the training distribution. To address this issue, it is common to hold out a calibration set to calibrate the predicted probabilities and obtain more reliable aleatoric uncertainty estimates.

Following the setup described in Appendix~\ref{tab:other-protocol}, we train two UQ networks: one on the training set and the other on the held-out calibration set. This set of experiments aims to investigate three key questions:  
(1) Does held-out set calibration lead to better-calibrated aleatoric uncertainty than raw softmax? 
(2) Is it feasible to rely solely on the calibration set for estimating epistemic uncertainty?  
(3) What training/calibration data split is appropriate in practical applications?

As shown in Table~\ref{tab:calib-cls-as}, across all experimental settings, the UQ networks trained on the calibration set \(\mathcal{D}_{\mathrm{C}}\) consistently achieve significantly lower ECE scores. In response to Question 1, these results underscore the importance of a held-out calibration set for enhancing the calibration of classification probabilities.

However, compared to UQ networks trained on the training set \(\mathcal{D}_{\mathrm{T}}\), these calibration-set-based models exhibit lower accuracy in epistemic uncertainty estimation, as indicated by consistently smaller AUROC across all experimental conditions. In response to Question 2, these results indicate that a mismatch between the calibration set and the training distribution degrades the UQ network’s capacity to model epistemic uncertainty.

\begin{figure*}
  \centering
  \includegraphics[width=1\linewidth]{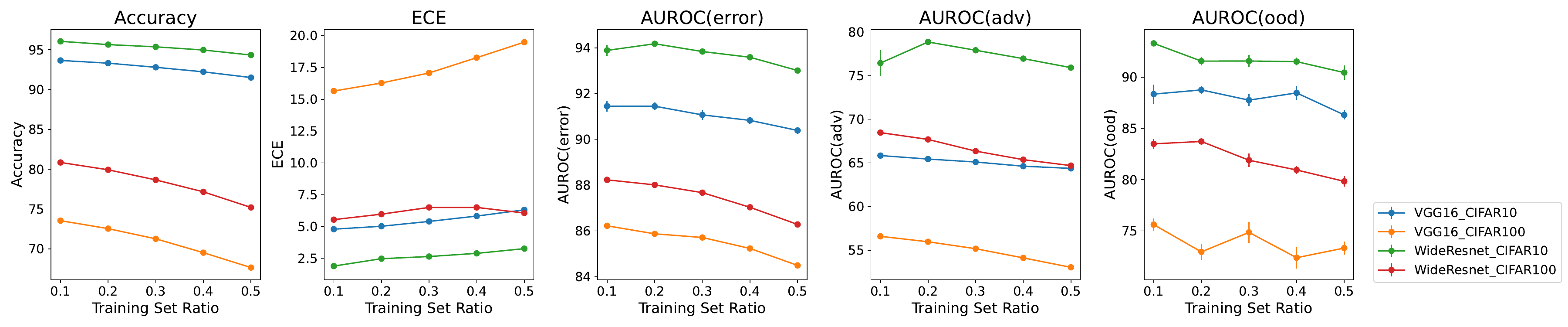}
\caption{Effect of training/calibration split ratio on UQ network trained on training set.}
  \label{fig:calib-training-cls1}
\end{figure*}

\begin{figure*}
  \centering
  \includegraphics[width=1\linewidth]{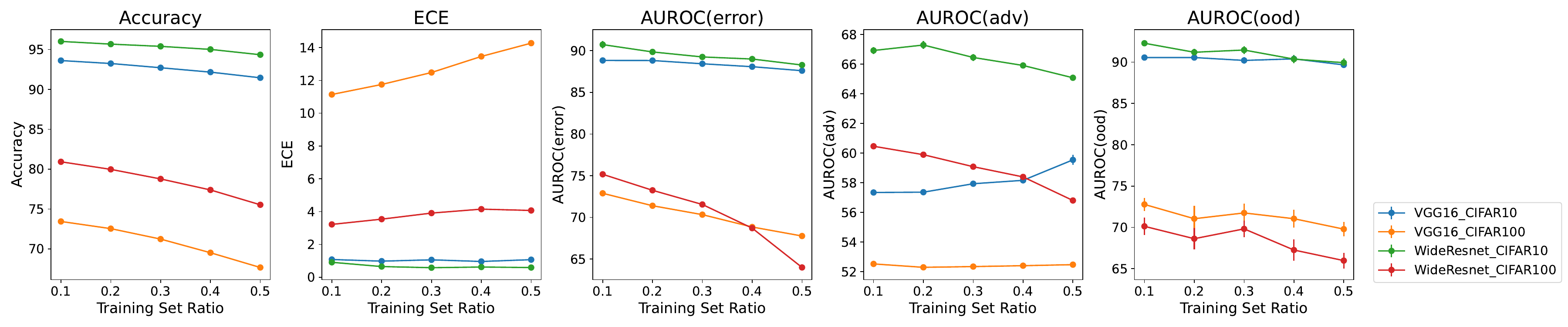}
\caption{Effect of training/calibration split ratio on UQ network trained on calibration set.}
  \label{fig:calib-training-cls2}
\end{figure*}

In response to Question 3, we plot the performance of both UQ networks across varying training/calibration split ratios in Figures~\ref{fig:calib-training-cls1} and~\ref{fig:calib-training-cls2}.  As the calibration set proportion increases, we observe a consistent decline in classification accuracy, calibration quality, and epistemic uncertainty estimation for all models and datasets.  This decline stems from two main factors: the reduced diversity of the training subset and the distributional mismatch introduced by reallocating data to calibration.  

In summary, these results demonstrate that while training the UQ network on a held-out calibration set improves probability calibration, fitting MAR heads on that subset hampers epistemic uncertainty estimation due to distributional mismatch.  Furthermore, allocating more data to calibration tends to degrade both base-model accuracy and UQ performance.  Empirically, a calibration-set proportion of 0.1 offers a good balance, and it is preferable to train MAR heads on the full original training set to preserve alignment with the base model’s learned distribution.

\begin{table*}
\caption{Results on Classification Benchmark under Different Training/Calibration Settings}
\label{tab:calib-cls-as}
\centering
\scalebox{0.71}{
\begin{tabular}{cc|c|lllll|lllll}
\hline
\multirow{2}{*}{Dataset}   & \multirow{2}{*}{\begin{tabular}[c]{@{}c@{}}Calibration\\ proportion\end{tabular}} & \multirow{2}{*}{\begin{tabular}[c]{@{}c@{}}Training\\ base\end{tabular}} & \multicolumn{5}{c|}{VGG16}                                                                                            & \multicolumn{5}{c}{Wide-ResNet}                                                                                      \\ \cline{4-13} 
                           &                                                                                   &                                                                          & Accuracy              & ECE                   & AUROC(error)          & AUROC(adv)            & AUROC(ood)            & Accuracy              & ECE                  & AUROC(error)          & AUROC(adv)            & AUROC(ood)            \\ \hline
\multirow{10}{*}{CIFAR10}  & \multirow{2}{*}{\(|\mathcal{D}| \times 0.1\)}                                     & \(\mathcal{D}_{\mathrm{T}}\)                                             & \textbf{93.66 ± 0.05} & 4.79 ± 0.06           & \textbf{91.45 ± 0.24} & \textbf{65.84 ± 0.24} & 88.34 ± 0.95          & \textbf{96.06 ± 0.04} & 1.89 ± 0.18          & \textbf{93.89 ± 0.24} & \textbf{76.44 ± 1.50} & \textbf{93.29 ± 0.23} \\
                           &                                                                                   & \(\mathcal{D}_{\mathrm{C}}\)                                             & 93.61 ± 0.05          & \textbf{1.08 ± 0.06}  & 88.82 ± 0.19          & 57.34 ± 0.15          & \textbf{90.56 ± 0.33} & 96.01 ± 0.05          & \textbf{0.91 ± 0.15} & 90.72 ± 0.43          & 66.92 ± 0.24          & 92.29 ± 0.27          \\ \cline{3-13} 
                           & \multirow{2}{*}{\(|\mathcal{D}| \times 0.2\)}                                     & \(\mathcal{D}_{\mathrm{T}}\)                                             & \textbf{93.32 ± 0.05} & 5.02 ± 0.05           & \textbf{91.45 ± 0.16} & \textbf{65.44 ± 0.27} & 88.75 ± 0.39          & 95.65 ± 0.04          & 2.47 ± 0.05          & \textbf{94.18 ± 0.12} & \textbf{78.86 ± 0.11} & \textbf{91.56 ± 0.44} \\
                           &                                                                                   & \(\mathcal{D}_{\mathrm{C}}\)                                             & 93.24 ± 0.05          & \textbf{0.98 ± 0.06}  & 88.81 ± 0.13          & 57.36 ± 0.19          & \textbf{90.56 ± 0.31} & \textbf{95.67 ± 0.04} & \textbf{0.65 ± 0.05} & 89.84 ± 0.16          & 67.29 ± 0.27          & 91.20 ± 0.44          \\ \cline{3-13} 
                           & \multirow{2}{*}{\(|\mathcal{D}| \times 0.3\)}                                     & \(\mathcal{D}_{\mathrm{T}}\)                                             & \textbf{92.80 ± 0.05} & 5.40 ± 0.06           & \textbf{91.07 ± 0.21} & \textbf{65.10 ± 0.18} & 87.75 ± 0.58          & 95.37 ± 0.04          & 2.64 ± 0.03          & \textbf{93.84 ± 0.11} & \textbf{77.91 ± 0.15} & \textbf{91.57 ± 0.59} \\
                           &                                                                                   & \(\mathcal{D}_{\mathrm{C}}\)                                             & 92.71 ± 0.04          & \textbf{1.06 ± 0.05}  & 88.42 ± 0.08          & 57.93 ± 0.18          & \textbf{90.21 ± 0.38} & \textbf{95.39 ± 0.05} & \textbf{0.58 ± 0.03} & 89.24 ± 0.19          & 66.45 ± 0.24          & 91.47 ± 0.47          \\ \cline{3-13} 
                           & \multirow{2}{*}{\(|\mathcal{D}| \times 0.4\)}                                     & \(\mathcal{D}_{\mathrm{T}}\)                                             & \textbf{92.23 ± 0.05} & 5.82 ± 0.04           & \textbf{90.83 ± 0.16} & \textbf{64.62 ± 0.29} & 88.46 ± 0.69          & 94.97 ± 0.05          & 2.89 ± 0.03          & \textbf{93.59 ± 0.08} & \textbf{76.95 ± 0.08} & \textbf{91.51 ± 0.39} \\
                           &                                                                                   & \(\mathcal{D}_{\mathrm{C}}\)                                             & 92.16 ± 0.06          & \textbf{0.96 ± 0.04}  & 88.07 ± 0.16          & 58.16 ± 0.16          & \textbf{90.40 ± 0.21} & \textbf{95.01 ± 0.05} & \textbf{0.62 ± 0.03} & 88.99 ± 0.15          & 65.91 ± 0.19          & 90.37 ± 0.49          \\ \cline{3-13} 
                           & \multirow{2}{*}{\(|\mathcal{D}| \times 0.5\)}                                     & \(\mathcal{D}_{\mathrm{T}}\)                                             & \textbf{91.51 ± 0.09} & 6.31 ± 0.08           & \textbf{90.39 ± 0.14} & \textbf{64.37 ± 0.28} & 86.31 ± 0.46          & \textbf{94.34 ± 0.05} & 3.26 ± 0.06          & \textbf{93.01 ± 0.06} & \textbf{75.92 ± 0.16} & \textbf{90.44 ± 0.70} \\
                           &                                                                                   & \(\mathcal{D}_{\mathrm{C}}\)                                             & 91.45 ± 0.09          & \textbf{1.07 ± 0.06}  & 87.59 ± 0.11          & 59.54 ± 0.34          & \textbf{89.67 ± 0.26} & \textbf{94.34 ± 0.05} & \textbf{0.59 ± 0.05} & 88.27 ± 0.16          & 65.08 ± 0.17          & 89.94 ± 0.50          \\ \hline
\multirow{10}{*}{CIFAR100} & \multirow{2}{*}{\(|\mathcal{D}| \times 0.1\)}                                     & \(\mathcal{D}_{\mathrm{T}}\)                                             & \textbf{73.54 ± 0.09} & 15.65 ± 0.08          & \textbf{86.22 ± 0.14} & \textbf{56.58 ± 0.14} & \textbf{75.61 ± 0.61} & 80.85 ± 0.08          & 5.54 ± 0.06          & \textbf{88.23 ± 0.14} & \textbf{68.47 ± 0.05} & \textbf{83.49 ± 0.47} \\
                           &                                                                                   & \(\mathcal{D}_{\mathrm{C}}\)                                             & 73.44 ± 0.09          & \textbf{11.14 ± 0.06} & 72.89 ± 0.20          & 52.52 ± 0.13          & 72.78 ± 0.81          & \textbf{80.92 ± 0.09} & \textbf{3.22 ± 0.09} & 75.17 ± 0.24          & 60.46 ± 0.10          & 70.13 ± 1.04          \\ \cline{3-13} 
                           & \multirow{2}{*}{\(|\mathcal{D}| \times 0.2\)}                                     & \(\mathcal{D}_{\mathrm{T}}\)                                             & \textbf{72.54 ± 0.13} & 16.28 ± 0.10          & \textbf{85.87 ± 0.12} & \textbf{55.96 ± 0.11} & \textbf{72.95 ± 0.80} & 79.93 ± 0.09          & 5.97 ± 0.10          & \textbf{88.01 ± 0.11} & \textbf{67.69 ± 0.08} & \textbf{83.72 ± 0.38} \\
                           &                                                                                   & \(\mathcal{D}_{\mathrm{C}}\)                                             & \textbf{72.54 ± 0.13} & \textbf{11.75 ± 0.12} & 71.40 ± 0.22          & 52.29 ± 0.07          & 71.05 ± 1.57          & \textbf{79.97 ± 0.09} & \textbf{3.54 ± 0.13} & 73.25 ± 0.22          & 59.89 ± 0.10          & 68.63 ± 1.30          \\ \cline{3-13} 
                           & \multirow{2}{*}{\(|\mathcal{D}| \times 0.3\)}                                     & \(\mathcal{D}_{\mathrm{T}}\)                                             & \textbf{71.26 ± 0.09} & 17.07 ± 0.09          & \textbf{85.71 ± 0.08} & \textbf{55.16 ± 0.08} & \textbf{74.85 ± 1.02} & 78.66 ± 0.07          & 6.50 ± 0.09          & \textbf{87.67 ± 0.13} & \textbf{66.35 ± 0.08} & \textbf{81.89 ± 0.66} \\
                           &                                                                                   & \(\mathcal{D}_{\mathrm{C}}\)                                             & 71.23 ± 0.08          & \textbf{12.48 ± 0.09} & 70.33 ± 0.22          & 52.34 ± 0.09          & 71.75 ± 1.14          & \textbf{78.77 ± 0.07} & \textbf{3.91 ± 0.08} & 71.55 ± 0.24          & 59.08 ± 0.10          & 69.82 ± 1.00          \\ \cline{3-13} 
                           & \multirow{2}{*}{\(|\mathcal{D}| \times 0.4\)}                                     & \(\mathcal{D}_{\mathrm{T}}\)                                             & \textbf{69.52 ± 0.11} & 18.27 ± 0.13          & \textbf{85.23 ± 0.12} & \textbf{54.11 ± 0.10} & \textbf{72.38 ± 1.04} & 77.16 ± 0.08          & 6.50 ± 0.11          & \textbf{87.03 ± 0.10} & \textbf{65.37 ± 0.08} & \textbf{80.94 ± 0.38} \\
                           &                                                                                   & \(\mathcal{D}_{\mathrm{C}}\)                                             & \textbf{69.52 ± 0.10} & \textbf{13.46 ± 0.16} & 68.85 ± 0.18          & 52.40 ± 0.12          & 71.05 ± 1.07          & \textbf{77.39 ± 0.06} & \textbf{4.15 ± 0.11} & 68.71 ± 0.28          & 58.39 ± 0.08          & 67.26 ± 1.29          \\ \cline{3-13} 
                           & \multirow{2}{*}{\(|\mathcal{D}| \times 0.5\)}                                     & \(\mathcal{D}_{\mathrm{T}}\)                                             & 67.65 ± 0.10          & 19.49 ± 0.09          & \textbf{84.49 ± 0.13} & \textbf{53.03 ± 0.12} & \textbf{73.32 ± 0.62} & 75.20 ± 0.04          & 6.06 ± 0.13          & \textbf{86.28 ± 0.12} & \textbf{64.69 ± 0.10} & \textbf{79.83 ± 0.53} \\
                           &                                                                                   & \(\mathcal{D}_{\mathrm{C}}\)                                             & \textbf{67.67 ± 0.11} & \textbf{14.27 ± 0.13} & 67.77 ± 0.16          & 52.47 ± 0.12          & 69.79 ± 0.89          & \textbf{75.53 ± 0.02} & \textbf{4.07 ± 0.10} & 64.00 ± 0.11          & 56.80 ± 0.11          & 65.98 ± 0.94          \\ \hline
\end{tabular}
}
\end{table*}

\begin{table*}
\caption{Results on Regression Benchmark under Different UQ-head Architectures}
\label{tab:model-reg-as}
\centering
\scalebox{1}{
\begin{tabular}{ccllllll}
\hline
\multirow{2}{*}{Metrics}                                                 & \multirow{2}{*}{\begin{tabular}[c]{@{}c@{}}Hidden\\ layer\end{tabular}} & \multicolumn{6}{c}{Datasets}                                                                                                                                                  \\ \cline{3-8} 
                                                                         &                                                                         & Concrete                    & Energy                     & Kin8nm                     & Naval                      & Power                       & Wine                       \\ \hline
\multirow{3}{*}{\begin{tabular}[c]{@{}c@{}}Winkler\\ Score\end{tabular}} & 1                                                                       & \underline{\textbf{33.37 ± 0.89}} & 7.82 ± 0.15                & 0.41 ± 0.01                & \underline{\textbf{0.00 ± 0.00}} & 18.73 ± 0.24                & \underline{\textbf{3.13 ± 0.06}} \\
                                                                         & 2                                                                       & \underline{\textit{34.22 ± 0.88}} & \underline{\textbf{7.66 ± 0.13}} & \underline{\textbf{0.40 ± 0.00}} & \underline{\textbf{0.00 ± 0.00}} & \underline{\textit{18.43 ± 0.25}} & \underline{\textit{3.18 ± 0.07}} \\
                                                                         & 3                                                                       & 34.35 ± 0.89                & \underline{\textit{7.80 ± 0.16}} & \underline{\textbf{0.40 ± 0.00}} & \underline{\textbf{0.00 ± 0.00}} & \underline{\textbf{18.34 ± 0.27}} & 3.25 ± 0.07                \\ \hline
\multirow{3}{*}{PIECE}                                                   & 1                                                                       & \underline{\textbf{0.06 ± 0.00}}  & \underline{\textbf{0.08 ± 0.01}} & \underline{\textbf{0.02 ± 0.00}} & \underline{\textbf{0.02 ± 0.00}} & \underline{\textbf{0.02 ± 0.00}}  & \underline{\textbf{0.03 ± 0.00}} \\
                                                                         & 2                                                                       & \underline{\textit{0.07 ± 0.00}}  & \underline{\textit{0.10 ± 0.01}} & \underline{\textit{0.03 ± 0.00}} & 0.03 ± 0.00                & \underline{\textbf{0.02 ± 0.00}}  & \underline{\textit{0.04 ± 0.00}} \\
                                                                         & 3                                                                       & \underline{\textit{0.07 ± 0.00}}  & 0.11 ± 0.01                & 0.04 ± 0.00                & \underline{\textbf{0.02 ± 0.00}} & \underline{\textbf{0.02 ± 0.00}}  & 0.05 ± 0.00                \\ \hline
\multirow{3}{*}{$\text{PIECE}^+$}                                        & 1                                                                       & \underline{\textbf{0.04 ± 0.01}}  & \underline{\textbf{0.09 ± 0.01}} & \underline{\textbf{0.02 ± 0.00}} & \underline{\textbf{0.01 ± 0.00}} & \underline{\textbf{0.01 ± 0.00}}  & \underline{\textbf{0.03 ± 0.01}} \\
                                                                         & 2                                                                       & \underline{\textit{0.05 ± 0.01}}  & \underline{\textit{0.10 ± 0.01}} & \underline{\textbf{0.02 ± 0.00}} & 0.03 ± 0.00                & \underline{\textbf{0.01 ± 0.00}}  & \underline{\textit{0.04 ± 0.01}} \\
                                                                         & 3                                                                       & 0.06 ± 0.01                 & 0.13 ± 0.01                & 0.03 ± 0.00                & \underline{\textbf{0.01 ± 0.00}} & \underline{\textbf{0.01 ± 0.00}}  & 0.05 ± 0.01                \\ \hline
\multirow{3}{*}{$\text{PIECE}^-$}                                        & 1                                                                       & \underline{\textbf{0.03 ± 0.01}}  & \underline{\textbf{0.04 ± 0.01}} & \underline{\textbf{0.01 ± 0.00}} & \underline{\textbf{0.01 ± 0.00}} & \underline{\textbf{0.01 ± 0.00}}  & \underline{\textbf{0.02 ± 0.00}} \\
                                                                         & 2                                                                       & \underline{\textit{0.04 ± 0.01}}  & \underline{\textit{0.05 ± 0.01}} & \underline{\textit{0.03 ± 0.00}} & \underline{\textit{0.02 ± 0.00}} & \underline{\textbf{0.01 ± 0.00}}  & \underline{\textit{0.03 ± 0.01}} \\
                                                                         & 3                                                                       & 0.05 ± 0.01                 & \underline{\textit{0.05 ± 0.01}} & 0.04 ± 0.00                & \underline{\textit{0.02 ± 0.00}} & \underline{\textbf{0.01 ± 0.00}}  & \underline{\textit{0.03 ± 0.01}} \\ \hline
\multirow{3}{*}{Correlation}                                             & 1                                                                       & 0.40 ± 0.03                 & 0.60 ± 0.02                & \underline{\textbf{0.30 ± 0.01}} & 0.70 ± 0.01                & 0.27 ± 0.01                 & 0.23 ± 0.01                \\
                                                                         & 2                                                                       & \underline{\textit{0.44 ± 0.02}}  & \underline{\textit{0.65 ± 0.02}} & \underline{\textit{0.29 ± 0.01}} & \underline{\textit{0.78 ± 0.01}} & \underline{\textit{0.31 ± 0.01}}  & \underline{\textit{0.24 ± 0.01}} \\
                                                                         & 3                                                                       & \underline{\textbf{0.47 ± 0.02}}  & \underline{\textbf{0.68 ± 0.02}} & \underline{\textit{0.29 ± 0.01}} & \underline{\textbf{0.82 ± 0.01}} & \underline{\textbf{0.33 ± 0.01}}  & \underline{\textbf{0.26 ± 0.01}} \\ \hline
\end{tabular}
}
\end{table*}

\begin{figure*}
  \centering
  \includegraphics[width=0.98\linewidth]{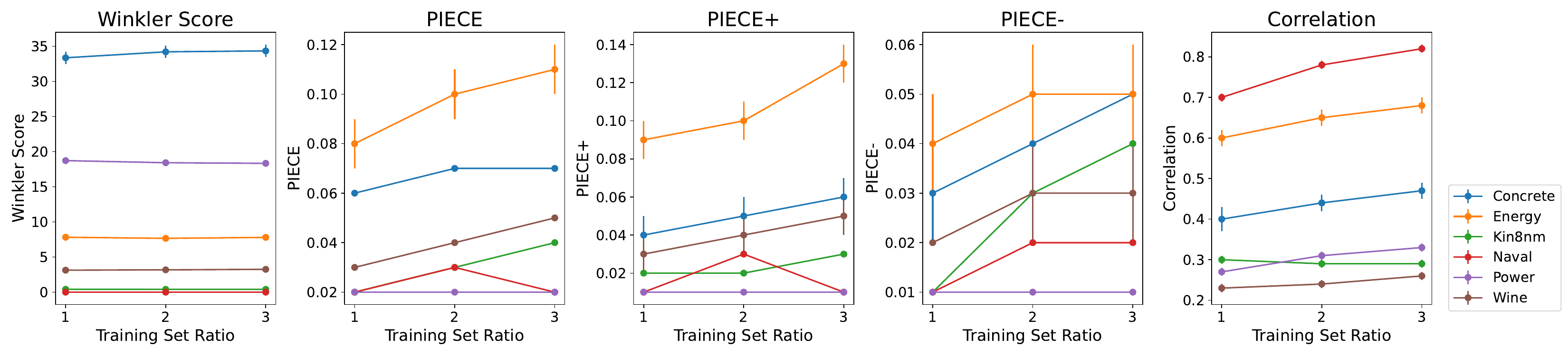}
  \caption{Effect of UQ-head Architectures on Post-hoc Uncertainty Estimation in Regression Benchmark}
  \label{fig:model-reg}
\end{figure*}

\begin{table*}
\caption{Results on Classification Benchmark under Different UQ-head Architectures}
\label{tab:model-cls-as}
\centering
\scalebox{0.78}{
\begin{tabular}{cc|lllll|lllll}
\hline
\multirow{2}{*}{Dataset}  & \multirow{2}{*}{\begin{tabular}[c]{@{}c@{}}Hidden\\ layer\end{tabular}} & \multicolumn{5}{c|}{VGG16}                                                                                                                          & \multicolumn{5}{c}{Wide-ResNet}                                                                                                                    \\ \cline{3-12} 
                          &                                                                         & Accuracy                    & ECE                         & AUROC(error)                & AUROC(adv)                  & AUROC(ood)                  & Accuracy                    & ECE                        & AUROC(error)                & AUROC(adv)                  & AUROC(ood)                  \\ \hline
\multirow{3}{*}{CIFAR10}  & 1                                                                       & \underline{\textbf{93.62 ± 0.03}} & \underline{\textbf{1.08 ± 0.06}}  & \underline{\textbf{89.67 ± 0.15}} & 68.36 ± 0.17                & \underline{\textbf{88.41 ± 0.43}} & \underline{\textit{96.01 ± 0.03}} & \underline{\textbf{0.91 ± 0.15}} & 93.74 ± 0.18                & 78.48 ± 0.77                & \underline{\textbf{92.84 ± 0.23}} \\
                          & 2                                                                       & 93.43 ± 0.06                & \underline{\textit{2.10 ± 0.06}}  & 88.95 ± 0.16                & \underline{\textit{70.23 ± 0.11}} & \underline{\textit{86.07 ± 0.59}} & 95.89 ± 0.06                & \underline{\textit{1.15 ± 0.19}} & \underline{\textit{93.94 ± 0.14}} & \underline{\textit{79.09 ± 0.62}} & \underline{\textit{92.75 ± 0.26}} \\
                          & 3                                                                       & \underline{\textit{93.46 ± 0.05}} & 3.23 ± 0.07                 & \underline{\textit{89.03 ± 0.17}} & \underline{\textbf{70.24 ± 0.10}} & 85.21 ± 0.71                & \underline{\textbf{96.03 ± 0.03}} & 1.97 ± 0.04                & \underline{\textbf{93.98 ± 0.12}} & \underline{\textbf{79.66 ± 0.32}} & 92.63 ± 0.27                \\ \hline
\multirow{3}{*}{CIFAR100} & 1                                                                       & \underline{\textbf{73.51 ± 0.06}} & \underline{\textbf{11.14 ± 0.06}} & \underline{\textbf{85.62 ± 0.09}} & \underline{\textit{56.57 ± 0.07}} & 75.81 ± 0.37                & \underline{\textbf{80.88 ± 0.05}} & \underline{\textbf{3.22 ± 0.09}} & \underline{\textbf{86.58 ± 0.07}} & \underline{\textbf{67.82 ± 0.05}} & 83.65 ± 0.26                \\
                          & 2                                                                       & \underline{\textit{73.36 ± 0.10}} & \underline{\textit{12.03 ± 0.10}} & 85.41 ± 0.09                & \underline{\textit{56.57 ± 0.07}} & \underline{\textbf{76.10 ± 0.37}} & \underline{\textit{80.72 ± 0.06}} & \underline{\textit{4.19 ± 0.07}} & \underline{\textit{86.57 ± 0.07}} & 67.80 ± 0.05                & \underline{\textbf{83.75 ± 0.25}} \\
                          & 3                                                                       & 73.33 ± 0.08                & 13.64 ± 0.09                & \underline{\textit{85.44 ± 0.09}} & \underline{\textbf{56.81 ± 0.07}} & \underline{\textit{76.08 ± 0.37}} & 80.64 ± 0.07                & 4.91 ± 0.07                & 86.53 ± 0.07                & \underline{\textbf{67.82 ± 0.05}} & \underline{\textit{83.72 ± 0.25}} \\ \hline
\end{tabular}
}
\end{table*}

\begin{figure*}
  \centering
  \includegraphics[width=0.98\linewidth]{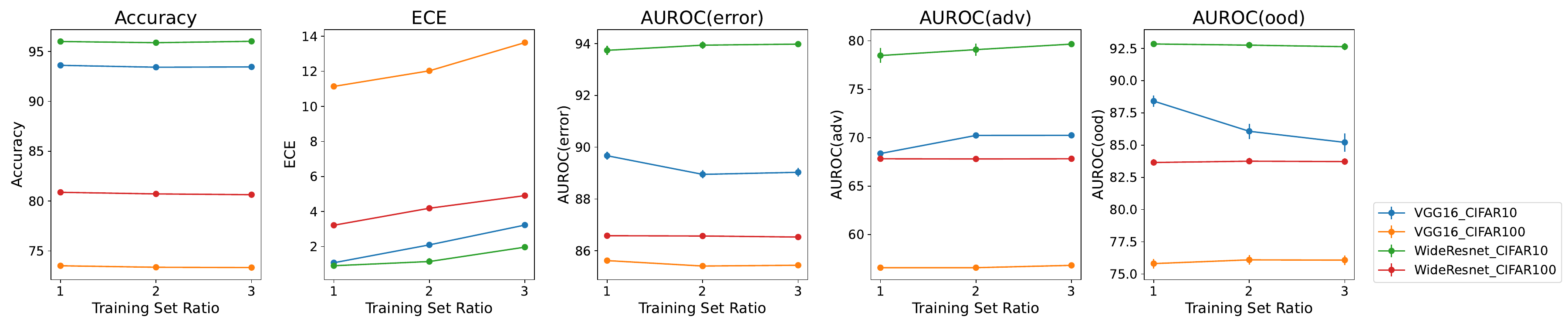}
  \caption{Effect of UQ-head Architectures on Post-hoc Uncertainty Estimation in Classification Benchmark}
  \label{fig:model-cls}
\end{figure*}

\subsection{Sensitivity to UQ Network Architectures}
Finally, we evaluate the impact of UQ network architectural complexity by varying the depth of the MLP used in the UQ networks. This analysis provides practical insights for model design and selection.

\subsubsection{Results on Regression}
Table~\ref{tab:model-reg-as} and Figure~\ref{fig:model-reg} show the results on the regression benchmark. We observe that increasing the depth of UQ networks does not improve PI quality, as measured by Winkler Score. On the contrary, it leads to larger PIECE values, suggesting that deeper UQ networks produce sharper intervals but are more prone to overfitting the training data, which compromises coverage on the test set.
In contrast, the correlation between uncertainty and prediction error improves with depth, indicating that deeper UQ networks better capture the base model’s knowledge distribution and therefore yield more accurate epistemic uncertainty estimates.

Based on these controlled experiments, for regression tasks, it is advisable to use shallow MLPs for estimating PI boundaries (i.e., ${q}^{+}$ and ${q}^{-}$) to improve calibration performance. In contrast, deeper MLPs can be used to model MARs, as they enhance the alignment with the base model’s knowledge and improve epistemic uncertainty estimation. Designing separate UQ networks with different MLP depths for different uncertainty types offers a balanced and effective solution.

\subsubsection{Results on Image Classification}
Table~\ref{tab:model-cls-as} and Figure~\ref{fig:model-cls} summarize our findings on standard image classification benchmarks.  As the depth of the UQ network increases, we consistently observe higher expected calibration error (ECE), suggesting that deeper heads yield overly sharp confidence maps, fit the calibration data too closely, and therefore generalize poorly to held-out test examples.  Interestingly, this sensitivity to network depth is specific to classification tasks; in our three detection benchmarks, where the primary concern is epistemic uncertainty, performance remains essentially unchanged across different UQ head depths.

Based on these controlled experiments, we recommend defaulting to shallow MLP architectures for all UQ networks in classification settings, since they strike a favorable balance between calibration fidelity and robustness to overfitting.  

In general, we emphasize, however, that these practical guidance derives from the datasets and model families we evaluated.  In new domains or with substantially different data characteristics, practitioners should still carry out a conventional model selection procedure, such as grid search or cross-validation over depth, width, and other hyperparameters, via cross-validation, to identify the UQ architecture best suited to their specific application.


\bibliographystyle{IEEEtran}
\bibliography{references}

\vfill

\end{document}